# Improving Classifier Training Efficiency for Automatic Cyberbullying Detection with Feature Density

Juuso Eronen †, Michal Ptaszynski †, Fumito Masui †, Aleksander Smywiński-Pohl ‡, Gniewosz Leliwa§, Michal Wroczynski§

*† Kitami Institute of Technology, Japan*
*‡ AGH University of Science and Technology, Poland*
*§ Samurailabs, Poland*

**Abstract**

We study the effectiveness of Feature Density (FD) using different linguistically-backed feature preprocessing methods in order to estimate dataset complexity, which in turn is used to comparatively estimate the potential performance of machine learning (ML) classifiers prior to any training. We hypothesise that estimating dataset complexity allows for the reduction of the number of required experiments iterations. This way we can optimize the resource-intensive training of ML models which is becoming a serious issue due to the increases in available dataset sizes and the ever rising popularity of models based on Deep Neural Networks (DNN). The problem of constantly increasing needs for more powerful computational resources is also affecting the environment due to alarmingly-growing amount of $CO_2$ emissions caused by training of large-scale ML models. The research was conducted on multiple datasets, including popular datasets, such as Yelp business review dataset used for training typical sentiment analysis models, as well as more recent datasets trying to tackle the problem of cyberbullying, which, being a serious social problem, is also a much more sophisticated problem form the point of view of linguistic representation. We use cyberbullying datasets collected for multiple languages, namely English, Japanese and Polish. The difference in linguistic complexity of datasets allows us to additionally discuss the efficacy of linguistically-backed word preprocessing.

## 1. Introduction

Performance estimation of different classification algorithms with a given dataset has always been one of the difficulties in Machine Learning (ML) [1]. While there are classifiers that appear to be extremely successful on a number of different topics, they may easily be outperformed by others on a dataset-specific scale, as it is difficult to find a classifier that would have the greatest performance for any dataset [1]. So it is up to the individual (researcher, or ML practitioner) to decide experimentally which classifier might be suitable for the task they are trying to solve, based on their field experience and accumulated theoretical knowledge.

A typical method of estimating the performance of various classifiers is to pick a number of potential classifiers to experiment on and train the chosen classifiers with cross-validation in order to achieve the best possible average estimation of their performance. Although getting accurate estimations of the classifier performance is possible this way, it becomes very costly and time-inefficient.

In the past, several attempts have been made to estimate the output of the ML model. One of the suggestions for this issue was to extrapolate the results from training small datasets to model the performance when utilizing larger datasets [2]. Another strategy was the usage of meta-learning and training a model using dataset characteristics to predict the output of the classifier [3].

The importance of addressing this issue derives not only from the increased demands in computational power, but also from its environmental consequences. This is mainly attributed to the growing success of the fields of Artificial Intelligence (AI) and ML. Training classifiers on massive datasets are both time consuming and computationally expensive while leaving behind a noticeable carbon footprint [4]. In order to take a step forward towards greener AI [5], the core methods of ML strategies must be assessed and future areas of development identified. It would be helpful to theoretically estimate, even if roughly, the efficacy of the classifier prior to training, in order to conserve computing resources and reduce pollution.

The ability to estimate the performance of a classifier prior to training will also have significant functional consequences. One of the areas where this is urgently required is the monitoring of negative and abusive behaviour encountered online, known as cyberbullying (CB), defined as the exploitation of open online means of communication, such as Internet forum boards, or social network services (SNS) to convey harmful and disturbing information about private individuals, often children and students [6]. Users' realization of the anonymity of online communications is one of the factors that make this activity attractive for bullies since they rarely face consequences of their improper behavior. The problem was further exacerbated by the popularization of smartphones and tablet computers that enable almost continuous usage of SNS anywhere, at home, work/school or in motion [7].

Messages that can be identified as cyberbullying usually ridicule someone's personality, body type or appearance, or include slandering or spreading rumors about the individual. This may drive its victims to even as far as self-mutilation or suicide, or, on the contrary, to a retaliation assault on their perpetrators [8]. Global spike in cyberbullying cases[1] opened a world wide discussion about whether such messages should be identified early to deter harm, and on freedom of speech on the Internet.

In certain nations, such as in Japan, the issue has been severe enough to be seen at ministerial level [9]. As one of the ways to solve the issue, Internet Patrol (IP) consisting of school workers has begun to track online forum pages and SNS featuring cyberbullying content. Unfortunately, as IP is carried out manually, reading through vast numbers of websites and SNS content makes it an uphill battle. To aid in this struggle, some research have started to develop methods for automatic detection of CB [10, 11, 12, 13].

Even with numerous improvements, the findings have unfortunately remained only slightly satisfactory. This is due to the plethora of language and vocabulary ambiguities and styles used in CB. Solving the problem of CB has become even more crucial after introducing global policies such as General Data Protection Regulation (GDPR) in EU[2], which put the weight of spotting and weeding out online harassment on the platforms themselves. Therefore the process of efficient implementing of automatic cyberbullying detection for different languages and social networking sites is one of the current most burning problems, which could greatly benefit from a method allowing to loosely approximate which classifier configurations can be rejected without the experimental process.

[1]https://cyberbullying.org/summary-of-our-cyberbullying-research
[2]https://edps.europa.eu/data-protection/data-protection/reference-library/anti-harassment-procedures_en

In order to contribute more to solving this problem, we are performing an in-depth study of the efficacy of the notion of Feature Density (FD) previously proposed by Ptaszynski et al. [14] to comparatively estimate the efficiency of various ML classifiers prior to training. Additionally we evaluate the usefulness of numerous linguistically-backed feature preprocessing approaches, including lemmas, Named Entity Recognition (NER) and dependency information-based features, in an application for automated cyberbullying detection.

To optimize the use of resources when dealing with NLP problems, it would be useful to be able to have information about the complexity of different datasets, meaning how difficult are the datasets for a classifier to learn and generalize upon. This problem has been recognized in other fields such as image recognition [15]. Even though natural language complexity has been studied through lexical [16, 17, 18, 19] and syntactic complexity [20, 21, 22, 23], mostly within the second language education field, there have been very few or no applications to use this kind of measures in estimating dataset complexity in ML tasks.

This research uses a total of four datasets, three of which are about cyberbullying and one verification dataset from a different field. All of the three CB datasets are in different languages, namely Japanese, English and Polish, to additionally study dataset complexity across different languages. We chose those three languages for two main reasons. Primarily, because they were some of the best prepared datasets for cyberbullying detection, but also because each of those languages represents different family of languages (English - Germanic; Polish - Slavic; Japanese - Koreano-Japonic language family) and different type of language (English - isolating/analytic language; Polish - fusional language, Japanese - agglutinative language).

The size of the CB datasets range from about 2,000 entries to 20,000. The Japanese dataset was created originally by Ptaszynski et al. [10] based on the definition by the Japanese Ministry of Education [9]. This dataset was also used in the first experiments with Feature Density [14]. The English dataset is a reannotated [24] version of the Formspring Dataset for Cyberbullying Detection [25]. The reannotation was necessary to ensure quality of expert annotations, as the original dataset was annotated by laypeople using Amazon Mechanical Turk. The verification dataset, which is a subset of the Yelp reviews [3] is substantially larger, as the subset used in this research holds 500,000 entries.

The various sizes of datasets were chosen to study the effect of Feature Density (FD) and verify its potential to comparatively estimate the performance of classifiers. The preprocessing methods applied here were extended from previous research with Japanese data [14]. For traditional classifiers we applied TF-IDF weighting on a Bag-of-Features language model. Additionally to applying a language complexity measure of a dataset (FD) to estimate performance of various traditional ML classifiers, we also specifically propose and test different linguistically-backed embeddings trained as neural language models used in neural networks (CNN). In order to perform the most thorough analysis, we are applying the methods to multiple languages and topics. This also allows us to study the effect of linguistic and cultural differences in classifier performance when the same classification methods are used.

The paper outline is as follows. In Section 2 we describe previous research in all areas that are addressed in this paper, namely, automatic estimation of classifier performance, dataset complexity estimation, feature density, linguistically-backed pre-processings, and cyberbullying detection. In Section 3 we describe all the datasets applied to this research and present their features. In Section 4 we go through the feature engineering and classification methods used in this research. In Section 5 we go through all the results from the conducted experiments. In Section 6 we discuss the results in general and bring out the most interesting findings in relation to the research goals.

[3] `http://www.yelp.com/dataset`

## 2. Research Background

### *2.1. Classifier Performance Estimation*

The first research that studied automatic estimation of ML classifier performance was based on extrapolating results from a smaller dataset to simulate the performance of a larger dataset. Basavanhally et al. [2] applied this method to estimating classifier performance in the field of computer aided medical diagnostics, where the available data is in many cases limited in quantity. Their research showed that using a repeated random sampling method on small datasets to estimate the performance on a larger set often had high error rates and should not be generalized as holding true when higher quantities of data become available. Later, a cross-validation sampling strategy was added, which improved the method and resulted in lower error rates [26].

However, they only used three different classifiers, two of which are widely regarded as low-performance baseline classifiers (Naive Bayes, k-Nearest Neighbor) and Support Vector Machines. Their results suggested that the ranking of the classifiers would not change as the dataset size increases but considering the classifiers used, the results seem very predictable and match the average rankings of these classifiers [27]. Also, they did not consider any deep learning -based approaches. This increases the questionability of the results.

In the field of NLP, Johnson et al.[28] used the fastText classifier [29] and applied the extrapolation method to document classification. They discovered that biased power law model with binomial weights could be used as a good baseline extrapolation model for NLP tasks. As the authors suggested, the method needs to be studied further using different classifiers and extrapolation models. Even though the extrapolation method would aid in estimating the classifier performance, at least some training is still required.

Gama et al. [3] proposed another method, in which classifier performance could be estimated by training a regression model based on meta-level characteristics of a dataset. The characteristics used included simple measures like number of instances and number of attributes, statistical measures like standard deviation ratio and various information based measures like class entropy. These measures are defined in the STATLOG project [30].

This meta-learning based method was adopted and further developed by introducing the Landmarking process [31], which uses the learners themselves to characterize the datasets. This involves using computationally non-demanding classifiers, such as Naive Bayes (NB), to gain valuable insights into datasets. The system outperformed the previous characterization approach and had modest results in the ranking of learners.

Later, Blachnik et al. [32] enhanced the Landmarking method. He proposed that the information from instance selection methods could be used as landmarks. These instance selection methods are usually used for cleaning the dataset, reducing its size by removing redundant information. They found out that it is possible to use the relation between the initial and reduced datasets as a landmark in order to predict classifier performance with lower error rates.

A more recent method focused on generalizing meta information of the dataset to quantify an overall dataset complexity measure and use it to find correlations between this measure and classifier performance without any training [15]. We describe this research in detail in the following Section 2.2.

In our research, instead of concentrating on training an additional classifier on meta information extracted from the dataset or performance simulation and extrapolation, we directly focused on feature engineering and the relation between the available feature space and classifier performance. This is a novel method that can be utilized together with the existing methods to better estimate the performance of different classifiers.

### *2.2. Dataset Complexity Estimation*

In the field of image recognition, Rahane et al. [15] proposed a method to estimate dataset complexity using various entropy-based information measures on image grayscale values. All of the used measures correlated with the known performance metrics of the applied datasets. They managed to properly rank their datasets using these metrics, which showed that there is a potential in ranking classifiers using these metrics. In the future, a similar method based on word similarity could be applied in NLP.

It has been widely known in linguistics that various languages have different complexity. Therefore language complexity in general is a topic that has been studied throughout the years in linguistics [33]. Especially, attempts to estimate a general lexical complexity of a language [16, 17, 18, 19] or a general syntactic complexity of a language [20, 21, 22, 23] are well documented within the fields of language education. It has been used especially in second language learning, to estimate the level of difficulty to certain first language speakers to learn a specific second language (e.g., how hard it is for native speakers of English to learn Polish or Japanese, which have completely different syntactic complexity). However, to the best of our knowledge, there has not been relevant research in applying this to locally estimate linguistic dataset complexity. One of the measures used for estimating linguistic complexity in general has been Lexical Density (LD) [20]. It is a score representing an estimated measure of content per lexical units for a given corpus, calculated as the number of all distinct words, or vocabulary size, divided by the number of all words in the corpus: $LD = V/N$. This notation means that the corpus is the more complex (or "dense"), the more distinct words it contains. In this research we take a look at an extension of Lexical Density, namely, Feature Density [14], and study its uses in estimating the complexity of different natural language datasets, depending on the kinds of feature sets were used to represent the dataset.

Also, information criterion based methods, liken Akaike Information Criterion (AIC) [34] and Bayesian Information Criterion (BIC) [35], have been proposed for model selection since the same time as lexical density was introduced. However using these metrics requires training all of the models which is exactly what we are trying to avoid as feature density can be calculated from the dataset directly without any training.

### *2.3. Feature Density*

The concept of Feature Density (FD) was introduced by Ptaszynski et al. [14] based on the notion of LD [20] from linguistics. The score is called Feature Density as it includes not only lexemes (words/tokens) but also other features, like parts-of-speech, dependency information, or any other applied feature set, in addition or instead of words/tokens

$$FD = \frac{V_x}{N_x}, \tag{1}$$

where $V_x$ is the number of distinct features in the corpus and $N_x$ is the total number of features in the corpus. By feature we mean preprocessed (lemmatized, added POS, dependency information etc.) words/tokens. The preprocessing types used are introduced in section 4.1. In this research we applied this notation to estimate dataset complexity, by defining it as the complexity of the language dataset itself.

Furthermore, in this research, after calculating FD for all applied datasets and all preprocessing methods, we calculated Pearson's correlation coefficient ($\rho$-value) between dataset generalization (FD) and classifier results (macro F-scores). If FD has a clear positive or negative correlation with classifier performance, it could be useful in comparatively estimating the performance of various classifiers within a dataset. Using the Japanese CB dataset, Ptaszynski et al. [14] showed that CNNs work well with higher FD while other classifiers' scores were usually higher with the lower FD datasets. This shows that it might be possible to improve the performance of CNNs by increasing the FD of the dataset in question, whereas other classifiers could score higher if FD was reduced [14].

We also compare different datasets in terms of comparative complexity. Using Feature Density and the experiment results, we study the usefulness of estimating dataset complexity in optimizing classifier performance or even estimating it prior to experimentation. In this research, the concept of Feature Density is applied to multiple languages in the field of cyberbullying as well as Sentiment Analysis.

### *2.4. Linguistically-backed Preprocessing*

In almost all cases, word embeddings are learned only from pure tokens (words) or in some cases, lemmas (unconjugated forms of words). This also applies to the recently popularized pre-trained language models like BERT [36]. Although word embeddings have been used to help predict parts of speech [37], named entity recognition [38], or dependency parsing [39], utilizing such linguistic information in the process of training the embeddings themselves have not yet been researched extensively, with only a few related introductory studies being proposed so far [40, 41, 42].

To conduct further investigation on the potential of capturing deeper relations between lexical items and structures and to filter out redundant information, we propose to preserve the morphological, syntactic and other types of linguistic information by combining them with the pure tokens or lemmas. This means, for example, including parts-of-speech or dependency information within the used lexical features. The word embeddings can then be trained using these features instead of just plain tokens. It is also possible to later apply this method to the pre-training of huge language models and possibly enhance their performance. The linguistically-backed preprocessing methods used in this research are described in-depth in section 4.1.

### *2.5. Cyberbullying Detection*

Even though the issue of CB has been researched in social sciences and child psychology for over ten years [6, 43], only a small number of significant attempts have been made so far to use information technology to help solve the problem. Here we introduce the most relevant studies up to the day.

For the first time, [10, 11] in 2010 performed affect analysis on a small CB dataset and discovered that the use of vulgar words were the distinctive features for CB. They trained an SVM classifier using a lexicon of such words and with multiple optimizations, they managed to detect CB with an F-score of 88.2%. However, as the amount of data increased, it caused a decrease in results, which caused the authors to abandon SVM as not ideal for language ambiguities typical for CB.

In other research, Sood et al. [44] focused on detecting personal insults and negative influence which could at most cause the Internet community to fall into recession, meaning if the harmful content would be left uncontrolled, people would start to leave the community. Their study used single words and bigrams as features, and weighted them using Boolean weighting (1/0), term frequency and tf-idf. These were used to train an SVM classifier. Their dataset was a corpus collected from multiple online fora, totaling at six thousand entries. They used a crowd-sourcing approach (Mechanical Turk) with non-professional laypersons hired for the classification task to annotate the data.

Later, Dinakar et al. [12] introduced their method to detect and mitigate cyberbullying. Their paper had a wider perspective, as they did not focus only on the detection of cyberbullying, but also included methods for mitigating the problem. This was an improvement compared to previous research. Their classifiers scored up to 58-77% of F-score in an English dataset. The results varied depending on the type of harassment they were attempting to classify. The best classifier they proposed was again SVM, which further confirms the effectiveness of SVMs for detecting cyberbullying, similarly to the research done in 2010 using a Japanese dataset [10].

An interesting work was done by Kontostathis et al. [45], who performed a thorough analysis of cyberbullying entries on Formspring.me. They identified usual cyberbullying patterns and used a machine learning method based on Essential Dimensions of Latent Semantic Indexing (EDLSI) to apply them in classification.

Cano et al. [46] introduced a Violence Detection Model (VDM), a weakly supervised Bayesian model. However, their focus was not strictly restricted on cyberbullying, but consisted of a more widened scope of generally understood "violence". This simplified the problem and made it more feasible for untrained annotators to work with. The datasets were extracted from violence-related topics on Twitter and DBPedia.

Nitta et al. [47] proposed a method extending Turney's SO-PMI-IR score [48] to automatically detect harmful entries. They used the score to calculate the relevance of a document with harmful contents. The seed words were grouped into three categories (abusive, violent, obscene) and the relevance of categories was maximized. The method was evaluated comparatively high as the best achieved Precision was around 91% (although with Recall less then 10%).

A re-evaluation of their method two years later unfortunately suggested that the method lost a major amount of its Precision (over 30 percentage-point drop) over the span of two years [13]. They hypothesized that this could be the cause of external factors like Web page re-ranking, or changes in SNS user policies, etc. The method was improved by acquiring and filtering new harmful seed words automatically with some success (P=76%), but they were unable to achieve results close to the original performance. Later an automatic method for the seed word acquisition [49] was developed with positive results. However, this method was deemed inefficient and impractical compared to a more direct machine learning based approach with a properly annotated dataset.

Sarna et al. [50] based their method on a set of features like "bad words", positive/negative sentiment words, and other common features like pronouns, etc., to estimate user credibility. These features were applied to commonly used classifiers (Naive Bayes, kNN, Decision Trees, SVM). The obtained classification results were further used in User Behavior Analysis model (BAU), and User Credibility Analysis (CAU) model. Even though their approach included the use of phenomena such as irony, or rumors, in practice they unfortunately only focused on messages containing "bad words." Moreover, neither the words themselves, the dataset, nor its annotation schema were sufficiently described in the paper.

Ptaszynski et al. [51] suggested a pattern-based language modeling system. The patterns, identified as ordered combinations of sentence elements, were extracted with the use of a Brute-Force search-inspired algorithm and used for classification. They reported promising initial findings and further developed the system by adding several data pre-processing techniques [52].

Finally, Ptaszynski et al. [14] proposed a method of using Linguistically-backed preprocessing methods and increased Feature Density to find an optimal way to preprocess the data in order to achieve higher performance, particularly with Convolutional Neural Networks. The experiments performed on actual cyberbullying data showed a major advantage of this approach to all previous methods, including the best performing method so far based on Brute Force Search algorithm.

The later research in cyberbullying detection has mainly concentrated on using recurrent neural networks and pretrained language models with plain tokens to train embeddings [53, 54, 55]. The exceptions being Balakrishnan et al. [56] and Rosa et al. [57], who used psychological features, like personalities, sentiments and emotions to improve automatic cyberbullying detection. However, these were done using simple models. Using linguistic preprocessing and linguistic embeddings to improve classifier performance has not been studied further even though its potential was confirmed earlier [14].

### *2.6. Contributions of This Study*

This research aims to answer the need of developing greener AI by ultimately reducing the $CO_2$ emissions caused by model training by proposing a method to estimate dataset complexity, and thus to comparatively estimate the potential performance of machine learning (ML) classifiers on a particular dataset prior to any training. The problem of constantly increasing needs for more powerful computational resources is affecting the environment due to alarmingly-growing amount of $CO_2$ emissions caused by training of large-scale ML models. Our approach to this problem is to find a way to train classifiers faster and more efficiently by decreasing the training load, which could be possible by looking at some general characteristics of a dataset, in this case, its complexity. This would allow for the reduction of the number of required experiments iterations.

The problem with a meta-learning based approach is that we would need a lot of datasets and features in order to produce a proper model to estimate classifier performance. The example meta dataset used by Gama et al. [3] was considerably small, consisting of around 20 datasets, which makes its statistical reliability very weak as the error rates get high [58, 59, 60]. The later experiments [31] used around 200 datasets, which is still very underwhelming considering the dataset sizes used today, leading the results with meta-learning being modest at best. Similar problems apply to extrapolation as the subset will get too small to represent the whole dataset and the algorithm is unable to thoroughly capture the characteristics and variations associated with each class. This particularly hurts models like CNNs that rely on larger quantities of data [60]. Also, extrapolation loses vocabulary in the field of NLP, which hinders the applicability of results [61].

This research takes a somewhat similar approach as Rahane et al. [15] did for image recognition by directly using a characteristic of the dataset in order to estimate complexity. In the case of our research, this characteristic is Feature Density, based on the notation of Lexical Density [20] that has been used to estimate language complexity. This characteristic was chosen as in previous research [14] where it was shown that there could be a correlation between Feature Density and classifier performance. The research also studies the possibility of preserving linguistic information as parts of used word embeddings in order to improve performance.

The research is conducted by studying the effectiveness of FD using different linguistically-backed feature preprocessing methods in order estimate dataset complexity, which in turn is used to comparatively estimate the potential performance of ML classifiers prior to any training. We hypothesise that FD will show correlations between various preprocessings and results, and by estimating dataset complexity, allows for the reduction of the number of required experiments iterations. This way we can optimize the resource-intensive training of ML models which is becoming a serious issue due to the increases in available dataset sizes and the ever rising popularity of models based on Deep Neural Networks.

Another goal is trying to aid in tackling the issue of cyberbullying, which, being a serious social problem, is also a much more sophisticated problem from the point of view of linguistic representation. The difference in linguistic complexity of datasets allows us to additionally discuss the efficacy of novel linguistically-backed word embeddings. As the recent trends in NLP mostly focus on using words, like with BERT [36], there could be potential in preserving deeper relations between lexical items and structures, by for example including linguistic information like parts-of-speech or dependency information. In order to explore this potential, we propose to preserve the morphological, syntactic and other types of linguistic information by combining them with the pure tokens or lemmas.

There are many classifiers and different ways to produce features that need to be considered when developing harmful and abusive behavior detection methods which is both time consuming and computationally intensive. Also, there are vast amounts of SNS platforms each of which are operating in one or multiple languages. It is practically impossible to develop a one-size-fits-all system for these platforms because of, for example, different user policies, like the definition of harmful content. It is difficult to deal with all languages at once due to the limitations of multilingual models still being widely unknown [62] and machine translation having its own issues with for example language specific semantics.

Our approach concentrates on the classifiers and feature engineering. In practice, we try to estimate what kind of feature engineering methods would work best for different classifiers in different languages. This would allow us to ignore feature engineering methods not viable for a particular classifier or language and only keep those that we believe could yield the highest performance without doing any actual training.

Table 1: Statistics of the applied cyberbullying datasets.

| | English | Japanese | Polish |
|---|---|---|---|
| Number of samples | 12,772 | 2,998 | 11,041 |
| Number of CB samples | 913 | 1,490 | 985 |
| Number of non-CB samples | 11,859 | 1,508 | 10,056 |
| Number of all tokens | 308,939 | 39,283 | 142,811 |
| Number of distinct tokens | 25,106 | 6,947 | 27,444 |
| Avg. length (chars) of a sample | 125.5 | 35.2 | 95.5 |
| Avg. length (words) of a sample | 28.7 | 13.3 | 14.3 |
| Avg. length (chars) of a CB sample | 115.4 | 33.5 | 105.1 |
| Avg. length (words) of a CB sample | 26.8 | 12.6 | 15.0 |
| Avg. length (chars) of a non-CB sample | 126.3 | 37.0 | 94.5 |
| Avg. length (words) of a non-CB sample | 26.8 | 14.1 | 14.2 |

## 3. Applied Datasets

To achieve a thorough and validated analysis, we applied our proposed methods to multiple languages, namely, Japanese, English and Polish. This also allowed us to study the effect of linguistic and cultural differences in classifier performance when the same classification methods are used for different languages. The research uses a total of four datasets, three from the field of automatic cyberbullying detection and one verification set from sentiment analysis. Key statistics of the applied cyberbullying datasets are shown in Table 1.

### *3.1. English Cyberbullying Dataset*

The first dataset for our experiments was the Kaggle Formspring Dataset for Cyberbullying Detection [25]. There was one major problem with the original dataset however, as the original annotations for the data were carried out by untrained laypeople. It has been proven before that the annotations for topics like online harassment and cyberbullying should be done by experts [63]. Therefore, the dataset was re-annotated with the help of experts with sufficient psychological background to assure high quality annotations [24]. In our research we applied the re-annotated version for more accurate results.

Table 1 reports some key statistics of the improved annotation of the dataset. The dataset contains approximately 300 thousand of tokens. There was no visible difference in length between the posted questions and answers, both being approximately 12 words long on average. On the contrary, the harmful (CB) entries were usually slightly but insignificantly shorter compared to the non-harmful (non-CB) samples (approx. 23 vs. 25 words). The amount of harmful samples was also substantially smaller compared to the amount of non-harmful samples, around 7% of the whole dataset, which is approximately the same as the real-life amount of profanity encountered on SNS [63].

### *3.2. Japanese Cyberbullying Dataset*

The Japanese dataset we used for experiments was originally created by Ptaszynski et al. [10], and also widely used by others [47, 51, 52, 13, 14]. It contains 1,490 harmful and 1,508 non-harmful entries in Japanese collected from unofficial school websites and fora. The original data was provided by the Human Rights Research Institute Against All Forms for Discrimination and Racism in Mie Prefecture, Japan. The entries were collected and labeled by Internet Patrol members (expert annotators) with the help of the government supplied manual [9]. The instructions given by the manual are briefly described below.

The definition given by MEXT suggests that cyberbullying occurs when a person is directly offended on the Internet. This includes publication of the person's identity, personal information and other aspects of privacy. Thus, as the first distinguishable features for cyberbullying, MEXT identifies private names (also initials and nicknames), names of organisations and affiliations and private information (address, phone numbers, personal information disclosure, etc.)

In addition, cyberbullying literature reveals vulgarities as one of the most distinguishing characteristics of cyberbullying [6, 64]. Also according to MEXT, vulgar language and cyberbullying can be distinguished from each other as cyberbullying conveys offenses against real individuals. In the prepared dataset, all entries containing at least one of the above characteristics is listed as harmful.

### *3.3. Polish Cyberbullying Dataset*

The Polish dataset originates from PolEval workshop from 2019 [65], collected from Twitter discussions. As feature selection and feature engineering have been proven to be integral parts of cyberbullying detection [14, 66], the tweets are provided as such, without additional preprocessing to allow researchers apply their own preprocessing methods. The only preprocessing applied to the dataset was done only to mask private information, such as personal information of individuals (Twitter users).

The dataset contains 11,041 entries in total, with 10,041 included in the training set and 1000 in the test set. The dataset was initially annotated by laypeople, but was later corrected by an expert in the case of disagreements. The laypeople agreed on majority of the annotations at 91.38%. The number seems very high, but it is mostly due to the fact that the annotators mostly agreed upon non-harmful tweets, which take up most of the dataset at 89.76%. When considering the harmful class, the annotators only agreed upon 1.62% of the entries. Moreover, some of the fully-agreed tweets needed to be corrected to the opposite class in the end by the expert annotator, which shows that using laypeople does not provide accurate enough annotations in the field of cyberbullying. It could be said that layperson annotators can tell with a decent level of confidence that an entry is not harmful (even if it contains some vulgar words), and they can spot, to some extent, if the entry is somehow harmful. Though in most cases they are unable to provide a reasoning for their choice. This provides further proof that for specific problems such as cyberbullying, an expert annotation is required [63]. Comparing the training and test sets, it can be noted that the latter contained a slightly higher ratio of harmful tweets (8.48% for training set vs. 13.40% for test set), which might end up showing in the classifier evaluations.

There was also a reasonably high number of retweets that have slipped into both the data processing phase and the annotation (709 or 6.42%). All of these tweets were not official retweets made using the retweet function, but tweet quotes beginning with a brief ”RT” statement, which differentiates them from normal replies and comments. This needs to be taken into account in the future.

### 3.4. Verification Dataset: Yelp User Reviews Sentiment Dataset

As a dataset for verification of the claimes posed in this study, we applied a subset of Yelp's user reviews data. It contains business reviews about restaurants, shops, etc. from North American metropolises with a rating from one to five stars along with other information. The dataset was originally assembled for Yelp's dataset challenge in order to promote innovation.

The ratings were binarized for our experiment by assigning one star reviews to negative class and five star reviews to positive class. Other reviews were discarded. Also, reviews containing less than three words or more than two standard deviations from the mean were filtered out to avoid over-lengthy review samples differing too much with average sample length in other datasets. We took a random subset of 250,000 positive and 250,000 negative reviews. This way we could study the influence of the change in size of a dataset by a roughly over one magnitude but no more than two magnitudes. With these constraints, we had a large binary dataset of a different topic that is also much simpler than the cyberbullying datasets. Although the whole dataset was much larger and had multiple classes, difference of more than two magnitudes between other datasets and inclusion of more classes might introduce uncontrollable and untraceable variations in dataset statistics influencing the measurements. Therefore, applying these limitations would allow us study our proposed methods with sufficient control. The subset was split into training and test datasets containing 80% and 20% of the data respectively with even number of positive and negative samples in each part.

We also chose the dataset deliberately from a different field in order to verify the general potential of the performance of the proposed method and its universal applicability. The dataset was also considerably larger, more specifically, almost fifty times that of the Polish or English datasets. This allowed us to study how the proposed methods perform with a larger amount of data. The dataset was binarized to provide comparability with the performance metrics between all of the datasets.

## 4. Proposed Methods

### *4.1. Preprocessing and Feature Density*

In order to train the linguistically-backed embeddings, we first preprocessed the dataset in various ways, similarly to previous research [14]. This was done for three reasons. Firstly, to see how traditional classifiers managed the data from similar domain (cyberbullying), but in different languages. Secondly, to later verify the correlation between the classification results and Feature Density (FD) [14]. Finally, to verify the performance of various versions of the proposed linguistically-backed embeddings. Also, because we were trying to make the proposed method entirely systemic and automated, we did not focus on any hand-made features, such as offensive word lexicons, etc., used in previous research [10]. The preprocessing was done using spaCy NLP toolkit [67]. After assembling combinations from the listed preprocessing types, we ended up with a total of 68 possible preprocessing methods for the experiments. All types of preprocessing we applied to generate preprocessing type combinations were listed below. The FDs for all preprocessing types used in this research are shown in Tables 2, 3, 4 and 5.

- **Tokenization:** includes words, punctuation marks, etc. separated by spaces (later: TOK).
- **Lemmatization:** like the above but with generic (dictionary) forms of words ("lemmas") (later: LEM).
- **Parts of speech (separate):** parts of speech information is added in the form of separate features (later: POSS).
- **Parts of speech (combined):** parts of speech information is merged with other applied features (later: POS).
- **Named Entity Recognition (without replacement):** information on what named entities (private name of a person, organization, numericals, etc.) appear in the sentence are added to the applied word (later: NER).
- **Named Entity Recognition (with replacement):** same as above but information replaces the applied word (later: NERR).

- **Dependency structure:** noun- and verb-phrases with syntactic relations between them (later: DEP).
- **Chunking:** like above but without dependency relations ("chunks", later: CHNK).
- **Stopword filtering:** redundant words are filtered out using spaCy's stopword lists (later: STOP)
- **Filtering of non-alphabetics:** non-alphabetic characters are filtered out for English and Polish. For Japanese, *kanji* and *kana* characters are also retained (later: ALPHA)

*4.2. Feature Extraction*

For classifiers other than those based on neural networks, we generated a Bag-of-Features language model was from each of the 68 processed dataset versions, producing a separate model for each of the preprocessing types (Bag-of-Words, Bag-of-Lemmas, Bag-of-POS, etc.). This was done for all of the four datasets. The language models generated from the entries of the datasets were used later in the input layer of classification. We also applied a traditional weight calculation scheme, namely term frequency with inverse document frequency $tf * idf$, where term frequency $tf(t, d)$ refers to raw frequency (number of times a term $t$ (word, token) occurs in a document $d$), and inverse document frequency $idf(t, D)$ is the logarithm of the total number of documents $|D|$ in the corpus divided by the number of documents containing the term $n_t$. Finally, $tf * idf$ refers to term frequency multiplied by inverse document frequency as in equation2.

$$idf(t, D) = log(\frac{|D|}{n_t}) \tag{2}$$

With the Neural Network models, MLP and CNNs, we trained the embeddings as a part of the network using the previously-described preprocessed datasets. Similarly to other classifiers, we trained a separate model for each of the 68 datasets (Word/token Embeddings, Lemmas Embeddings, POS Embeddings, Chunks Embeddings, etc.). The embeddings were fully trained on the datasets themselves as part of the network using using Keras' [68] embedding layer with random initial weights, meaning no pretraining was used.

Table 2: English Cyberbullying Dataset: Feature Density of preprocessing types.

| Preprocessing type | Uniq.1grams | All1grams | FD |
|---|---|---|---|
| POS | 18 | 357616 | .0001 |
| POSALPHA | 18 | 357616 | .0001 |
| POSSTOP | 18 | 194606 | .0001 |
| POSSTOPALPHA | 17 | 129076 | .0001 |
| LEMPOSSALPHA | 17875 | 579664 | .0308 |
| LEMPOSS | 21238 | 660653 | .0321 |
| TOKPOSSALPHA | 21737 | 579624 | .0375 |
| TOKPOSS | 25122 | 660612 | .038 |
| LEMNERALPHA | 14815 | 289868 | .0511 |
| LEMNERR | 17327 | 309124 | .0561 |
| CHNKNERRALPHA | 12293 | 215096 | .0572 |
| LEMNERRALPHA | 17877 | 305481 | .0585 |
| CHNKNERALPHA | 14007 | 228146 | .0614 |
| LEMALPHA | 17860 | 289868 | .0616 |
| LEMPOSSSTOP | 20948 | 334870 | .0626 |
| TOKNERRALPHA | 18595 | 289828 | .0642 |
| CHNKALPHA | 13991 | 215096 | .065 |
| LEMNER | 21239 | 325173 | .0653 |
| LEMPOSSSTOPALPHA | 17554 | 258103 | .068 |
| TOKNERR | 21119 | 309084 | .0683 |
| LEM | 21222 | 308434 | .0688 |
| TOKNERALPHA | 21737 | 305441 | .0712 |
| TOKPOSSSTOP | 24472 | 334869 | .0731 |
| LEMPOS | 26232 | 357657 | .0733 |
| TOKALPHA | 21722 | 289828 | .0749 |
| LEMPOSALPHA | 22206 | 289868 | .0766 |
| TOKNER | 25121 | 325132 | .0773 |
| TOK | 25106 | 308393 | .0814 |
| TOKPOSSSTOPALPHA | 21037 | 258103 | .0815 |
| TOKPOS | 31121 | 357616 | .087 |
| TOKPOSALPHA | 27013 | 289828 | .0932 |
| LEMNERRSTOPALPHA | 14509 | 129076 | .1124 |
| LEMNERRSTOP | 17047 | 146549 | .1163 |
| LEMNERSTOPALPHA | 17557 | 142289 | .1234 |
| CHNKNERR | 33025 | 262529 | .1258 |
| LEMNERRSTOPALPHA | 20950 | 160269 | .1307 |
| LEMPOSSTOP | 25669 | 194674 | .1319 |
| LEMSTOPALPHA | 17540 | 129076 | .1359 |
| TOKNERRSTOPALPHA | 17911 | 129076 | .1387 |
| CHNKNER | 38044 | 272581 | .1396 |
| TOKNERRSTOP | 20480 | 146549 | .1397 |
| LEMSTOP | 20933 | 145866 | .1435 |
| CHNKNERSTOPALPHA | 13356 | 92782 | .144 |
| CHNKNERRSTOPALPHA | 11656 | 80896 | .1441 |
| CHNK | 38029 | 261990 | .1452 |
| TOKNERSTOPALPHA | 21037 | 142289 | .1478 |
| TOKNERSTOP | 24471 | 160268 | .1527 |
| TOKPOSSTOP | 30040 | 194673 | .1543 |
| TOKSTOPALPHA | 21022 | 129076 | .1629 |
| CHNKSTOPALPHA | 13340 | 80896 | .1649 |
| LEMPOSSTOPALPHA | 21626 | 129076 | .1675 |
| TOKSTOP | 24456 | 145865 | .1677 |
| TOKPOSSTOPALPHA | 25925 | 129076 | .2009 |
| CHNKNERRSTOP | 32452 | 126357 | .2568 |
| CHNKNERSTOP | 37462 | 135357 | .2768 |
| CHNKSTOP | 37447 | 125824 | .2976 |
| DEPNERALPHA | 95404 | 240302 | .397 |
| DEPNERRALPHA | 94928 | 215096 | .4413 |
| DEPALPHA | 95386 | 215096 | .4435 |
| DEPNER | 143197 | 321835 | .4449 |
| DEPNERSTOPALPHA | 47159 | 104940 | .4494 |
| DEPNERR | 141479 | 308704 | .4583 |
| DEP | 143179 | 308704 | .4638 |
| DEPNERSTOP | 94539 | 184130 | .5134 |
| DEPNERRSTOP | 92730 | 172086 | .5389 |
| DEPSTOP | 94521 | 172086 | .5493 |
| DEPNERRSTOPALPHA | 46552 | 80896 | .5755 |
| DEPSTOPALPHA | 47141 | 80896 | .5827 |

Table 3: Japanese Cyberbullying Dataset: Feature Density of preprocessing types.

| Preprocessing type | Uniq.1grams | All1grams | FD |
|---|---|---|---|
| POS | 19 | 40015 | 0.0005 |
| POSALPHA | 19 | 40015 | 0.0005 |
| POSSTOP | 19 | 25777 | 0.0007 |
| POSSTOPALPHA | 16 | 18444 | 0.0009 |
| LEMPOSS | 6495 | 78685 | 0.0825 |
| TOKPOSS | 6964 | 79226 | 0.0879 |
| LEMPOSSALPHA | 6152 | 65314 | 0.0942 |
| TOKPOSSALPHA | 6579 | 65314 | 0.1007 |
| LEMPOSSSTOP | 6392 | 50203 | 0.1273 |
| LEMNERR | 5065 | 39024 | 0.1298 |
| TOKPOSSSTOP | 6818 | 50742 | 0.1344 |
| TOKNERR | 5474 | 39380 | 0.1390 |
| LEMNER | 6607 | 44046 | 0.1500 |
| LEMNERRALPHA | 4918 | 32665 | 0.1506 |
| TOKNER | 7078 | 44509 | 0.1590 |
| TOKNERRALPHA | 5329 | 32665 | 0.1631 |
| LEMPOSSSTOPALPHA | 6049 | 36846 | 0.1642 |
| LEM | 6478 | 38955 | 0.1663 |
| LEMNERALPHA | 6266 | 36835 | 0.1701 |
| LEMPOS | 6870 | 40017 | 0.1717 |
| TOKPOSSSTOPALPHA | 6433 | 36846 | 0.1746 |
| TOK | 6947 | 39283 | 0.1768 |
| CHNKNERR | 5911 | 32935 | 0.1795 |
| TOKNERALPHA | 6694 | 36835 | 0.1817 |
| TOKPOS | 7505 | 40018 | 0.1875 |
| LEMALPHA | 6138 | 32665 | 0.1879 |
| CHNKNERRALPHA | 5289 | 26694 | 0.1981 |
| LEMPOSALPHA | 6520 | 32665 | 0.1996 |
| LEMNERRSTOP | 4961 | 24791 | 0.2001 |
| TOKALPHA | 6564 | 32665 | 0.2009 |
| TOKNERRSTOP | 5328 | 25143 | 0.2119 |
| CHNKNER | 7797 | 36061 | 0.2162 |
| TOKPOSALPHA | 7109 | 32665 | 0.2176 |
| LEMNERSTOP | 6504 | 29641 | 0.2194 |
| CHNKNERALPHA | 6659 | 29873 | 0.2229 |
| TOKNERSTOP | 6932 | 30102 | 0.2303 |
| CHNK | 7672 | 32877 | 0.2334 |
| CHNKALPHA | 6534 | 26694 | 0.2448 |
| LEMSTOP | 6375 | 24722 | 0.2579 |
| LEMPOSSTOP | 6700 | 25779 | 0.2599 |
| LEMNERRSTOPALPHA | 4814 | 18444 | 0.2610 |
| TOKSTOP | 6801 | 25046 | 0.2715 |
| LEMNERSTOPALPHA | 6163 | 22444 | 0.2746 |
| TOKPOSSTOP | 7237 | 25780 | 0.2807 |
| TOKNERRSTOPALPHA | 5183 | 18444 | 0.2810 |
| CHNKNERRSTOP | 5774 | 20190 | 0.2860 |
| TOKNERSTOPALPHA | 6548 | 22444 | 0.2917 |
| LEMSTOPALPHA | 6035 | 18444 | 0.3272 |
| CHNKNERSTOP | 7659 | 23283 | 0.3290 |
| LEMPOSSTOPALPHA | 6350 | 18444 | 0.3443 |
| TOKSTOPALPHA | 6418 | 18444 | 0.3480 |
| CHNKNERRSTOPALPHA | 5152 | 14003 | 0.3679 |
| TOKPOSSTOPALPHA | 6841 | 18444 | 0.3709 |
| CHNKSTOP | 7534 | 20132 | 0.3742 |
| CHNKNERSTOPALPHA | 6521 | 17132 | 0.3806 |
| CHNKSTOPALPHA | 6396 | 14003 | 0.4568 |
| DEPNERSTOPALPHA | 12078 | 19727 | 0.6123 |
| DEPNERALPHA | 21542 | 32541 | 0.6620 |
| DEPNERSTOP | 16800 | 23802 | 0.7058 |
| DEPNER | 26264 | 36581 | 0.7180 |
| DEPNERR | 26089 | 33357 | 0.7821 |
| DEP | 26139 | 33357 | 0.7836 |
| DEPNERRALPHA | 21354 | 26694 | 0.8000 |
| DEPALPHA | 21417 | 26694 | 0.8023 |
| DEPNERRSTOP | 16619 | 20611 | 0.8063 |
| DEPSTOP | 16675 | 20611 | 0.8090 |
| DEPNERRSTOPALPHA | 11884 | 14003 | 0.8487 |
| DEPSTOPALPHA | 11953 | 14003 | 0.8536 |

Table 4: Polish Cyberbullying Dataset: Feature Density of preprocessing types.

| Preprocessing type | Uniq. 1grams | All 1grams | FD |
|---|---|---|---|
| POS | 15 | 157137 | 0.0001 |
| POSALPHA | 15 | 157137 | 0.0001 |
| POSSTOP | 15 | 104715 | 0.0001 |
| POSSTOPALPHA | 15 | 63004 | 0.0002 |
| LEMPOSS | 16403 | 294706 | 0.0557 |
| LEMPOSSALPHA | 14952 | 230795 | 0.0648 |
| LEMPOSSSTOP | 16246 | 189900 | 0.0856 |
| TOKPOSS | 27458 | 294717 | 0.0932 |
| LEMNERR | 14509 | 142929 | 0.1015 |
| LEMPOS | 17217 | 157137 | 0.1096 |
| LEMNER | 16393 | 147531 | 0.1111 |
| TOKPOSSALPHA | 26067 | 230795 | 0.1129 |
| LEMNERRALPHA | 13167 | 115398 | 0.1141 |
| LEM | 16388 | 142800 | 0.1148 |
| LEMPOSSSTOPALPHA | 14756 | 125979 | 0.1171 |
| LEMNERALPHA | 14943 | 121392 | 0.1231 |
| LEMALPHA | 14938 | 115398 | 0.1294 |
| LEMPOSALPHA | 15479 | 115398 | 0.1341 |
| TOKPOSSSTOP | 26839 | 189911 | 0.1413 |
| LEMNERRSTOP | 14350 | 90569 | 0.1584 |
| LEMPOSSTOP | 16923 | 104722 | 0.1616 |
| LEMNERSTOP | 16236 | 95020 | 0.1709 |
| TOKNERR | 24922 | 142940 | 0.1744 |
| CHNKNERR | 24922 | 142940 | 0.1744 |
| LEMSTOP | 16231 | 90441 | 0.1795 |
| TOKPOS | 28521 | 157137 | 0.1815 |
| TOKNER | 27450 | 147542 | 0.1860 |
| CHNKNER | 27450 | 147542 | 0.1860 |
| TOK | 27444 | 142811 | 0.1922 |
| CHNK | 27444 | 142811 | 0.1922 |
| TOKPOSSSTOPALPHA | 25379 | 125979 | 0.2015 |
| TOKNERRALPHA | 23630 | 115398 | 0.2048 |
| CHNKNERRALPHA | 23630 | 115398 | 0.2048 |
| LEMNERRSTOPALPHA | 12972 | 63004 | 0.2059 |
| CHNKNERALPHA | 26060 | 121997 | 0.2136 |
| LEMNERSTOPALPHA | 14747 | 68845 | 0.2142 |
| TOKNERALPHA | 26060 | 121392 | 0.2147 |
| TOKALPHA | 26054 | 115398 | 0.2258 |
| CHNKALPHA | 26054 | 115398 | 0.2258 |
| TOKPOSALPHA | 26784 | 115398 | 0.2321 |
| LEMSTOPALPHA | 14742 | 63004 | 0.2340 |
| LEMPOSSTOPALPHA | 15173 | 63004 | 0.2408 |
| TOKPOSSTOP | 27683 | 104722 | 0.2643 |
| TOKNERRSTOP | 24303 | 90580 | 0.2683 |
| CHNKNERRSTOP | 24303 | 90580 | 0.2683 |
| TOKNERSTOP | 26831 | 95031 | 0.2823 |
| CHNKNERSTOP | 26831 | 95031 | 0.2823 |
| TOKSTOP | 26825 | 90452 | 0.2966 |
| CHNKSTOP | 26825 | 90452 | 0.2966 |
| TOKNERRSTOPALPHA | 22946 | 63004 | 0.3642 |
| CHNKNERRSTOPALPHA | 22946 | 63004 | 0.3642 |
| CHNKNERSTOPALPHA | 25372 | 69449 | 0.3653 |
| TOKNERSTOPALPHA | 25372 | 68845 | 0.3685 |
| TOKSTOPALPHA | 25366 | 63004 | 0.4026 |
| CHNKSTOPALPHA | 25366 | 63004 | 0.4026 |
| TOKPOSSTOPALPHA | 25930 | 63004 | 0.4116 |
| DEPNERSTOP | 68279 | 111161 | 0.6142 |
| DEPNER | 102460 | 163736 | 0.6258 |
| DEPNERRSTOP | 67378 | 104715 | 0.6434 |
| DEPNERR | 101554 | 157137 | 0.6463 |
| DEPNERSTOPALPHA | 51860 | 80173 | 0.6469 |
| DEPNERALPHA | 86044 | 132723 | 0.6483 |
| DEPSTOP | 68273 | 104715 | 0.6520 |
| DEP | 102454 | 157137 | 0.6520 |
| DEPNERRALPHA | 85138 | 115398 | 0.7378 |
| DEPALPHA | 86038 | 115398 | 0.7456 |
| DEPNERRSTOPALPHA | 50959 | 63004 | 0.8088 |
| DEPSTOPALPHA | 51854 | 63004 | 0.8230 |

Table 5: Verification Dataset (English Yelp Reviews): Feature Density of preprocessing types.

| Preprocessing type | Uniq.1grams | All1grams | FD |
|---|---|---|---|
| POS | 18 | 22123101 | 0.0000 |
| POSALPHA | 18 | 22123101 | 0.0000 |
| POSSTOP | 17 | 10976927 | 0.0000 |
| POSSTOPALPHA | 17 | 7913333 | 0.0000 |
| LEMPOSSALPHA | 108272 | 37473399 | 0.0029 |
| LEMPOSS | 124750 | 41602171 | 0.0030 |
| TOKPOSSALPHA | 125796 | 37473386 | 0.0034 |
| TOKPOSS | 142140 | 41602328 | 0.0034 |
| LEMNERALPHA | 81750 | 18736707 | 0.0044 |
| LEMNERR | 92299 | 19691797 | 0.0047 |
| TOKNERRALPHA | 100174 | 18736694 | 0.0053 |
| CHNKNERALPHA | 67433 | 12588242 | 0.0054 |
| LEMNERRALPHA | 108270 | 19681446 | 0.0055 |
| TOKNERR | 110454 | 19691940 | 0.0056 |
| LEMALPHA | 108260 | 18736707 | 0.0058 |
| LEMNER | 124748 | 20691409 | 0.0060 |
| CHNKNERALPHA | 82630 | 13295729 | 0.0062 |
| LEM | 124828 | 19686486 | 0.0063 |
| TOKNERALPHA | 125792 | 19681433 | 0.0064 |
| LEMPOSSSTOP | 124382 | 19312556 | 0.0064 |
| CHNKALPHA | 82619 | 12588242 | 0.0066 |
| TOKALPHA | 125783 | 18736694 | 0.0067 |
| LEMPOSSSTOPALPHA | 107695 | 15826663 | 0.0068 |
| TOKNER | 142136 | 20691566 | 0.0069 |
| LEMPOS | 157276 | 22126373 | 0.0071 |
| TOK | 142145 | 19686629 | 0.0072 |
| LEMPOSALPHA | 136041 | 18736707 | 0.0073 |
| TOKPOSSSTOP | 141540 | 19312673 | 0.0073 |
| TOKPOSSSTOPALPHA | 124822 | 15826663 | 0.0079 |
| TOKPOS | 191314 | 22126347 | 0.0086 |
| TOKPOSALPHA | 169800 | 18736694 | 0.0091 |
| LEMNERRSTOPALPHA | 81202 | 7913333 | 0.0103 |
| LEMNERRSTOP | 91950 | 8548830 | 0.0108 |
| LEMNERSTOPALPHA | 107692 | 8613858 | 0.0125 |
| TOKNERRSTOPALPHA | 99226 | 7913333 | 0.0125 |
| TOKNERSTOP | 109868 | 8548956 | 0.0129 |
| LEMNERSTOP | 124379 | 9302597 | 0.0134 |
| LEMSTOPALPHA | 107682 | 7913333 | 0.0136 |
| LEMPOSSTOP | 156259 | 10980340 | 0.0142 |
| TOKNERSTOPALPHA | 124817 | 8613858 | 0.0145 |
| LEMSTOP | 124369 | 8517231 | 0.0146 |
| TOKNERSTOP | 141535 | 9302714 | 0.0152 |
| TOKSTOPALPHA | 124808 | 7913333 | 0.0158 |
| CHNKNERRSTOPALPHA | 66524 | 4096850 | 0.0162 |
| TOKSTOP | 141526 | 8517348 | 0.0166 |
| LEMPOSSTOPALPHA | 134752 | 7913333 | 0.0170 |
| TOKPOSSTOP | 189338 | 10980327 | 0.0172 |
| CHNKNERSTOPALPHA | 81699 | 4699728 | 0.0174 |
| CHNKSTOPALPHA | 81688 | 4096850 | 0.0199 |
| TOKPOSSTOPALPHA | 167504 | 7913333 | 0.0212 |
| CHNKNERR | 764785 | 15908671 | 0.0481 |
| CHNKNER | 861736 | 16452253 | 0.0524 |
| CHNK | 861725 | 15890287 | 0.0542 |
| CHNKNERRSTOP | 764246 | 7115714 | 0.1074 |
| CHNKNERSTOP | 861185 | 7556935 | 0.1140 |
| CHNKSTOP | 861174 | 7097343 | 0.1213 |
| DEPNERALPHA | 2240284 | 13497593 | 0.1660 |
| DEPNERRALPHA | 2184827 | 12588242 | 0.1736 |
| DEPALPHA | 2240266 | 12588242 | 0.1780 |
| DEPNERR | 3758037 | 18251727 | 0.2059 |
| DEPNER | 3933231 | 18959225 | 0.2075 |
| DEP | 3933213 | 18251727 | 0.2155 |
| DEPNERSTOPALPHA | 1300041 | 4901505 | 0.2652 |
| DEPNERSTOP | 2972316 | 10059025 | 0.2955 |
| DEPNERRSTOP | 2795319 | 9456061 | 0.2956 |
| DEPNERRSTOPALPHA | 1241849 | 4096850 | 0.3031 |
| DEPSTOP | 2972298 | 9456061 | 0.3143 |
| DEPSTOPALPHA | 1300023 | 4096850 | 0.3173 |

*4.3. Classification*

In the experiment we applied the following classification algorithms. The assumption was that each classifier presents different correlation with FD, and characteristics of this correlation can be further exploited to minimize the time required for training optimal solutions by choosing a small number of feature sets which usually perform best, or at least eliminating feature sets which always perform below an acceptable performance threshold.

**SVM** or Support-vector machines [69] are a set of classifiers well established in AI and NLP. They represent data, belonging to specified categories, as points in space (vectors), and find an optimal hyperplane to separate the examples from each category. SVM has been very successful in previous cyberbullying research [10, 12, 14]. In this research, we used two SVM functions, the **linear SVM**, as it had the greatest performance out of all of the SVM kernel functions that were used in previous research [14] and a linear SVM function supported with Stochastic Gradient Descent optimizer (**SGD**).

**NaïveBayes** (NB) classifier is a supervised learning algorithm applying Bayes' theorem that has a strong (naïve) assumption of independence between pairs of features. It is traditionally used as a baseline in different text classification tasks and is known for working well with smaller datasets. Also, it is considerably fast to train compared to some other popular classifiers e.g. Random Forest.

**kNN** or the k-Nearest Neighbors classifier takes as input k-closest training samples with assigned classes and classifies the input sample by a majority vote. It is often applied as a baseline alongside Naïve Bayes. The classifier is fast and simple to train but on the contrary, it is very susceptible to outliers and overfitting. In this research, we used k=1 setting in which the input sample is simply assigned to the class of the first nearest neighbor.

**Random Forest** (RF) in training phase creates multiple decision trees to output the optimal class (mode of classes) in classification phase [70]. An improvement of RF when comparing to standard decision trees is the ability to correct overfitting to the training set, which is very common in decision trees [71]. In practice, Random Forest starts by taking a random bootstrap sample with replacement from the dataset. It then selects a random subset of features in order to reduce the dimensionality of the sample.

Next, an unpruned decision tree is trained on this bootstrap sample. This process is repeated for the desired ensemble size. The predicted value of an unknown instance is obtained by taking a majority vote over the entire ensemble of trees [70].

**Logistic Regression** (LR) is a statistical model that calculates class probabilities using a logistic function (sigmoid) instead of a straight line (linear regression) or a hyperplane. Logistic regression models are usually fit using Maximum Likelihood Estimation [71]. The model assigns a probability value between [0,1] for each input, which is used to determine the class it belongs. The experiments in this research are conducted using two different solvers, **Newton's method** and **l-bfgs**, which is a quasi-Newton method that uses approximations and memory saving features in order to improve performance with the cost of possible minor convergence problems.

**Boosting** includes algorithms such as **AdaBoost** [72] and Extreme Gradient Boosting (**XGBoost**) [73], which is a more generalized and optimized version. In boosting, the weak learners, which are usually decision trees, evolve over time as they are trained sequentially to perform better on the residuals of the previous learner. The members cast a weighted vote instead of generating random predictors (RF) and averaging their result [71].

**MLP** (Multilayer Perceptron) is a type of feed-forward artificial neural network consisting of an input layer, an output layer and one or more hidden layers. In this experiment MLP refers to a neural network using regular dense layers. We applied an MLP implementation with Rectified Linear Units (ReLU) as a neuron activation function [74] and one hidden layer with dropout regularization to reduce overfitting and improve generalization by randomly dropping out some of the hidden neurons during training [74].

**CNN** or Convolutional Neural Networks are a type of feed-forward artificial neural network utilizing convolutional and pooling layers. Although originally designed for image recognition, the effectiveness of CNNs has been shown in multiple other tasks, including NLP [75] and sentence classification [76]. We implemented CNNs with Rectified Linear Units (ReLU) as a neuron activation function, and max pooling [77], which applies a max filter to non-overlying sub-parts of the input to reduce dimensionality and as a result, helps to prevent overfitting. We also applied dropout regularization on penultimate layer for the same reason. We applied two versions of CNNs. First, with only one hidden convolutional layer containing 128 units. The second network consisted of two hidden convolutional layers with 128 feature maps each, 4x4 size of patch and 2x2 max-pooling. We used Adaptive Moment Estimation (Adam), a variant of Stochastic Gradient Descent [78] as the optimization function.

## 5. Experiments

### *5.1. Setup*

The four preprocessed datasets were additionally preprocessed according to methodology described in Section 4.1, which resulted in the creation of 68 separate training sets for each original dataset. The experiment was performed once for every preprocessing type for every dataset. Each of the classifiers (sect. 4.3) were tested on all of the versions of the datasets in a 10-fold cross validation procedure for the English and Japanese Cyberbullying datasets and with predetermined train-test set splits for the Polish and verification datasets, to retain the originally proposed method of evaluation for each dataset. This gave us an opportunity to evaluate how effective different preprocessing methods were for each classifier and for each language, also with comparison to previous methods. As some of the datasets were not balanced, we oversampled the minority class using Synthetic Minority Over-sampling Technique (SMOTE) [79] in order to balance out the classes. The preprocessing methods represent a wide range of Feature Densities, which can be used to evaluate the correlation with classifier performance. This gave us nearly eighteen thousand experiment runs[4] which we use as a basis for our discussions on the results and applicability of FD. The hardware used for the experiments included Intel i9 7920X, running at stock 2.90 GHz, for non-neural classifiers and Nvidia GTX 1080ti for neural classifiers. The power consumptions were calculated expecting 100% power usage for the device in question.

Table 6: Runtimes and approximate power usage of the training processes. Non-neural classifiers: Intel i9 7920X@2.90 GHz, 163W. Neural classifiers: Nvidia GTX 1080ti, 250W. Expecting 100% power usage.

| | English | | Japanese | |
|---|---|---|---|---|
| **Classifier** | **Runtime (s)** | **Power usage (Wh)** | **Runtime (s)** | **Power usage (Wh)** |
| **Logistic Regression** | 321.6 | 145.61 | 82.75 | 37.47 |
| **CGD Logistic Regression** | 249.74 | 113.08 | 76.74 | 34.75 |
| **SGD SVM** | 176.26 | 79.81 | 11.35 | 5.14 |
| **Linear SVM** | 1543.06 | 698.67 | 41.41 | 18.75 |
| **KNN** | 556.44 | 251.94 | 43.22 | 19.57 |
| **Naive Bayes** | 97.54 | 44.16 | 36.36 | 16.46 |
| **Random Forest** | 3982.49 | 1803.18 | 420.19 | 190.25 |
| **AdaBoost** | 10425.4 | 4720.39 | 611.21 | 276.74 |
| **XGBoost** | 17917.74 | 8112.76 | 47093.99 | 21323.11 |
| **MLP** | 53845.89 | 37392.98 | 18235.85 | 12663.78 |
| **CNN1** | 62361.45 | 43306.56 | 21288.6 | 14783.75 |
| **CNN2** | 62054.46 | 43093.37 | 16116.72 | 11192.17 |
| | **Polish** | | **Yelp** | |
| **Classifier** | **Runtime (s)** | **Power usage (Wh)** | **Runtime (s)** | **Power usage (Wh)** |
| **Logistic Regression** | 56.66 | 25.65 | 3429.78 | 1552.93 |
| **CGD Logistic Regression** | 45.42 | 20.56 | 3592.51 | 1626.61 |
| **SGD SVM** | 160.06 | 72.47 | 2876.84 | 1302.57 |
| **Linear SVM** | 36.82 | 16.67 | 28365.17 | 12843.12 |
| **KNN** | 56.75 | 25.69 | 2687.75 | 1216.95 |
| **Naive Bayes** | 33.77 | 15.29 | 35976.27 | 16289.26 |
| **Random Forest** | 118.39 | 53.61 | 5420.16 | 2454.13 |
| **AdaBoost** | 136.26 | 61.7 | 3429.78 | 1552.93 |
| **XGBoost** | 2042.64 | 924.86 | 3592.51 | 1626.61 |
| **MLP** | 7095.01 | 4927.09 | 121782.6 | 84571.25 |
| **CNN1** | 9939.68 | 6902.56 | 144275.23 | 100191.13 |
| **CNN2** | 9418.16 | 6540.39 | 156572.35 | 108730.8 |

### *5.2. Effect of Feature Density*

#### *5.2.1. English Cyberbullying Dataset*

We trained each of the classifiers using the proposed preprocessing methods. The classification results are presented in Table 7. As the results for using only parts-of-speech tags, which had the lowest FD by far, were extremely low (close to a coinflip). Thus, we can say that POS tags alone do not contain enough information to successfully classify the entries. We also analyzed the correlation between Feature Density and classifier F-score, which is shown in Table 8.

After excluding the preprocessing methods that only used POS tags, we can see that all classifiers, except CNNs have a strong negative correlation with Feature Density. So these classifiers seem to have a weaker performance if a lot of linguistic information is added, and the best results being usually within the range of .05 to .15 of FD depending on the classifier. This range includes 38 of the 68 preprocessing methods (Table 2). The sweet spot for performance can be seen from, for example, the highest performing classifier, SVM with SGD optimizer (Figure 1), where the maximum classifier performance starts high at around .05 of FD and slowly falls until .14 after which there is a noticeable drop. The performance only falls further as the FD rises. If the weaker feature sets were to be left out, the power savings are approximately 35Wh calculated from Table 6 for training the SGD SVM classifier, which is not very much. But the classifier was very power efficient to train to begin.

[4]2 datasets x 10 fold x 68 preprocessings x 12 classifiers (16,320 runs) + 2 datasets x 1 train-test x 68 preprocessings x 12 classifiers (1,632) = 17,952 experiment runs.

For CNNs however, there was a very weak positive or no correlation between FD and the classifier performance, with the higher FD datasets performing equally or even slightly better when comparing to the low FD datasets. Taking a look at one layer CNN's performance, which was better than the CNN with two layers, we can see from Figure 1 that the maximum performance starts at a moderate level and stays stable throughout the whole range of feature densities. The best results can be found between .05 to .1 and after around .45 FD. From Table 2 we can see that around half of the preprocessing types fall into these ranges. We can calculate from Table 6 that the power savings for the CNN are approximately 21kWh, which is huge compared to the SVM's power savings.

The results suggest that for non-CNN classifiers there is no need to consider preprocessings with a high FD, such as chunking or dependencies, as they had a considerably lower performance. The performance seems to start falling rapidly at around .15 FD with most of the classifiers. For CNNs, as there was almost no correlation between FD and F-score, we are unable to estimate an ideal range for Feature Density. However, this means that there is a potential in the higher FD preprocessing types, namely, dependencies for CNNs.

The reason for CNNs relatively low performance could be explained by the relatively small size of the dataset, especially when considering the amount of actual cyberbullying entries, as adding even a second layer to the network already caused a loss of the most valuable features and ended up degrading the performance. With such small amount of data, it does not seem useful to train deep learning models to solve the classification problem. Still, the dependency based features are showing some potential with CNNs.

With a considerably larger dataset it could be possible to outperform other classifiers and especially the traditional approach to word embeddings, namely, using plain tokens, and supplementing them with dependency based features when using deep learning, as was previously proposed by [40].

The experiments show that changing Feature Density in moderate amount can yield good results when using other classifiers than CNNs. However, excessive changes to either too low or too high always showed diminishing results. The threshold was in all cases approximately between 50% and 200% of the original density (TOK), most optimal FDs only slightly varying with each classifier. The exception being Random Forest, which showed a clear spike at around .12 FD. As the usage of high Feature Density datasets showed potential with CNNs, their usage needs to be confirmed with a larger dataset. Also, more exact ideal feature densities need to be confirmed for each classifier using datasets of different sizes and fields to make as accurate ranking of classifiers by FD as possible.

*5.2.2. Japanese Cyberbullying Dataset*

Similarly to the English dataset, we trained all of the classifiers and analyzed the correlation of Feature Density with each of the classifiers using the proposed preprocessing methods. All classification results are represented in Table 9 and correlations in Table 10 We also excluded preprocessing types that only use POS-tags for the same reason as with the English dataset.

From Table 10 we can see that all classifiers have a strong negative correlation with Feature Density, meaning that adding too much linguistic information, such as dependency relations, push the score down. The best results are usually within the range of .08 to .30 FD depending on the classifier. This applies to, for example, the highest performing classifier, one-layer CNN (Figure 2), where the maximum classifier performance starts high at around .08 of FD and slightly increases until .30 after which there is a noticeable drop. The performance only falls further as the FD rises, stabilizing after .40 FD. This range includes 43 of the 68 preprocessing methods (Table 3). If the weaker feature sets were to be left out, the power savings are approximately 5.4kWh for the one-layer CNN, when calculated from Table 6.

Table 7: English Dataset: F1 for all preprocessing types & classifiers; best classifier for each dataset in **bold**; best preprocessing type for each underlined

| | LBFGS LR | Newton LR | Linear SVM | SGD SVM | KNN | NaiveBayes | RandomForest | AdaBoost | XGBoost | MLP | CNN1 | CNN2 |
|---|---|---|---|---|---|---|---|---|---|---|---|---|
| CHNK | 0.727 | 0.726 | 0.718 | **0.736** | 0.57 | 0.674 | 0.613 | 0.649 | 0.667 | 0.724 | 0.657 | 0.666 |
| CHNKNERR | 0.688 | 0.695 | 0.702 | 0.699 | 0.58 | 0.653 | 0.603 | 0.608 | 0.642 | **0.704** | 0.645 | 0.662 |
| CHNKNERRALPHA | 0.66 | 0.663 | 0.651 | 0.657 | 0.603 | 0.626 | 0.616 | 0.599 | 0.653 | **0.674** | 0.566 | 0.6 |
| CHNKNERRSTOP | 0.686 | 0.684 | 0.684 | **0.694** | 0.577 | 0.629 | 0.635 | 0.621 | 0.652 | 0.693 | 0.402 | 0.344 |
| CHNKNERRSTOPALPHA | 0.618 | 0.617 | 0.591 | 0.607 | 0.404 | 0.598 | 0.62 | 0.582 | **0.648** | 0.623 | 0.451 | 0.34 |
| CHNKNER | 0.718 | 0.723 | 0.721 | **0.737** | 0.582 | 0.669 | 0.603 | 0.63 | 0.673 | 0.722 | 0.654 | 0.642 |
| CHNKNERALPHA | 0.675 | 0.676 | 0.663 | 0.663 | 0.599 | 0.641 | 0.618 | 0.609 | 0.649 | **0.684** | 0.557 | 0.614 |
| CHNKNERSTOP | **0.724** | **0.724** | 0.715 | **0.724** | 0.582 | 0.663 | 0.635 | 0.652 | 0.679 | 0.72 | 0.501 | 0.298 |
| CHNKNERSTOPALPHA | 0.666 | 0.661 | 0.644 | **0.668** | 0.386 | 0.615 | 0.659 | 0.625 | 0.656 | 0.647 | 0.431 | 0.406 |
| CHNKALPHA | 0.684 | 0.681 | 0.669 | 0.683 | 0.607 | 0.643 | 0.647 | 0.616 | 0.676 | **0.695** | 0.587 | 0.583 |
| CHNKSTOP | 0.722 | 0.721 | 0.711 | **0.723** | 0.577 | 0.67 | 0.667 | 0.648 | 0.679 | 0.715 | 0.386 | 0.342 |
| CHNKSTOPALPHA | 0.629 | 0.637 | 0.606 | 0.619 | 0.395 | 0.608 | 0.649 | 0.654 | **0.664** | 0.628 | 0.455 | 0.374 |
| DEP | 0.617 | 0.619 | 0.568 | 0.587 | 0.243 | 0.617 | 0.536 | 0.566 | 0.598 | 0.594 | 0.682 | **0.694** |
| DEPNERR | 0.61 | 0.614 | 0.571 | 0.587 | 0.241 | 0.611 | 0.533 | 0.562 | 0.596 | 0.595 | 0.67 | **0.695** |
| DEPNERRALPHA | 0.606 | 0.605 | 0.589 | 0.602 | 0.312 | 0.596 | 0.537 | 0.556 | 0.595 | 0.593 | 0.585 | **0.622** |
| DEPNERRSTOP | 0.602 | 0.599 | 0.564 | 0.568 | 0.273 | 0.615 | 0.543 | 0.572 | 0.6 | 0.578 | **0.726** | 0.702 |
| DEPNERRSTOPALPHA | 0.584 | 0.584 | 0.56 | 0.581 | 0.386 | 0.599 | 0.544 | 0.561 | 0.595 | 0.574 | 0.583 | **0.619** |
| DEPNER | 0.624 | 0.621 | 0.574 | 0.585 | 0.242 | 0.611 | 0.528 | 0.564 | 0.595 | 0.592 | 0.686 | **0.692** |
| DEPNERALPHA | 0.585 | 0.589 | 0.561 | 0.579 | 0.213 | 0.607 | 0.578 | 0.497 | 0.593 | 0.603 | 0.606 | **0.623** |
| DEPNERSTOP | 0.611 | 0.602 | 0.564 | 0.576 | 0.274 | 0.604 | 0.527 | 0.563 | 0.604 | 0.577 | **0.725** | 0.708 |
| DEPNERSTOPALPHA | 0.535 | 0.531 | 0.523 | 0.523 | 0.297 | 0.543 | 0.563 | 0.422 | 0.576 | 0.564 | 0.63 | **0.632** |
| DEPALPHA | 0.609 | 0.612 | 0.588 | 0.601 | 0.314 | 0.6 | 0.545 | 0.552 | 0.604 | 0.598 | 0.606 | **0.62** |
| DEPSTOP | 0.606 | 0.595 | 0.562 | 0.571 | 0.276 | 0.616 | 0.544 | 0.576 | 0.603 | 0.584 | **0.741** | 0.648 |
| DEPSTOPALPHA | 0.586 | 0.587 | 0.564 | 0.588 | 0.388 | 0.594 | 0.539 | 0.568 | 0.595 | 0.578 | **0.629** | 0.625 |
| LEM | 0.781 | 0.786 | 0.784 | **0.79** | 0.634 | 0.715 | 0.724 | 0.72 | 0.744 | 0.786 | 0.67 | 0.665 |
| LEMNERR | 0.74 | 0.737 | 0.742 | 0.74 | 0.601 | 0.692 | 0.697 | 0.683 | 0.724 | **0.749** | 0.658 | 0.663 |
| LEMNERRALPHA | 0.729 | 0.728 | 0.725 | 0.725 | 0.614 | 0.685 | 0.699 | 0.68 | 0.71 | **0.74** | 0.645 | 0.652 |
| LEMNERRSTOP | 0.737 | 0.734 | 0.726 | 0.732 | 0.609 | 0.682 | 0.727 | 0.69 | 0.72 | **0.741** | 0.371 | 0.364 |
| LEMNERRSTOPALPHA | 0.732 | 0.732 | 0.714 | 0.727 | 0.624 | 0.674 | 0.723 | 0.682 | 0.704 | **0.737** | 0.372 | 0.348 |
| LEMPOSS | 0.764 | 0.765 | 0.769 | 0.767 | 0.564 | 0.713 | 0.658 | 0.679 | 0.717 | **0.773** | 0.662 | 0.736 |
| LEMPOSSALPHA | **0.76** | 0.758 | 0.753 | 0.758 | 0.406 | 0.705 | 0.669 | 0.674 | 0.712 | 0.756 | 0.603 | 0.715 |
| LEMPOSSSTOP | 0.763 | 0.766 | 0.767 | **0.774** | 0.566 | 0.709 | 0.706 | 0.691 | 0.72 | 0.773 | 0.683 | 0.725 |
| LEMPOSSSTOPALPHA | 0.762 | **0.766** | 0.748 | 0.765 | 0.49 | 0.702 | 0.713 | 0.681 | 0.714 | 0.757 | 0.593 | 0.716 |
| LEMNER | 0.784 | 0.782 | 0.787 | **0.792** | 0.631 | 0.71 | 0.716 | 0.72 | 0.742 | 0.78 | 0.68 | 0.613 |
| LEMNERALPHA | 0.763 | 0.764 | 0.765 | 0.767 | 0.637 | 0.699 | 0.71 | 0.707 | 0.742 | **0.768** | 0.662 | 0.671 |
| LEMNERSTOP | 0.782 | 0.783 | 0.782 | **0.792** | 0.634 | 0.706 | 0.745 | 0.725 | 0.742 | 0.78 | 0.429 | 0.378 |
| LEMNERSTOPALPHA | **0.77** | 0.767 | 0.752 | 0.767 | 0.64 | 0.693 | 0.739 | 0.716 | 0.738 | 0.768 | 0.46 | 0.414 |
| LEMPOS | 0.778 | 0.778 | 0.788 | **0.79** | 0.517 | 0.711 | 0.663 | 0.727 | 0.741 | 0.783 | 0.665 | 0.64 |
| LEMPOSALPHA | 0.768 | 0.772 | 0.772 | 0.768 | 0.522 | 0.7 | 0.654 | 0.713 | 0.727 | **0.775** | 0.664 | 0.695 |
| LEMPOSSTOP | 0.78 | 0.781 | **0.788** | **0.788** | 0.642 | 0.708 | 0.708 | 0.721 | 0.735 | 0.783 | 0.715 | 0.707 |
| LEMPOSSTOPALPHA | 0.77 | 0.769 | 0.766 | 0.768 | 0.669 | 0.696 | 0.718 | 0.722 | 0.73 | **0.778** | 0.669 | 0.698 |
| LEMALPHA | 0.755 | 0.764 | 0.745 | **0.765** | 0.294 | 0.703 | 0.718 | 0.705 | 0.748 | 0.754 | 0.61 | 0.651 |
| LEMSTOP | 0.787 | 0.786 | 0.784 | **0.791** | 0.641 | 0.713 | 0.754 | 0.732 | 0.752 | 0.789 | 0.403 | 0.327 |
| LEMSTOPALPHA | 0.772 | 0.766 | 0.766 | **0.773** | 0.357 | 0.702 | 0.747 | 0.712 | 0.745 | 0.764 | 0.377 | 0.329 |
| POSS | 0.487 | 0.487 | 0.488 | 0.491 | 0.522 | 0.498 | **0.556** | 0.509 | 0.555 | 0.488 | 0.54 | 0.536 |
| POSSALPHA | 0.488 | 0.486 | 0.488 | 0.498 | 0.526 | 0.498 | **0.552** | 0.518 | 0.549 | 0.493 | 0.538 | 0.534 |
| POSSSTOP | 0.477 | 0.477 | 0.471 | 0.467 | 0.518 | 0.486 | **0.54** | 0.496 | 0.533 | 0.484 | 0.431 | 0.434 |
| POSSSTOPALPHA | 0.469 | 0.47 | 0.471 | 0.465 | 0.517 | 0.478 | **0.525** | 0.484 | 0.511 | 0.491 | 0.428 | 0.484 |
| TOK | 0.793 | 0.788 | 0.793 | **0.796** | 0.632 | 0.716 | 0.711 | 0.728 | 0.748 | **0.796** | 0.659 | 0.661 |
| TOKNERR | 0.741 | 0.744 | 0.737 | 0.743 | 0.6 | 0.696 | 0.688 | 0.671 | 0.719 | **0.749** | 0.655 | 0.631 |
| TOKNERRALPHA | 0.734 | 0.735 | 0.735 | 0.73 | 0.624 | 0.683 | 0.681 | 0.674 | 0.704 | **0.748** | 0.626 | 0.655 |
| TOKNERRSTOP | 0.736 | 0.736 | 0.728 | 0.732 | 0.609 | 0.68 | 0.73 | 0.678 | 0.71 | **0.751** | 0.406 | 0.317 |
| TOKNERRSTOPALPHA | 0.728 | 0.731 | 0.727 | 0.723 | 0.623 | 0.675 | 0.721 | 0.68 | 0.698 | **0.744** | 0.412 | 0.394 |
| TOKPOSS | 0.766 | 0.768 | 0.767 | **0.783** | 0.549 | 0.715 | 0.648 | 0.671 | 0.715 | 0.773 | 0.686 | 0.729 |
| TOKPOSSALPHA | 0.765 | 0.761 | 0.763 | 0.767 | 0.378 | 0.709 | 0.662 | 0.656 | 0.709 | **0.769** | 0.643 | 0.658 |
| TOKPOSSSTOP | 0.763 | 0.765 | 0.767 | **0.773** | 0.563 | 0.704 | 0.703 | 0.684 | 0.724 | 0.771 | 0.675 | 0.722 |
| TOKPOSSSTOPALPHA | 0.774 | 0.773 | 0.774 | 0.771 | 0.671 | 0.694 | 0.722 | 0.713 | 0.73 | **0.779** | 0.68 | 0.698 |
| TOKNER | **0.789** | 0.785 | 0.788 | **0.789** | 0.609 | 0.708 | 0.703 | 0.722 | 0.745 | 0.784 | 0.684 | 0.68 |
| TOKNERALPHA | 0.768 | 0.771 | 0.763 | **0.776** | 0.628 | 0.696 | 0.701 | 0.705 | 0.746 | 0.775 | 0.649 | 0.648 |
| TOKNERSTOP | 0.785 | **0.791** | 0.79 | 0.79 | 0.635 | 0.703 | 0.732 | 0.721 | 0.743 | 0.79 | 0.444 | 0.367 |
| TOKNERSTOPALPHA | 0.773 | 0.771 | 0.762 | **0.774** | 0.646 | 0.691 | 0.737 | 0.704 | 0.74 | 0.771 | 0.371 | 0.379 |
| TOKPOS | 0.781 | 0.783 | 0.791 | **0.798** | 0.565 | 0.713 | 0.656 | 0.72 | 0.739 | 0.787 | 0.626 | 0.705 |
| TOKPOSALPHA | 0.775 | 0.775 | 0.778 | **0.784** | 0.576 | 0.699 | 0.653 | 0.705 | 0.731 | 0.783 | 0.633 | 0.698 |
| TOKPOSSTOP | 0.786 | 0.783 | **0.794** | 0.792 | 0.645 | 0.7 | 0.711 | 0.733 | 0.739 | 0.789 | 0.706 | 0.691 |
| TOKPOSSTOPALPHA | 0.759 | **0.766** | 0.756 | 0.762 | 0.458 | 0.696 | 0.706 | 0.679 | 0.674 | 0.601 | 0.734 | 0.718 |
| TOKALPHA | 0.768 | 0.768 | 0.757 | **0.773** | 0.271 | 0.705 | 0.721 | 0.705 | 0.742 | 0.756 | 0.643 | 0.652 |
| TOKSTOP | 0.793 | 0.79 | 0.784 | **0.794** | 0.644 | 0.708 | 0.758 | 0.736 | 0.749 | 0.787 | 0.355 | 0.321 |
| TOKSTOPALPHA | 0.775 | **0.776** | 0.766 | **0.776** | 0.342 | 0.7 | 0.745 | 0.714 | 0.744 | 0.765 | 0.452 | 0.425 |

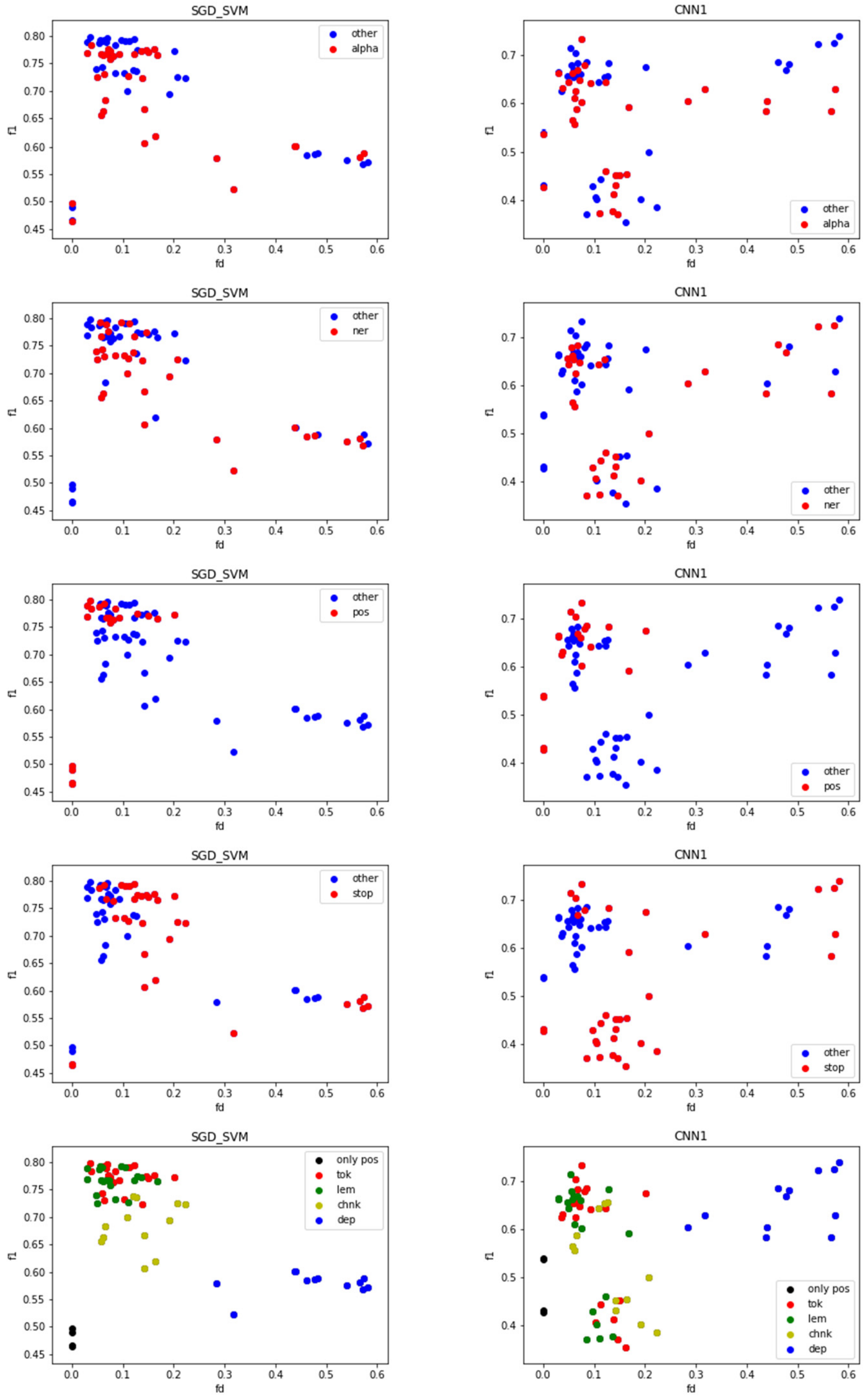

(i) Comparison of TOK (red), LEM (green), CHNK (yellow) and DEP (blue)

(j) Comparison of TOK (red), LEM (green), CHNK (yellow) and DEP (blue)

Figure 1: English dataset: FD & F1 score for SGD SVM (left) and CNN1 (right)

Table 8: English Dataset: Classifiers with best F1, preprocessing type and Pearson's correlation coefficient for FD and F1.

| Classifier | Best F1 | Best PP type | $\rho$(F1, FD) |
|---|---|---|---|
| SGD SVM | 0.798 | TOKPOS | -0.8239 |
| MLP | 0.7958 | TOK | -0.8599 |
| Linear SVM | 0.7941 | TOKPOSSTOP | -0.834 |
| L-BFGS LR | 0.7932 | TOKSTOP | -0.8024 |
| Newton LR | 0.7915 | TOKNERSTOP | -0.8097 |
| RandomForest | 0.7582 | TOKSTOP | -0.7873 |
| XGBoost | 0.7523 | LEMSTOP | -0.8303 |
| CNN1 | 0.7406 | DEPSTOP | 0.1633 |
| CNN2 | 0.7357 | LEMPOSS | 0.0951 |
| AdaBoost | 0.7356 | TOKSTOP | -0.7362 |
| NaiveBayes | 0.7165 | TOK | -0.7531 |
| KNN | 0.6711 | TOKPOSSSTOPALPHA | -0.7116 |

These results differ from the previous research [14], where CNN had a positive correlation with FD and the highest FD preprocessing methods, namely dependency relations, got the best results. The reason most likely lies in the parser differences (SpaCy used in our research vs. MeCAB[5] used in previous research [14]) and some differences in the network architecture, namely, previous research applied a simple BoW as a language model in CNNs, while we applied a more advanced word embeddings-based language model. The difference is very noticeable as the previous research achieved a score of .927 on the same dataset using CNNs with dependency relations, whereas we only managed to reach 0.802. This suggests two insights. Firstly, it is important to choose the the right and matured intermediary tools for handling the data (MeCab is a well established morphological parser, while the support for Japanese language in SpaCy was added only recently). Secondly, more advanced language models, even if achieving high results in many tasks, might be an overshoot for handling small-sized datasets, like the one used here.

The experiments show that changing Feature Density in moderate amounts could be useful for slightly increasing the F-score. However, excessive changes always showed diminishing results. The threshold was again in all cases approximately between 50% and 200% of the original density (TOK), most optimal FDs varying with each classifier in this range. Also, as the higher FD preprocessings performed quite poorly compared to previous research, we were unable to confirm the positive correlation between classifier

[5]https://taku910.github.io/mecab/

performance and FD for the Japanese dataset.

### 5.2.3. *Polish Cyberbullying Dataset*

Like with the previous two datasets, we analyzed the correlation of Feature Density with each of the classifiers using the proposed preprocessing methods. The classification results are represented in Table 12 and correlations in 13. Again, we excluded the preprocessing methods that only used POS tags due to poor performance caused by information loss.

From Table 13 we can see that most classifiers have a strong negative correlation with Feature Density, meaning that adding too much linguistic information, such as dependency relations, push the score down. The best results are within the range of .08 to .37 of FD depending on the classifier. This range includes 47 of the 68 feature sets (Table 4). If the weaker feature sets were to be left out, the power savings are approximately 2.1kWh for training the most expensive classifier, one-layer CNN, calculated from Table 6. This result is lower compared to English as cross-validation was not used.

One of the highest performing classifiers, SVM with SGD optimizer (Figure 3), indicates that the classifier performance decreases slightly as FD increases within this range. On the other hand, one-layer CNN shows that the maximum classifier performance first increases, peaks at around .20 and then decreases, which shows slight variance between classifiers.

Surprisingly, the performance of Neural Network-based classifiers was poor compared to other datasets. Their F1-score only reached the baseline classifier KNN. The reason behind this could be that Polish is grammatically a lot more complex language compared to, for example, English. The base Feature Density (TOK) of Polish is over double compared to English, which demonstrates this complexity. The linguistic complexity would most likely require more data to make the use of neural networks viable. The effect of FD with neural networks in Polish needs to be explored further using a larger dataset in the future.

The realization of cyberbullying in the Polish dataset had some differences to English as can be seen from Table 11. Polish tended to contain more indirect bullying as the amount of innuendos was almost three times higher than in the English dataset. The

Table 9: Japanese Cyberbullying Dataset: F1 for all preprocessing types & classifiers; best classifier for each dataset in **bold**; best preprocessing type for each underlined

| | LBFGS LR | Newton LR | Linear SVM | SGD SVM | KNN | NaiveBayes | RandomForest | AdaBoost | XGBoost | MLP | CNN1 | CNN2 |
|---|---|---|---|---|---|---|---|---|---|---|---|---|
| CHNK | 0.710 | 0.708 | 0.718 | 0.715 | 0.495 | 0.753 | 0.677 | 0.592 | 0.632 | 0.761 | **0.771** | 0.635 |
| CHNKNERR | 0.752 | 0.751 | 0.756 | 0.745 | 0.517 | 0.757 | 0.727 | 0.690 | 0.713 | 0.780 | **0.786** | 0.699 |
| CHNKNERRALPHA | 0.733 | 0.736 | 0.744 | 0.737 | 0.529 | 0.745 | 0.712 | 0.679 | 0.703 | 0.747 | **0.770** | 0.669 |
| CHNKNERRSTOP | 0.745 | 0.744 | 0.753 | 0.737 | 0.524 | 0.739 | 0.729 | 0.694 | 0.700 | 0.775 | **0.778** | 0.665 |
| CHNKNERRSTOPALPHA | 0.731 | 0.730 | **0.741** | 0.723 | 0.539 | 0.730 | 0.718 | 0.676 | 0.690 | 0.736 | 0.727 | 0.657 |
| CHNKNER | 0.758 | 0.758 | 0.769 | 0.763 | 0.523 | 0.772 | 0.738 | 0.696 | 0.714 | **0.796** | 0.789 | 0.735 |
| CHNKNERALPHA | 0.751 | 0.755 | 0.757 | 0.751 | 0.555 | 0.767 | 0.725 | 0.686 | 0.712 | 0.758 | **0.768** | 0.717 |
| CHNKNERSTOP | 0.754 | 0.755 | 0.764 | 0.776 | 0.543 | 0.767 | 0.738 | 0.696 | 0.709 | 0.784 | **0.791** | 0.768 |
| CHNKNERSTOPALPHA | 0.758 | 0.755 | 0.760 | **0.761** | 0.528 | 0.750 | 0.721 | 0.692 | 0.707 | 0.745 | 0.752 | 0.684 |
| CHNKALPHA | 0.714 | 0.719 | 0.709 | 0.702 | 0.510 | 0.740 | 0.684 | 0.590 | 0.622 | 0.738 | **0.747** | 0.629 |
| CHNKSTOP | 0.712 | 0.714 | 0.724 | 0.716 | 0.506 | 0.728 | 0.675 | 0.580 | 0.629 | **0.753** | 0.746 | 0.688 |
| CHNKSTOPALPHA | 0.695 | 0.701 | 0.714 | 0.700 | 0.518 | 0.721 | 0.666 | 0.572 | 0.628 | 0.702 | **0.722** | 0.604 |
| DEP | 0.678 | 0.681 | 0.685 | 0.680 | 0.495 | 0.682 | 0.608 | 0.534 | 0.583 | 0.764 | **0.765** | 0.573 |
| DEPNERR | 0.677 | 0.684 | 0.681 | 0.686 | 0.479 | 0.676 | 0.608 | 0.520 | 0.578 | **0.791** | 0.773 | 0.590 |
| DEPNERRALPHA | 0.667 | 0.671 | 0.675 | 0.674 | 0.461 | 0.670 | 0.585 | 0.524 | 0.576 | **0.767** | 0.762 | 0.587 |
| DEPNERRSTOP | 0.659 | 0.652 | 0.659 | 0.649 | 0.509 | 0.645 | 0.571 | 0.514 | 0.549 | **0.783** | 0.761 | 0.479 |
| DEPNERRSTOPALPHA | 0.617 | 0.617 | 0.619 | 0.618 | 0.493 | 0.627 | 0.534 | 0.504 | 0.523 | **0.755** | 0.738 | 0.592 |
| DEPNER | 0.727 | 0.727 | 0.742 | 0.746 | 0.502 | 0.724 | 0.708 | 0.690 | 0.707 | **0.792** | 0.748 | 0.547 |
| DEPNERALPHA | 0.730 | 0.731 | 0.741 | 0.748 | 0.427 | 0.718 | 0.707 | 0.689 | 0.703 | **0.782** | 0.756 | 0.575 |
| DEPNERSTOP | 0.717 | 0.716 | 0.728 | 0.723 | 0.544 | 0.701 | 0.693 | 0.681 | 0.690 | **0.802** | 0.750 | 0.654 |
| DEPNERSTOPALPHA | 0.715 | 0.711 | 0.722 | 0.713 | 0.448 | 0.731 | 0.695 | 0.683 | 0.685 | **0.779** | 0.739 | 0.671 |
| DEPALPHA | 0.669 | 0.668 | 0.681 | 0.674 | 0.464 | 0.674 | 0.593 | 0.529 | 0.571 | **0.756** | 0.705 | 0.621 |
| DEPSTOP | 0.655 | 0.658 | 0.658 | 0.653 | 0.507 | 0.647 | 0.575 | 0.523 | 0.556 | **0.776** | 0.731 | 0.659 |
| DEPSTOPALPHA | 0.627 | 0.623 | 0.622 | 0.622 | 0.499 | 0.628 | 0.537 | 0.516 | 0.535 | **0.746** | 0.743 | 0.608 |
| LEM | 0.803 | 0.799 | 0.817 | 0.806 | 0.592 | 0.823 | 0.771 | 0.669 | 0.723 | 0.863 | **0.868** | 0.721 |
| LEMNERR | 0.794 | 0.790 | 0.796 | 0.791 | 0.585 | 0.808 | 0.778 | 0.744 | 0.763 | **0.843** | 0.811 | 0.654 |
| LEMNERRALPHA | 0.797 | 0.796 | 0.805 | 0.793 | 0.585 | 0.811 | 0.781 | 0.748 | 0.768 | **0.845** | 0.828 | 0.708 |
| LEMNERRSTOP | 0.787 | 0.784 | 0.793 | 0.793 | 0.590 | 0.794 | 0.796 | 0.740 | 0.756 | **0.839** | 0.836 | 0.807 |
| LEMNERRSTOPALPHA | 0.790 | 0.791 | 0.798 | 0.793 | 0.587 | 0.796 | 0.788 | 0.743 | 0.764 | 0.810 | **0.814** | 0.776 |
| LEMPOSS | 0.815 | 0.817 | 0.851 | 0.842 | 0.711 | 0.855 | 0.785 | 0.718 | 0.780 | 0.866 | **0.877** | 0.768 |
| LEMPOSSALPHA | 0.825 | 0.823 | 0.848 | 0.837 | 0.734 | 0.847 | 0.770 | 0.690 | 0.759 | 0.856 | **0.867** | 0.572 |
| LEMPOSSSTOP | 0.816 | 0.818 | 0.850 | 0.846 | 0.529 | 0.851 | 0.774 | 0.718 | 0.764 | 0.868 | **0.876** | 0.820 |
| LEMPOSSSTOPALPHA | 0.819 | 0.817 | 0.841 | 0.829 | 0.722 | 0.844 | 0.779 | 0.686 | 0.743 | 0.850 | **0.851** | 0.688 |
| LEMNER | 0.806 | 0.801 | 0.818 | 0.816 | 0.633 | 0.823 | 0.790 | 0.747 | 0.775 | **0.870** | 0.865 | 0.616 |
| LEMNERALPHA | 0.809 | 0.807 | 0.827 | 0.823 | 0.612 | 0.827 | 0.797 | 0.744 | 0.776 | 0.857 | **0.860** | 0.772 |
| LEMNERSTOP | 0.799 | 0.798 | 0.819 | 0.823 | 0.632 | 0.831 | 0.802 | 0.742 | 0.767 | **0.875** | 0.870 | 0.807 |
| LEMNERSTOPALPHA | 0.805 | 0.808 | 0.830 | 0.826 | 0.643 | 0.834 | 0.802 | 0.744 | 0.769 | 0.848 | **0.849** | 0.741 |
| LEMPOS | 0.806 | 0.808 | 0.845 | 0.850 | 0.716 | 0.830 | 0.786 | 0.742 | 0.768 | 0.868 | **0.870** | 0.619 |
| LEMPOSALPHA | 0.807 | 0.803 | 0.838 | 0.846 | 0.704 | 0.828 | 0.778 | 0.728 | 0.772 | 0.856 | **0.860** | 0.663 |
| LEMPOSSTOP | 0.802 | 0.795 | 0.841 | 0.847 | 0.567 | 0.836 | 0.770 | 0.730 | 0.769 | 0.868 | **0.877** | 0.808 |
| LEMPOSSTOPALPHA | 0.791 | 0.789 | 0.830 | 0.836 | 0.700 | 0.837 | 0.768 | 0.707 | 0.754 | 0.835 | **0.860** | 0.783 |
| LEMALPHA | 0.800 | 0.808 | 0.824 | 0.807 | 0.587 | 0.815 | 0.758 | 0.666 | 0.727 | **0.854** | 0.853 | 0.805 |
| LEMSTOP | 0.799 | 0.801 | 0.813 | 0.802 | 0.614 | 0.812 | 0.785 | 0.659 | 0.723 | 0.858 | **0.862** | 0.809 |
| LEMSTOPALPHA | 0.803 | 0.804 | 0.813 | 0.807 | 0.619 | 0.821 | 0.776 | 0.665 | 0.727 | 0.842 | **0.844** | 0.801 |
| POSS | 0.645 | 0.646 | 0.634 | 0.627 | 0.558 | 0.620 | 0.645 | 0.643 | 0.632 | 0.645 | **0.670** | 0.490 |
| POSSALPHA | 0.646 | 0.650 | 0.635 | 0.619 | 0.557 | 0.618 | 0.644 | 0.646 | 0.640 | 0.649 | **0.658** | 0.567 |
| POSSSTOP | 0.641 | 0.643 | 0.634 | 0.637 | 0.553 | 0.626 | 0.638 | **0.643** | 0.630 | 0.635 | 0.623 | 0.492 |
| POSSSTOPALPHA | 0.615 | **0.616** | 0.603 | 0.610 | 0.506 | 0.614 | 0.610 | 0.609 | 0.595 | 0.614 | 0.587 | 0.508 |
| TOK | 0.801 | 0.797 | 0.813 | 0.802 | 0.592 | 0.817 | 0.766 | 0.647 | 0.724 | **0.864** | 0.857 | 0.792 |
| TOKNERR | 0.789 | 0.785 | 0.797 | 0.799 | 0.589 | 0.798 | 0.783 | 0.729 | 0.756 | 0.838 | **0.844** | 0.793 |
| TOKNERRALPHA | 0.791 | 0.792 | 0.802 | 0.792 | 0.567 | 0.797 | 0.778 | 0.737 | 0.759 | **0.836** | 0.821 | 0.722 |
| TOKNERRSTOP | 0.776 | 0.778 | 0.795 | 0.783 | 0.582 | 0.788 | 0.776 | 0.727 | 0.756 | 0.835 | **0.843** | 0.709 |
| TOKNERRSTOPALPHA | 0.785 | 0.787 | 0.796 | 0.795 | 0.571 | 0.783 | 0.778 | 0.736 | 0.761 | **0.818** | 0.817 | 0.742 |
| TOKPOSS | 0.815 | 0.812 | 0.843 | 0.835 | 0.695 | 0.849 | 0.775 | 0.719 | 0.756 | 0.876 | **0.876** | 0.683 |
| TOKPOSSALPHA | 0.788 | 0.789 | 0.829 | 0.833 | 0.716 | 0.823 | 0.752 | 0.715 | 0.748 | 0.864 | **0.865** | 0.646 |
| TOKPOSSSTOP | 0.823 | 0.814 | 0.834 | 0.828 | 0.713 | 0.848 | 0.767 | 0.701 | 0.760 | **0.855** | 0.821 | 0.643 |
| TOKPOSSSTOPALPHA | 0.817 | 0.822 | 0.837 | 0.835 | 0.518 | 0.846 | 0.764 | 0.703 | 0.764 | 0.876 | 0.883 | 0.759 |
| TOKNER | 0.814 | 0.812 | 0.830 | 0.828 | 0.718 | 0.835 | 0.763 | 0.667 | 0.731 | 0.846 | **0.847** | 0.686 |
| TOKNERALPHA | 0.798 | 0.800 | 0.816 | 0.820 | 0.610 | 0.825 | 0.790 | 0.734 | 0.762 | **0.870** | 0.864 | 0.834 |
| TOKNERSTOP | 0.803 | 0.805 | 0.825 | 0.824 | 0.608 | 0.822 | 0.790 | 0.745 | 0.768 | 0.842 | **0.857** | 0.811 |
| TOKNERSTOPALPHA | 0.794 | 0.794 | 0.819 | 0.820 | 0.616 | 0.820 | 0.797 | 0.735 | 0.759 | 0.878 | 0.875 | 0.849 |
| TOKPOS | 0.799 | 0.796 | 0.822 | 0.828 | 0.630 | 0.826 | 0.794 | 0.730 | 0.763 | 0.846 | **0.847** | 0.773 |
| TOKPOSALPHA | 0.808 | 0.807 | 0.844 | 0.852 | 0.706 | 0.828 | 0.777 | 0.724 | 0.770 | **0.870** | 0.865 | 0.674 |
| TOKPOSSTOP | 0.802 | 0.800 | 0.839 | 0.840 | 0.691 | 0.823 | 0.771 | 0.715 | 0.763 | 0.839 | **0.857** | 0.615 |
| TOKPOSSTOPALPHA | 0.795 | 0.794 | 0.834 | 0.844 | 0.560 | 0.832 | 0.766 | 0.720 | 0.762 | 0.863 | **0.873** | 0.661 |
| TOKALPHA | 0.801 | 0.799 | 0.807 | 0.804 | 0.584 | 0.817 | 0.761 | 0.655 | 0.712 | 0.847 | **0.848** | 0.798 |
| TOKSTOP | 0.803 | 0.801 | 0.815 | 0.803 | 0.584 | 0.809 | 0.778 | 0.646 | 0.717 | 0.857 | **0.861** | 0.782 |
| TOKSTOPALPHA | 0.800 | 0.800 | 0.811 | 0.813 | 0.591 | 0.821 | 0.779 | 0.650 | 0.717 | **0.827** | 0.818 | 0.710 |

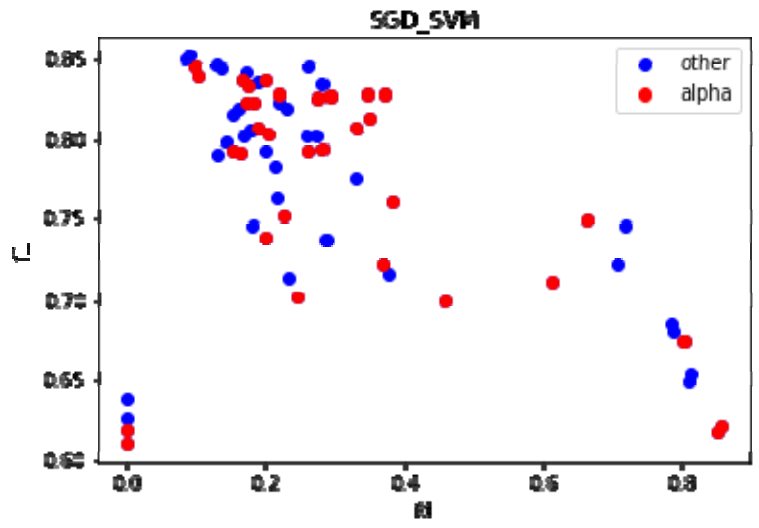

(a) Alphabetic filtering (red) vs others (blue)

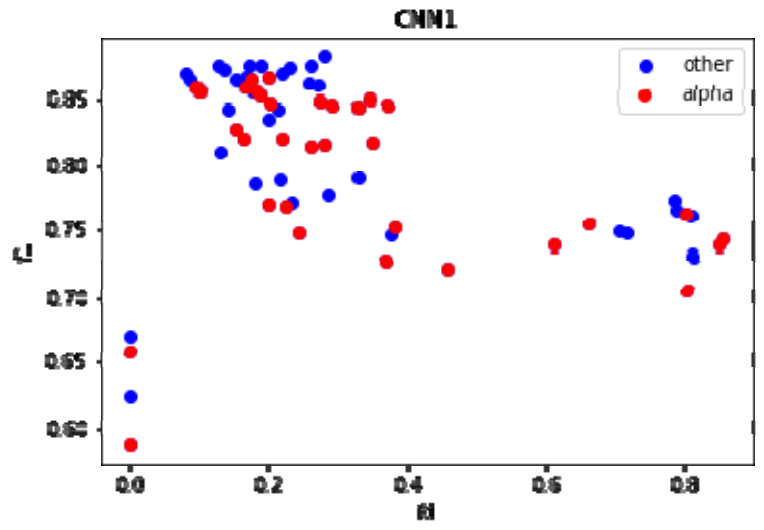

(b) Alphabetic filtering (red) vs others (blue)

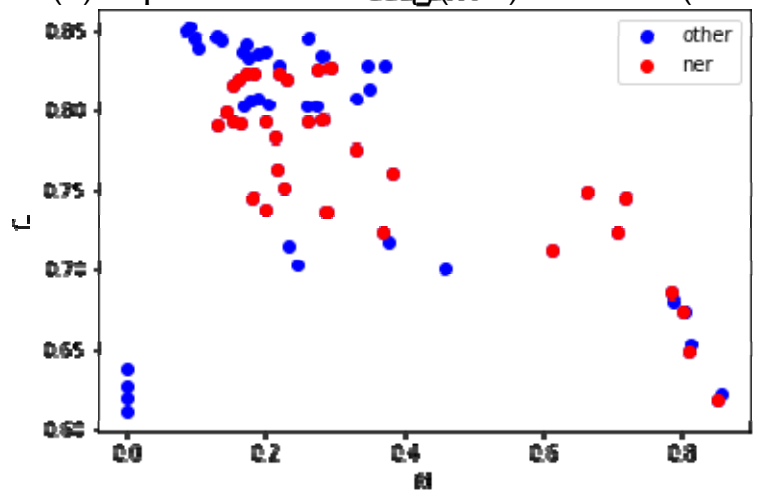

(c) NER (red) vs others (blue)

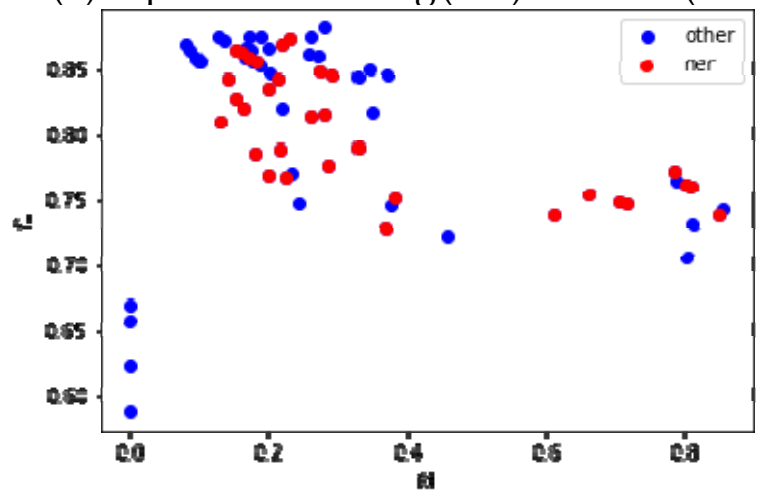

(d) NER (red) vs others (blue)

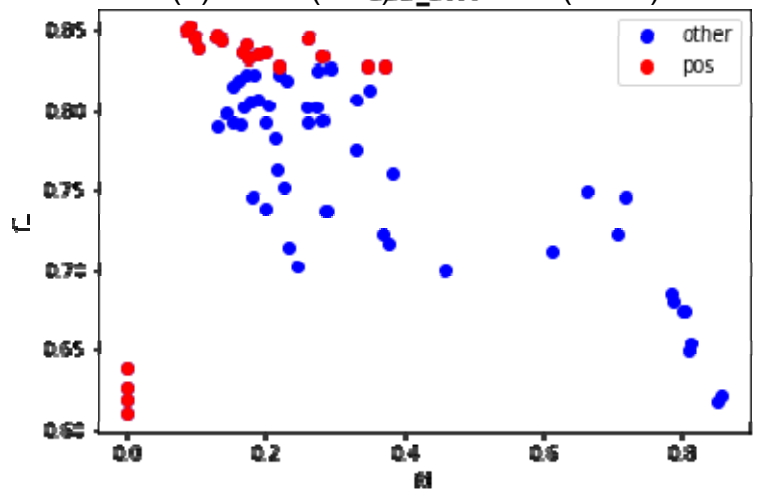

(e) POS (red) vs others (blue)

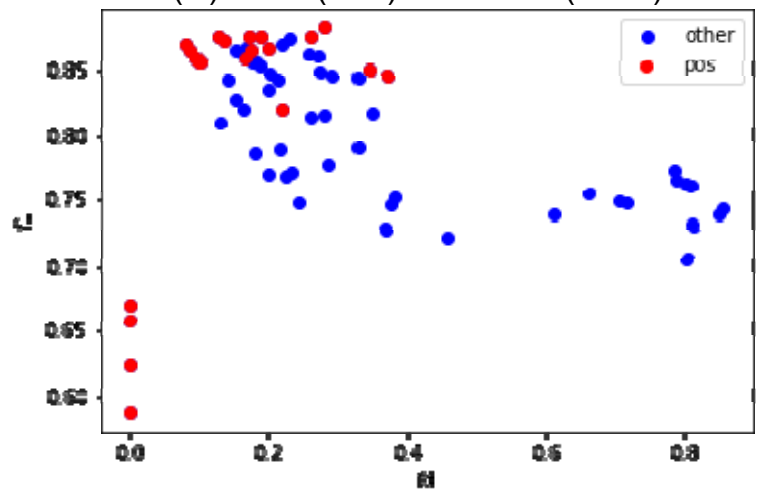

(f) POS (red) vs others (blue)

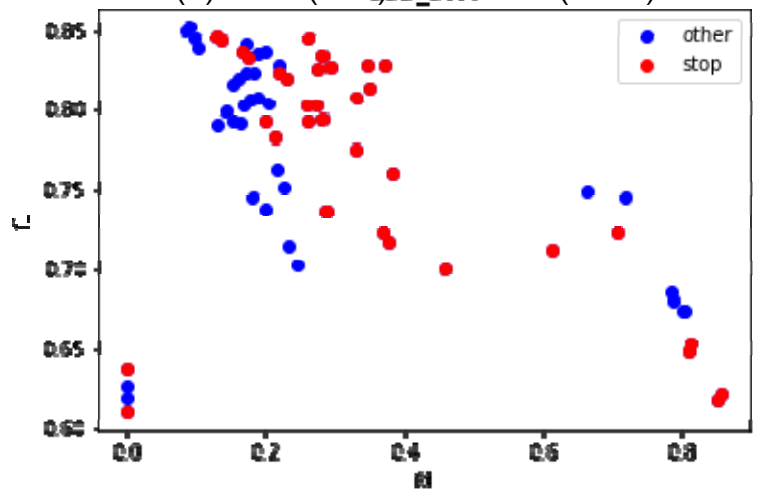

(g) Stopword filtering (red) vs others (blue)

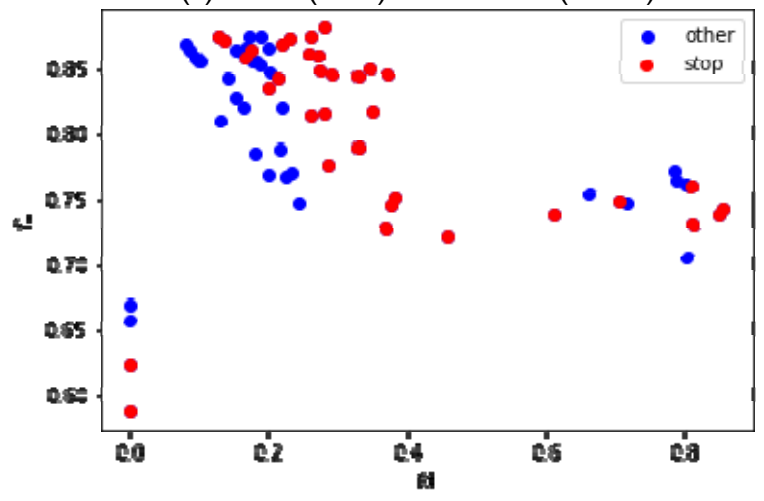

(h) Stopword filtering (red) vs others (blue)

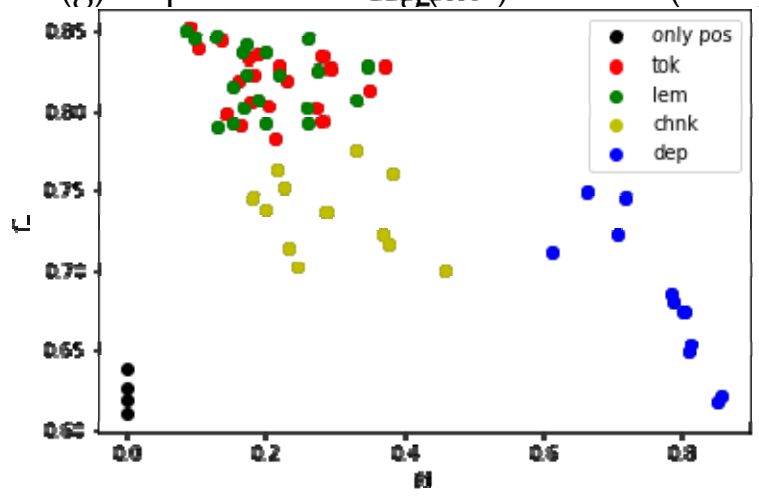

(i) Comparison of TOK (red), LEM (green), CHNK (yellow) and DEP (blue)

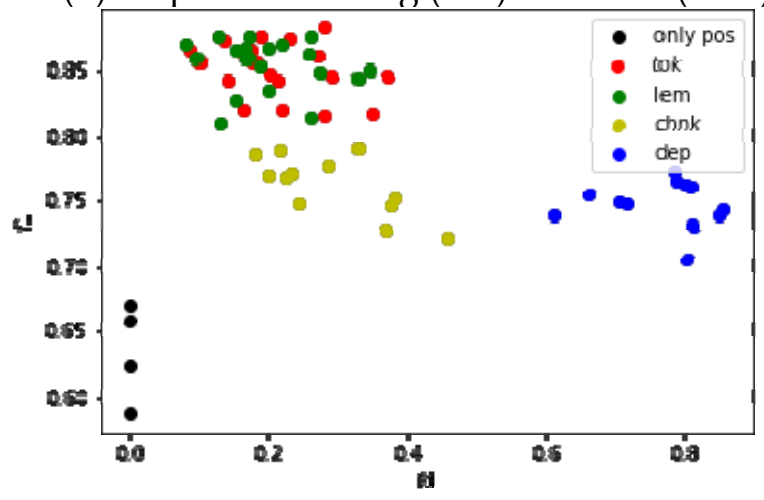

(j) Comparison of TOK (red), LEM (green), CHNK (yellow) and DEP (blue)

Figure 2: Japanese Dataset: FD & F1 score for SGD SVM (left) and CNN1 (right)

Table 10: Japanese Dataset: Classifiers with best F1, preprocessing type and Pearson's correlation coefficient for FD and F1.

| Classifier | Best F1 | Best PP type | $\rho$(F1, FD) |
|---|---|---|---|
| CNN1 | 0.8829 | TOKPOSSTOP | -0.7356 |
| MLP | 0.8776 | TOKNERSTOP | -0.6107 |
| NaiveBayes | 0.8552 | LEMPOS | -0.8675 |
| SGDSVM | 0.8523 | TOKPOSS | -0.8270 |
| Linear SVM | 0.8508 | LEMPOS | -0.8415 |
| CNN2 | 0.8488 | TOKNERRSTOP | -0.5637 |
| L-BFGSLR | 0.8251 | LEMPOSALPHA | -0.8506 |
| Newton LR | 0.8227 | LEMPOSALPHA | -0.8511 |
| RandomForest | 0.8022 | LEMNERSTOPALPHA | -0.8711 |
| XGBoost | 0.7801 | LEMPOS | -0.8461 |
| AdaBoost | 0.7484 | LEMNERRALPHA | -0.7902 |
| KNN | 0.7339 | LEMPOSALPHA | -0.6320 |

Table 11: Categories of cyberbullying in English and Polish datasets

| CB category | English # | Polish # | English % | Polish % |
|---|---|---|---|---|
| Threat | 29 | 25 | 3.2 | 2.5 |
| Innuendo | 24 | 68 | 2.6 | 6.9 |
| Sexual harassment | 92 | N/A | 10 | N/A |
| Insult | 198 | 459 | 21.6 | 46.6 |
| Blackmail | 5 | 0 | 0.5 | 0 |
| Mockery | 338 | 471 | 36.9 | 42.3 |
| Phishing, revealing private information | 0 | 23 | 0 | 2.3 |
| Vulgarism | 135 | 129 | 14.7 | 13.1 |

Polish dataset also contained some revealing of personal information while the English dataset had none. Also, Polish contained more personal insults than the English dataset. On the other hand, the English dataset contained some blackmailing, while Polish did not have any. The fact that Polish had more indirect bullying in the form of innuendos compared to English and that Polish language is more complex than English could be the reasons for the overall lower performance of the classifiers on the Polish dataset.

### *5.2.4. Verification Dataset*

Also for the verification dataset, we analyzed the correlation of Feature Density with each of the classifiers using the proposed preprocessing methods. The classification scores are shown in Table 14 and correlations in Table 15. After excluding the preprocessing methods that only used POS tags, we can see that Logistic Regression and SVMs have a strong negative correlation with Feature Density. So these classifiers seem to have a weaker performance if a lot of linguistic information is added, and the best results being usually within the range of .002 to .015 FD depending on the classifier.

Table 12: Polish Dataset: F1 for all preprocessing types & classifiers; best classifier for each dataset in **bold**; best preprocessing type for each underlined

| | LBFGS LR | Newton LR | Linear SVM | SGD SVM | KNN | NaiveBayes | RandomForest | AdaBoost | XGBoost | MLP | CNN1 | CNN2 |
|---|---|---|---|---|---|---|---|---|---|---|---|---|
| CHNK | 0.437 | 0.419 | 0.424 | 0.443 | 0.163 | **0.455** | 0.124 | 0.308 | 0.278 | 0.245 | 0.213 | 0.270 |
| CHNKNERR | 0.448 | 0.433 | 0.455 | **0.462** | 0.175 | 0.446 | 0.099 | 0.272 | 0.183 | 0.234 | 0.249 | 0.292 |
| CHNKNERRALPHA | 0.444 | 0.440 | 0.446 | **0.469** | 0.201 | 0.468 | 0.085 | 0.291 | 0.205 | 0.224 | 0.228 | 0.211 |
| CHNKNERRSTOP | **0.447** | 0.414 | 0.408 | 0.421 | 0.231 | 0.400 | 0.187 | 0.333 | 0.299 | 0.223 | 0.250 | 0.312 |
| CHNKNERRSTOPALPHA | 0.425 | 0.427 | 0.425 | **0.452** | 0.241 | 0.401 | 0.212 | 0.283 | 0.246 | 0.276 | 0.262 | 0.187 |
| CHNKNER | 0.439 | 0.435 | 0.416 | 0.445 | 0.170 | **0.459** | 0.113 | 0.304 | 0.205 | 0.271 | 0.155 | 0.192 |
| CHNKNERALPHA | 0.444 | 0.437 | 0.424 | 0.419 | 0.182 | **0.459** | 0.086 | 0.265 | 0.236 | 0.290 | 0.189 | 0.183 |
| CHNKNERSTOP | 0.417 | 0.414 | 0.396 | **0.430** | 0.225 | 0.413 | 0.176 | 0.308 | 0.295 | 0.254 | 0.194 | 0.307 |
| CHNKNERSTOPALPHA | 0.459 | **0.462** | 0.436 | 0.431 | 0.213 | 0.441 | 0.174 | 0.325 | 0.309 | 0.265 | 0.231 | 0.260 |
| CHNKALPHA | 0.435 | 0.452 | 0.412 | 0.435 | 0.180 | **0.457** | 0.085 | 0.256 | 0.192 | 0.259 | 0.228 | 0.268 |
| CHNKSTOP | 0.448 | **0.453** | 0.394 | 0.433 | 0.199 | 0.403 | 0.188 | 0.388 | 0.331 | 0.267 | 0.298 | 0.307 |
| CHNKSTOPALPHA | 0.450 | **0.459** | 0.441 | 0.459 | 0.225 | 0.438 | 0.208 | 0.330 | 0.266 | 0.279 | 0.264 | 0.309 |
| DEP | 0.316 | 0.316 | 0.204 | 0.199 | 0.235 | **0.346** | 0.029 | 0.250 | 0.070 | 0.125 | 0.200 | 0.123 |
| DEPNERR | 0.322 | 0.312 | 0.214 | 0.211 | 0.243 | **0.340** | 0.029 | 0.242 | 0.057 | 0.128 | 0.197 | 0.057 |
| DEPNERRALPHA | 0.261 | 0.251 | 0.200 | 0.233 | 0.263 | **0.335** | 0.043 | 0.260 | 0.071 | 0.176 | 0.115 | 0.213 |
| DEPNERRSTOP | 0.276 | 0.290 | 0.195 | 0.182 | 0.250 | **0.301** | 0.029 | 0.195 | 0.043 | 0.122 | 0.260 | 0.055 |
| DEPNERRSTOPALPHA | 0.159 | 0.158 | 0.154 | 0.165 | 0.260 | **0.264** | 0.058 | 0.119 | 0.057 | 0.158 | 0.171 | 0.027 |
| DEPNER | 0.345 | 0.339 | 0.191 | 0.210 | 0.224 | **0.345** | 0.043 | 0.269 | 0.068 | 0.140 | 0.201 | 0.044 |
| DEPNERALPHA | 0.250 | 0.240 | 0.201 | 0.222 | 0.193 | **0.329** | 0.044 | 0.230 | 0.043 | 0.111 | 0.044 | 0.000 |
| DEPNERSTOP | **0.303** | 0.299 | 0.173 | 0.160 | 0.221 | 0.289 | 0.044 | 0.178 | 0.042 | 0.077 | 0.206 | 0.146 |
| DEPNERSTOPALPHA | 0.172 | 0.181 | 0.166 | 0.167 | 0.175 | **0.239** | 0.058 | 0.093 | 0.056 | 0.201 | 0.177 | 0.168 |
| DEPALPHA | 0.230 | 0.251 | 0.200 | 0.210 | 0.260 | **0.358** | 0.043 | 0.234 | 0.071 | 0.197 | 0.263 | 0.133 |
| DEPSTOP | 0.273 | 0.289 | 0.184 | 0.193 | 0.240 | **0.304** | 0.043 | 0.215 | 0.070 | 0.111 | 0.134 | 0.014 |
| DEPSTOPALPHA | 0.155 | 0.168 | 0.176 | 0.152 | 0.256 | **0.276** | 0.071 | 0.105 | 0.044 | 0.106 | 0.049 | 0.135 |
| LEM | **0.481** | 0.473 | 0.428 | 0.444 | 0.281 | 0.460 | 0.125 | 0.361 | 0.317 | 0.285 | 0.281 | 0.275 |
| LEMNERR | 0.466 | 0.456 | 0.462 | **0.471** | 0.273 | 0.452 | 0.113 | 0.309 | 0.244 | 0.277 | 0.308 | 0.307 |
| LEMNERRALPHA | 0.455 | 0.449 | 0.452 | 0.464 | 0.239 | **0.478** | 0.112 | 0.341 | 0.214 | 0.324 | 0.262 | 0.233 |
| LEMNERRSTOP | 0.484 | 0.484 | 0.494 | 0.479 | 0.302 | 0.417 | 0.221 | 0.347 | 0.310 | 0.306 | 0.263 | 0.283 |
| LEMNERRSTOPALPHA | 0.482 | 0.484 | 0.471 | **0.496** | 0.272 | 0.439 | 0.211 | 0.364 | 0.287 | 0.296 | 0.333 | 0.312 |
| LEMPOSS | 0.494 | 0.483 | 0.444 | 0.456 | 0.265 | 0.459 | 0.112 | 0.390 | 0.214 | 0.287 | 0.313 | 0.190 |
| LEMPOSSALPHA | **0.467** | 0.462 | 0.421 | 0.438 | 0.287 | 0.444 | 0.125 | 0.354 | 0.221 | 0.279 | 0.260 | 0.187 |
| LEMPOSSSTOP | 0.467 | 0.451 | 0.457 | **0.483** | 0.322 | 0.437 | 0.237 | 0.448 | 0.345 | 0.297 | 0.338 | 0.241 |
| LEMPOSSSTOPALPHA | 0.490 | **0.498** | 0.448 | 0.474 | 0.352 | 0.426 | 0.255 | 0.421 | 0.385 | 0.312 | 0.258 | 0.288 |
| LEMNER | **0.466** | 0.465 | 0.419 | 0.450 | 0.281 | 0.464 | 0.125 | 0.381 | 0.241 | 0.267 | 0.239 | 0.237 |
| LEMNERALPHA | **0.487** | 0.464 | 0.414 | 0.431 | 0.225 | 0.482 | 0.099 | 0.386 | 0.232 | 0.275 | 0.245 | 0.265 |
| LEMNERSTOP | 0.461 | **0.494** | 0.467 | 0.490 | 0.301 | 0.430 | 0.224 | 0.391 | 0.378 | 0.348 | 0.310 | 0.334 |
| LEMNERSTOPALPHA | **0.486** | 0.473 | 0.477 | 0.467 | 0.235 | 0.461 | 0.232 | 0.391 | 0.343 | 0.346 | 0.354 | 0.315 |
| LEMPOS | 0.447 | 0.457 | **0.472** | 0.465 | 0.271 | 0.440 | 0.099 | 0.427 | 0.219 | 0.266 | 0.271 | 0.120 |
| LEMPOSALPHA | **0.484** | 0.473 | 0.420 | 0.469 | 0.286 | 0.462 | 0.110 | 0.361 | 0.186 | 0.275 | 0.258 | 0.193 |
| LEMPOSSTOP | 0.461 | 0.457 | 0.491 | **0.498** | 0.341 | 0.422 | 0.174 | 0.438 | 0.259 | 0.305 | 0.232 | 0.164 |
| LEMPOSSTOPALPHA | 0.449 | **0.498** | 0.466 | 0.481 | 0.357 | 0.419 | 0.192 | 0.438 | 0.256 | 0.325 | 0.307 | 0.246 |
| LEMALPHA | 0.491 | **0.493** | 0.440 | 0.436 | 0.225 | 0.462 | 0.112 | 0.381 | 0.220 | 0.315 | 0.254 | 0.250 |
| LEMSTOP | 0.490 | **0.500** | 0.469 | 0.500 | 0.281 | 0.430 | 0.231 | 0.408 | 0.383 | 0.305 | 0.299 | 0.264 |
| LEMSTOPALPHA | 0.494 | 0.502 | 0.452 | 0.465 | 0.242 | 0.453 | 0.231 | 0.425 | 0.371 | 0.330 | 0.292 | 0.319 |
| POSS | 0.313 | 0.314 | 0.298 | **0.315** | 0.184 | 0.301 | 0.141 | 0.293 | 0.172 | 0.293 | 0.204 | 0.198 |
| POSSALPHA | 0.310 | **0.316** | 0.297 | 0.301 | 0.180 | 0.301 | 0.113 | 0.298 | 0.179 | 0.282 | 0.235 | 0.170 |
| POSSSTOP | 0.295 | **0.297** | 0.288 | 0.266 | 0.188 | 0.251 | 0.193 | 0.282 | 0.223 | 0.257 | 0.258 | 0.230 |
| POSSSTOPALPHA | 0.256 | 0.246 | 0.247 | 0.229 | 0.078 | 0.239 | 0.203 | 0.264 | 0.240 | 0.246 | 0.240 | **0.267** |
| TOK | 0.437 | 0.444 | 0.422 | **0.456** | 0.166 | 0.455 | 0.099 | 0.303 | 0.171 | 0.278 | 0.229 | 0.248 |
| TOKNERR | 0.440 | 0.444 | 0.432 | **0.465** | 0.178 | 0.455 | 0.072 | 0.266 | 0.194 | 0.245 | 0.203 | 0.212 |
| TOKNERRALPHA | 0.424 | 0.432 | 0.443 | **0.472** | 0.190 | 0.451 | 0.099 | 0.275 | 0.181 | 0.266 | 0.230 | 0.262 |
| TOKNERRSTOP | 0.425 | **0.430** | 0.406 | 0.424 | 0.244 | 0.384 | 0.150 | 0.314 | 0.267 | 0.226 | 0.262 | 0.288 |
| TOKNERRSTOPALPHA | 0.434 | 0.432 | 0.425 | **0.458** | 0.229 | 0.408 | 0.151 | 0.284 | 0.267 | 0.242 | 0.265 | 0.236 |
| TOKPOSS | 0.443 | 0.441 | 0.450 | **0.464** | 0.264 | 0.461 | 0.071 | 0.321 | 0.205 | 0.201 | 0.258 | 0.056 |
| TOKPOSSALPHA | 0.393 | 0.400 | 0.403 | **0.427** | 0.317 | 0.411 | 0.185 | 0.317 | 0.251 | 0.208 | 0.208 | 0.040 |
| TOKPOSSSTOP | 0.443 | 0.449 | 0.424 | 0.427 | 0.273 | **0.457** | 0.086 | 0.246 | 0.160 | 0.253 | 0.174 | 0.173 |
| TOKPOSSSTOPALPHA | **0.430** | 0.420 | 0.400 | 0.404 | 0.306 | 0.416 | 0.137 | 0.345 | 0.277 | 0.190 | 0.251 | 0.212 |
| TOKNER | 0.429 | **0.442** | 0.425 | 0.429 | 0.327 | 0.432 | 0.197 | 0.297 | 0.263 | 0.284 | 0.227 | 0.257 |
| TOKNERALPHA | 0.444 | 0.439 | 0.416 | 0.448 | 0.170 | **0.453** | 0.085 | 0.314 | 0.191 | 0.309 | 0.210 | 0.188 |
| TOKNERSTOP | 0.447 | 0.435 | 0.428 | 0.435 | 0.176 | **0.461** | 0.085 | 0.282 | 0.192 | 0.267 | 0.220 | 0.162 |
| TOKNERSTOPALPHA | 0.429 | 0.427 | 0.398 | **0.430** | 0.242 | 0.405 | 0.174 | 0.335 | 0.291 | 0.271 | 0.221 | 0.281 |
| TOKPOS | 0.451 | **0.473** | 0.445 | 0.471 | 0.225 | 0.438 | 0.199 | 0.316 | 0.298 | 0.272 | 0.257 | 0.259 |
| TOKPOSALPHA | 0.422 | 0.434 | 0.421 | **0.459** | 0.255 | 0.450 | 0.058 | 0.302 | 0.171 | 0.271 | 0.212 | 0.071 |
| TOKPOSSTOP | 0.434 | 0.436 | 0.429 | **0.447** | 0.266 | 0.442 | 0.043 | 0.259 | 0.168 | 0.213 | 0.214 | 0.129 |
| TOKPOSSTOPALPHA | 0.388 | 0.393 | 0.404 | **0.428** | 0.286 | 0.398 | 0.138 | 0.300 | 0.264 | 0.208 | 0.208 | 0.040 |
| TOKALPHA | 0.442 | 0.446 | 0.414 | **0.456** | 0.176 | 0.448 | 0.112 | 0.303 | 0.215 | 0.261 | 0.232 | 0.286 |
| TOKSTOP | 0.438 | **0.440** | 0.392 | 0.429 | 0.220 | 0.423 | 0.188 | 0.387 | 0.287 | 0.234 | 0.288 | 0.352 |
| TOKSTOPALPHA | **0.463** | 0.448 | 0.434 | 0.423 | 0.217 | 0.451 | 0.172 | 0.335 | 0.271 | 0.344 | 0.283 | 0.276 |

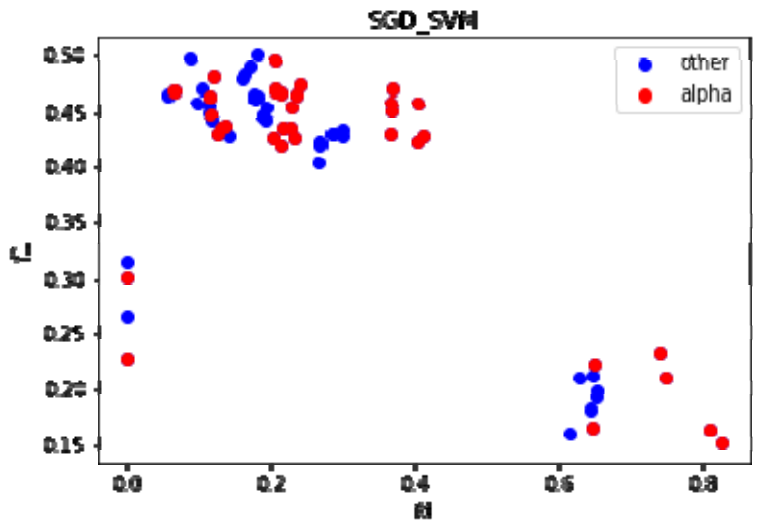

(a) Alphabetic filtering (red) vs others (blue)

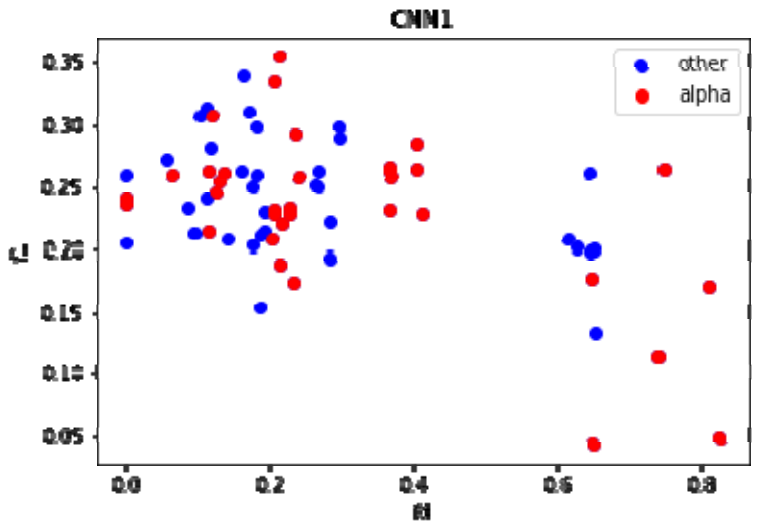

(b) Alphabetic filtering (red) vs others (blue)

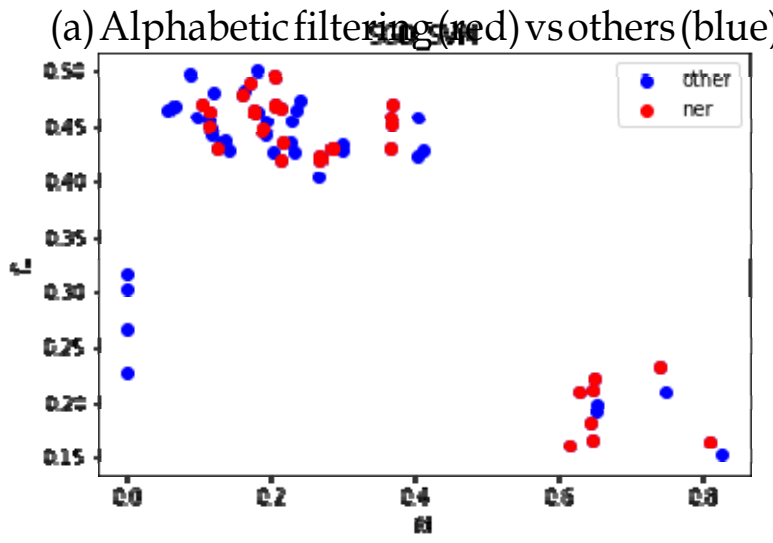

(c) NER (red) vs others (blue)

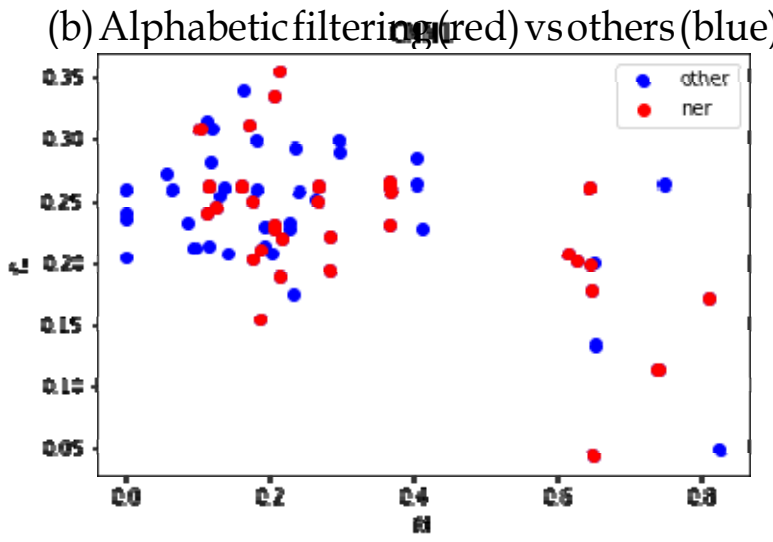

(d) NER (red) vs others (blue)

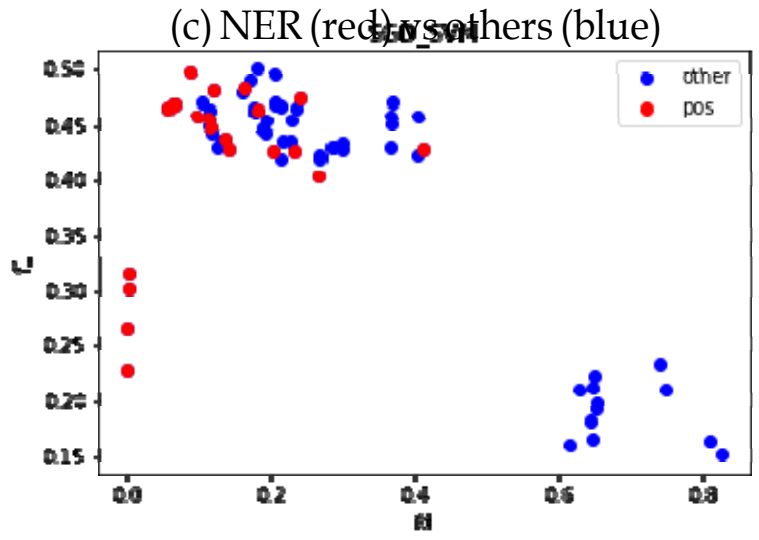

(e) POS (red) vs others (blue)

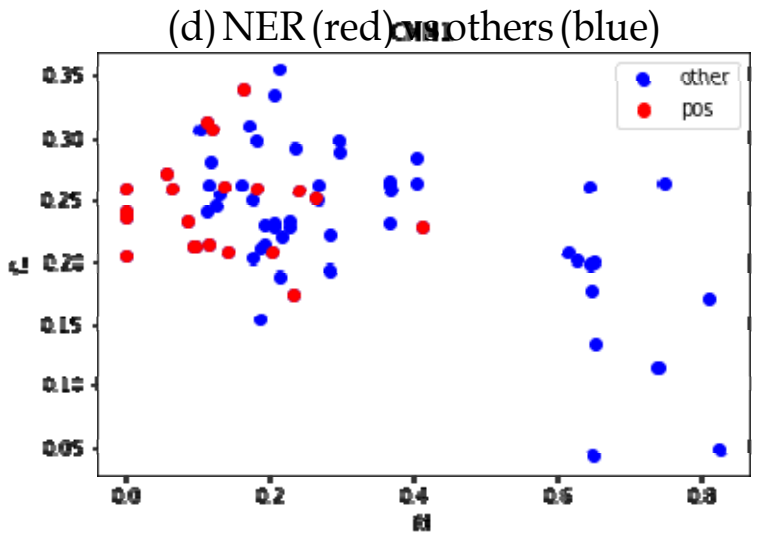

(f) POS (red) vs others (blue)

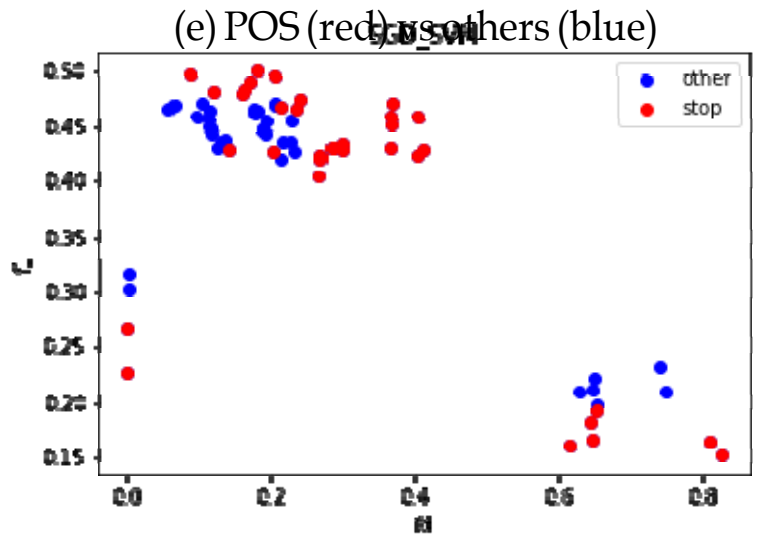

(g) Stopword filtering (red) vs others (blue)

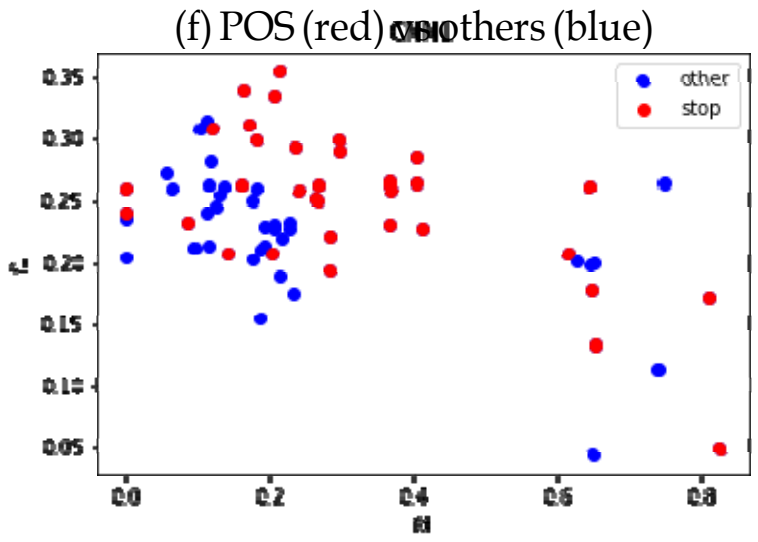

(h) Stopword filtering (red) vs others (blue)

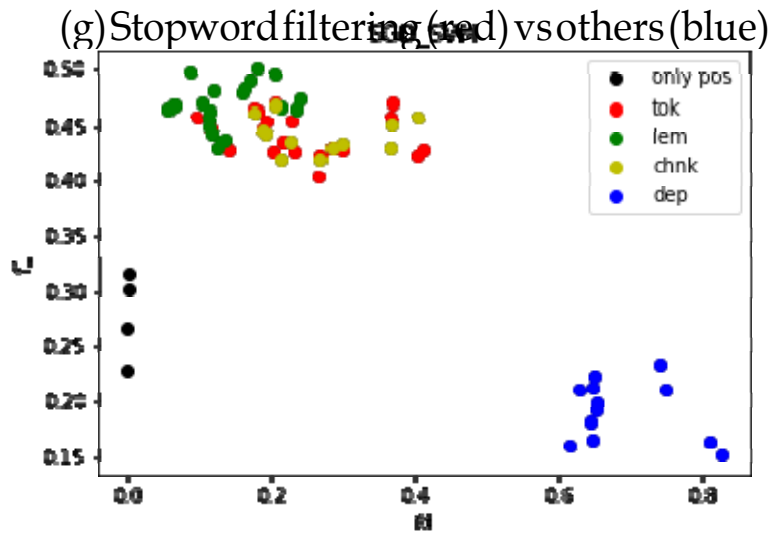

(i) Comparison of TOK (red), LEM (green), CHNK (yellow) and DEP (blue)

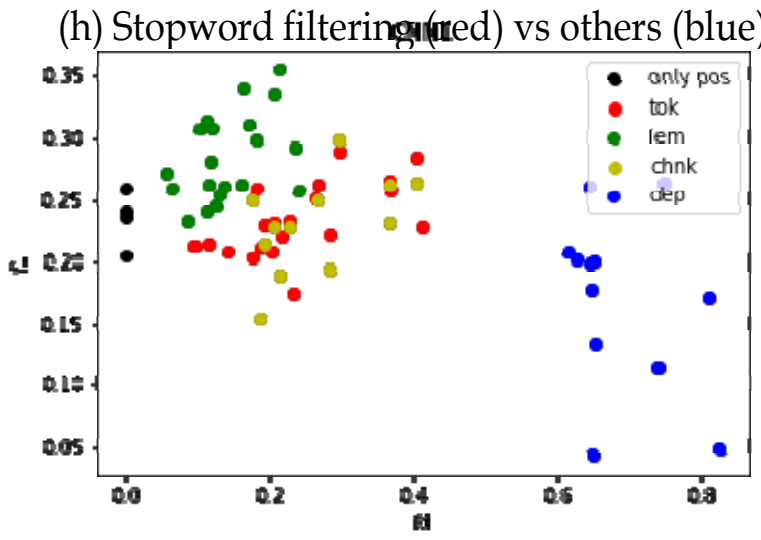

(j) Comparison of TOK (red), LEM (green), CHNK (yellow) and DEP (blue)

Figure 3: Polish Dataset: FD & F1 score for SGDSVM (left) and CNN1 (right)

Table 13: Polish Dataset: Classifiers with best F1, preprocessing type and Pearson's correlation coefficient for FD and F1.

| Classifier | Best F1 | Best PP type | $\rho$(F1, FD) |
|---|---|---|---|
| Newton LR | 0.5021 | LEMstopalpha | -0.8824 |
| SGD SVM | 0.5000 | LEMstop | -0.9154 |
| L-BFGS LR | 0.4944 | LEMPOS | -0.8869 |
| Linear SVM | 0.4936 | LEMNERRSTOP | -0.9117 |
| NaiveBayes | 0.4819 | LEMNERALPHA | -0.8719 |
| AdaBoost | 0.4480 | LEMPOSSTOP | -0.7298 |
| XGBoost | 0.3846 | LEMPOSSTOPALPHA | -0.6749 |
| KNN | 0.3571 | LEMPOSSSTOPALPHA | -0.1470 |
| CNN1 | 0.3535 | LEMNERSTOPALPHA | -0.5461 |
| CNN2 | 0.3517 | TOKSTOP | -0.4314 |
| MLP | 0.3483 | LEMNERSTOP | -0.7447 |
| RandomForest | 0.2548 | LEMPOSSTOPALPHA | -0.4224 |

This range includes 37 of the 68 preprocessing methods (Table 5). This can be seen from, for example, the highest performing classifier, SVM with SGD optimizer (Figure 4), where the maximum classifier performance increases until peaking at .007 after which there is a noticeable drop. The performance only falls further as the FD rises. If the feature sets outside this range would be left out, the power savings are approximately 400Wh for the SGD SVM, when calculated from Table 6.

For CNNs however, there was no correlation between FD and the classifier performance, with the higher FD datasets performing almost equally when comparing to the low FD dataset, similarly as it was for English Cyberbullying Dataset, suggesting this could be a more general characteristics of the language itself. Taking a look at one layer CNN's performance, which was better than the CNN with two layers, we can see from Figure 4 that the performance stays quite stable throughout the whole range of feature densities. The scores are also generally much higher due to the classification problem (sentiment analysis of 1- and 5-start reviews) being much simpler compared to cyberbullying. All of the classifiers got considerably high results across feature densities without extreme changes, unlike other datasets. In other words, all of the classifiers seem to have reached their maximum performance for the data and there is no more score to gain without overfitting. Due to this, we are unable to estimate the possible time save by feature density in this case, meaning that for simple tasks such as this one it does not matter that much which classifier or which preprocessing is used, as it will still reach a considerably high performance.

Table 14: Verification Dataset: F1 for all preprocessing types & classifiers; best classifier for each dataset in **bold**; best preprocessing type for each underlined

| | LBFGS LR | SGD SVM | KNN | NaiveBayes | RandomForest | XGBoost | MLP | CNN1 | CNN2 |
|---|---|---|---|---|---|---|---|---|---|
| CHNK | 0.958 | 0.948 | 0.791 | 0.924 | 0.909 | 0.918 | 0.968 | **0.975** | 0.971 |
| CHNKNERR | 0.956 | 0.947 | 0.785 | 0.925 | 0.908 | 0.922 | 0.967 | **0.975** | 0.972 |
| CHNKNERRALPHA | 0.945 | 0.939 | 0.783 | 0.928 | 0.899 | 0.916 | 0.950 | **0.962** | 0.958 |
| CHNKNERRSTOP | 0.949 | 0.940 | 0.759 | 0.935 | 0.923 | 0.901 | **0.967** | 0.965 | 0.965 |
| CHNKNERRSTOPALPHA | 0.933 | 0.926 | 0.609 | 0.921 | 0.905 | 0.891 | **0.943** | 0.939 | 0.942 |
| CHNKNER | 0.957 | 0.948 | 0.783 | 0.923 | 0.909 | 0.922 | 0.967 | 0.977 | 0.972 |
| CHNKNERALPHA | 0.947 | 0.940 | 0.781 | 0.929 | 0.904 | 0.918 | 0.953 | **0.964** | 0.960 |
| CHNKNERSTOP | 0.950 | 0.939 | 0.767 | 0.933 | 0.923 | 0.902 | **0.968** | 0.965 | 0.965 |
| CHNKNERSTOPALPHA | 0.935 | 0.927 | 0.611 | 0.923 | 0.909 | 0.893 | **0.945** | 0.941 | 0.943 |
| CHNKALPHA | 0.947 | 0.940 | 0.786 | 0.930 | 0.902 | 0.916 | 0.953 | **0.962** | 0.959 |
| CHNKSTOP | 0.951 | 0.942 | 0.769 | 0.934 | 0.925 | 0.898 | 0.969 | 0.966 | 0.964 |
| CHNKSTOPALPHA | 0.935 | 0.927 | 0.609 | 0.924 | 0.908 | 0.893 | 0.944 | 0.939 | **0.944** |
| DEP | **0.948** | 0.927 | 0.777 | 0.940 | 0.900 | 0.878 | 0.920 | 0.933 | 0.933 |
| DEPNERR | **0.949** | 0.928 | 0.777 | 0.941 | 0.901 | 0.881 | 0.921 | 0.934 | 0.934 |
| DEPNERRALPHA | **0.945** | 0.926 | 0.773 | 0.941 | 0.899 | 0.869 | 0.924 | 0.934 | 0.934 |
| DEPNERRSTOP | 0.928 | 0.898 | 0.726 | 0.946 | 0.891 | 0.846 | 0.946 | **0.948** | 0.947 |
| DEPNERRSTOPALPHA | 0.916 | 0.887 | 0.732 | 0.935 | 0.885 | 0.824 | 0.943 | **0.945** | 0.942 |
| DEPNER | **0.946** | 0.922 | 0.774 | 0.940 | 0.898 | 0.881 | 0.923 | 0.935 | 0.933 |
| DEPNERALPHA | **0.943** | 0.921 | 0.032 | 0.940 | 0.897 | 0.876 | 0.926 | 0.935 | 0.934 |
| DEPNERSTOP | 0.924 | 0.889 | 0.729 | 0.945 | 0.893 | 0.847 | 0.944 | **0.950** | 0.945 |
| DEPNERSTOPALPHA | 0.914 | 0.881 | 0.155 | 0.934 | 0.884 | 0.831 | 0.944 | **0.945** | 0.944 |
| DEPALPHA | **0.944** | 0.926 | 0.773 | 0.940 | 0.898 | 0.869 | 0.926 | 0.935 | 0.935 |
| DEPSTOP | 0.927 | 0.895 | 0.726 | 0.945 | 0.889 | 0.842 | 0.947 | **0.949** | 0.944 |
| DEPSTOPALPHA | 0.915 | 0.888 | 0.735 | 0.934 | 0.885 | 0.815 | 0.945 | **0.946** | 0.942 |
| LEM | 0.968 | 0.963 | 0.768 | 0.924 | 0.934 | 0.948 | 0.966 | **0.974** | 0.970 |
| LEMNERR | 0.967 | 0.962 | 0.772 | 0.925 | 0.934 | 0.948 | 0.963 | **0.972** | 0.970 |
| LEMNERRALPHA | 0.966 | 0.961 | 0.774 | 0.924 | 0.934 | 0.947 | 0.964 | **0.973** | 0.969 |
| LEMNERRSTOP | 0.960 | 0.954 | 0.766 | 0.920 | 0.939 | 0.936 | **0.965** | 0.962 | 0.963 |
| LEMNERRSTOPALPHA | 0.961 | 0.955 | 0.766 | 0.920 | 0.939 | 0.936 | **0.964** | 0.963 | 0.963 |
| LEMPOSS | **0.968** | 0.962 | 0.744 | 0.925 | 0.932 | 0.947 | 0.944 | 0.958 | 0.958 |
| LEMPOSSALPHA | **0.967** | 0.962 | 0.764 | 0.925 | 0.932 | 0.946 | 0.948 | 0.958 | 0.956 |
| LEMPOSSSTOP | **0.962** | 0.955 | 0.740 | 0.921 | 0.938 | 0.933 | 0.962 | 0.962 | 0.958 |
| LEMPOSSSTOPALPHA | 0.962 | 0.955 | 0.760 | 0.920 | 0.940 | 0.932 | 0.962 | **0.963** | 0.960 |
| LEMNER | 0.968 | 0.963 | 0.765 | 0.922 | 0.935 | 0.948 | 0.963 | **0.973** | 0.971 |
| LEMNERALPHA | 0.967 | 0.962 | 0.763 | 0.922 | 0.934 | 0.947 | 0.964 | **0.973** | 0.969 |
| LEMNERSTOP | 0.961 | 0.955 | 0.763 | 0.918 | 0.939 | 0.938 | **0.967** | 0.962 | 0.963 |
| LEMNERSTOPALPHA | 0.962 | 0.956 | 0.761 | 0.918 | 0.938 | 0.937 | **0.965** | 0.963 | 0.963 |
| LEMPOS | **0.967** | 0.958 | 0.741 | 0.923 | 0.927 | 0.949 | 0.806 | 0.895 | 0.891 |
| LEMPOSALPHA | **0.966** | 0.959 | 0.737 | 0.923 | 0.928 | 0.948 | 0.818 | 0.898 | 0.894 |
| LEMPOSSTOP | **0.961** | 0.953 | 0.765 | 0.919 | 0.935 | 0.937 | 0.912 | 0.930 | 0.930 |
| LEMPOSSTOPALPHA | **0.961** | 0.954 | 0.769 | 0.919 | 0.938 | 0.936 | 0.940 | 0.944 | 0.948 |
| LEMALPHA | 0.967 | 0.963 | 0.765 | 0.923 | 0.935 | 0.947 | 0.966 | **0.972** | 0.968 |
| LEMSTOP | 0.962 | 0.956 | 0.763 | 0.919 | 0.940 | 0.937 | **0.967** | 0.963 | 0.964 |
| LEMSTOPALPHA | 0.962 | 0.956 | 0.762 | 0.918 | 0.941 | 0.936 | **0.965** | 0.963 | 0.963 |
| POSS | 0.744 | 0.742 | 0.658 | 0.690 | 0.747 | 0.750 | 0.668 | 0.836 | **0.841** |
| POSSALPHA | 0.744 | 0.749 | 0.658 | 0.690 | 0.746 | 0.750 | 0.668 | 0.838 | **0.843** |
| POSSSTOP | 0.723 | 0.729 | 0.613 | 0.691 | 0.715 | 0.728 | 0.666 | 0.777 | **0.778** |
| POSSSTOPALPHA | 0.712 | 0.713 | 0.504 | 0.675 | 0.695 | 0.718 | 0.600 | 0.755 | **0.764** |
| TOK | 0.970 | 0.965 | 0.791 | 0.928 | 0.934 | 0.949 | 0.967 | **0.974** | 0.971 |
| TOKNERR | 0.969 | 0.964 | 0.787 | 0.929 | 0.931 | 0.948 | 0.967 | **0.975** | 0.971 |
| TOKNERRALPHA | 0.969 | 0.964 | 0.787 | 0.928 | 0.935 | 0.948 | 0.966 | **0.974** | 0.970 |
| TOKNERRSTOP | 0.962 | 0.957 | 0.763 | 0.924 | 0.941 | 0.935 | **0.967** | 0.964 | 0.965 |
| TOKNERRSTOPALPHA | 0.962 | 0.957 | 0.763 | 0.923 | 0.940 | 0.936 | **0.968** | 0.964 | 0.965 |
| TOKPOSS | 0.970 | 0.965 | 0.755 | 0.930 | 0.931 | 0.946 | 0.947 | 0.960 | 0.960 |
| TOKPOSSALPHA | **0.963** | 0.955 | 0.771 | 0.924 | 0.939 | 0.939 | 0.916 | 0.931 | 0.933 |
| TOKPOSSSTOP | **0.970** | 0.965 | 0.783 | 0.929 | 0.933 | 0.946 | 0.951 | 0.961 | 0.957 |
| TOKPOSSSTOPALPHA | 0.963 | 0.957 | 0.730 | 0.927 | 0.941 | 0.933 | 0.963 | **0.965** | 0.959 |
| TOKNER | 0.964 | 0.958 | 0.760 | 0.927 | 0.942 | 0.933 | **0.967** | 0.964 | 0.962 |
| TOKNERALPHA | 0.970 | 0.965 | 0.783 | 0.926 | 0.935 | 0.950 | 0.966 | **0.974** | 0.972 |
| TOKNERSTOP | 0.970 | 0.965 | 0.781 | 0.925 | 0.936 | 0.949 | 0.966 | **0.974** | 0.971 |
| TOKNERSTOPALPHA | 0.963 | 0.957 | 0.765 | 0.923 | 0.941 | 0.937 | **0.968** | 0.963 | 0.965 |
| TOKPOS | 0.962 | 0.957 | 0.762 | 0.923 | 0.942 | 0.937 | **0.968** | 0.964 | 0.966 |
| TOKPOSALPHA | **0.969** | 0.962 | 0.752 | 0.927 | 0.929 | 0.951 | 0.806 | 0.893 | 0.892 |
| TOKPOSSTOP | **0.969** | 0.962 | 0.753 | 0.926 | 0.930 | 0.950 | 0.825 | 0.897 | 0.895 |
| TOKPOSSTOPALPHA | **0.963** | 0.956 | 0.761 | 0.925 | 0.938 | 0.938 | 0.916 | 0.931 | 0.933 |
| TOKALPHA | 0.969 | 0.964 | 0.787 | 0.927 | 0.937 | 0.949 | 0.968 | **0.974** | 0.971 |
| TOKSTOP | 0.964 | 0.958 | 0.765 | 0.923 | 0.943 | 0.936 | **0.968** | 0.963 | 0.966 |
| TOKSTOPALPHA | 0.963 | 0.958 | 0.764 | 0.923 | 0.944 | 0.937 | **0.968** | 0.964 | 0.965 |

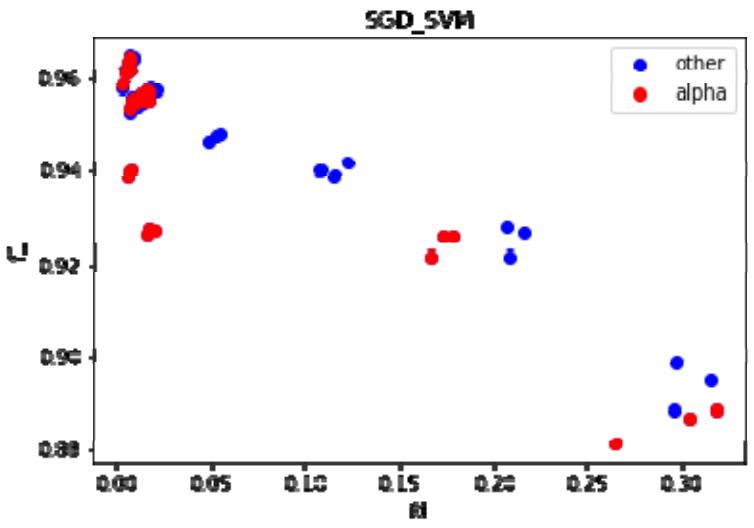

(a) Alphabetic filtering (red) vs others (blue)

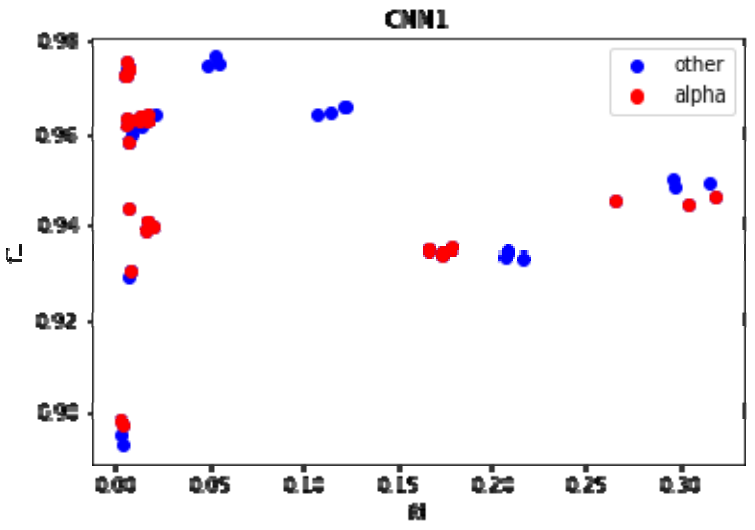

(b) Alphabetic filtering (red) vs others (blue)

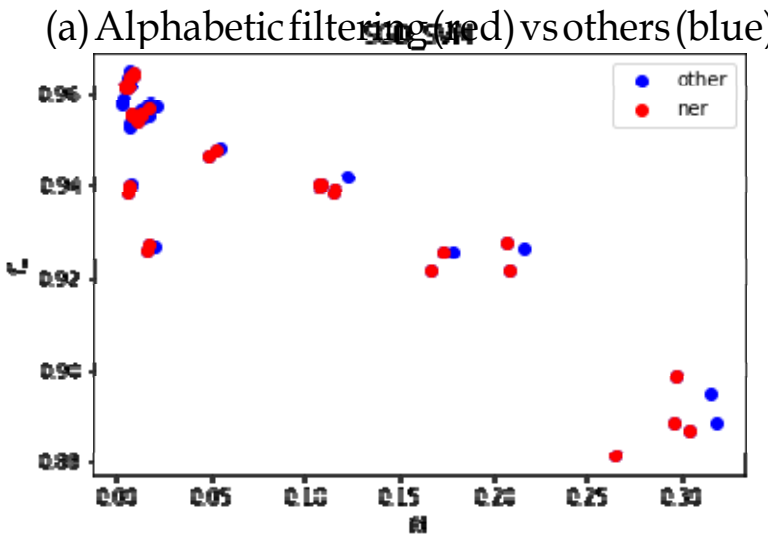

(c) NER (red) vs others (blue)

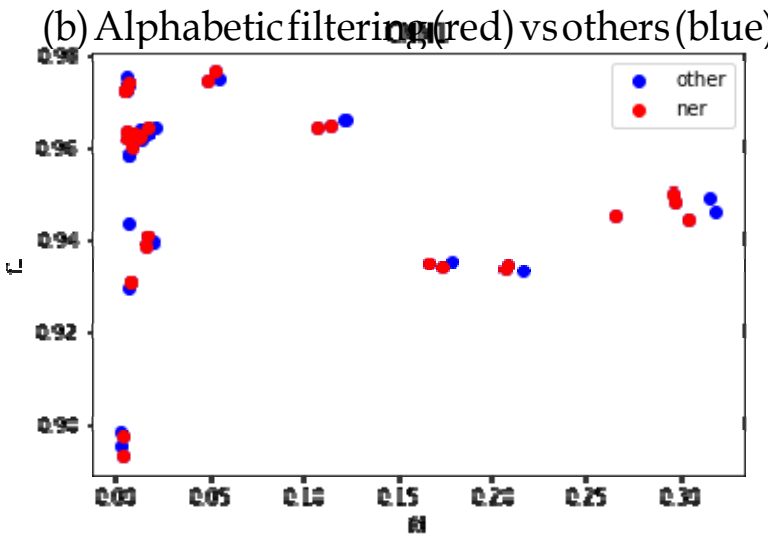

(d) NER (red) vs others (blue)

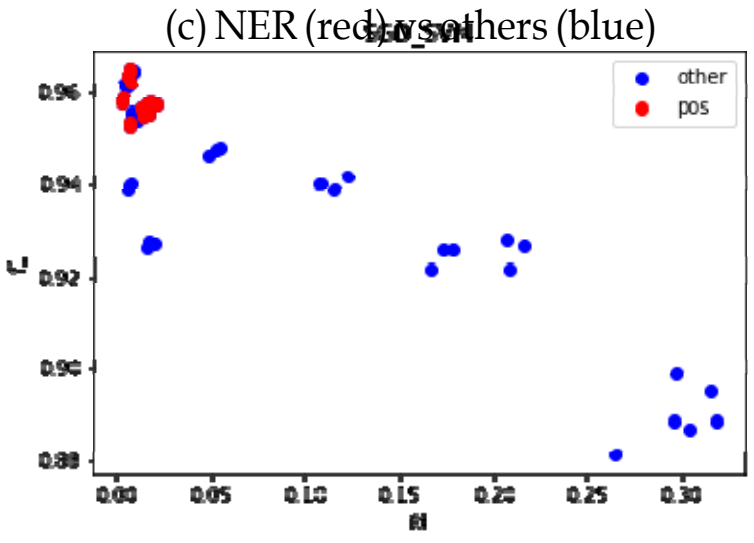

(e) POS (red) vs others (blue)

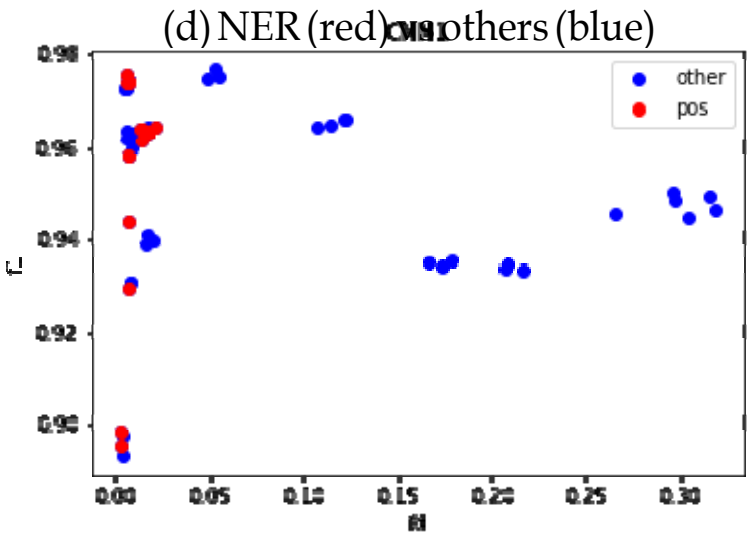

(f) POS (red) vs others (blue)

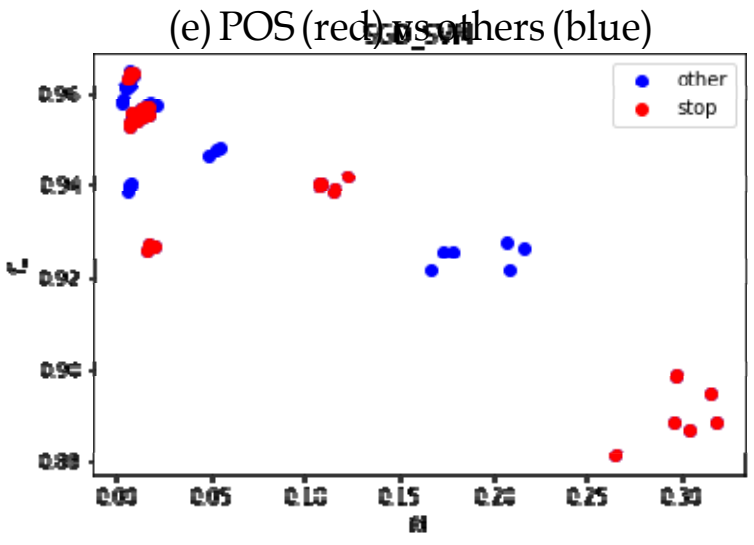

(g) Stopword filtering (red) vs others (blue)

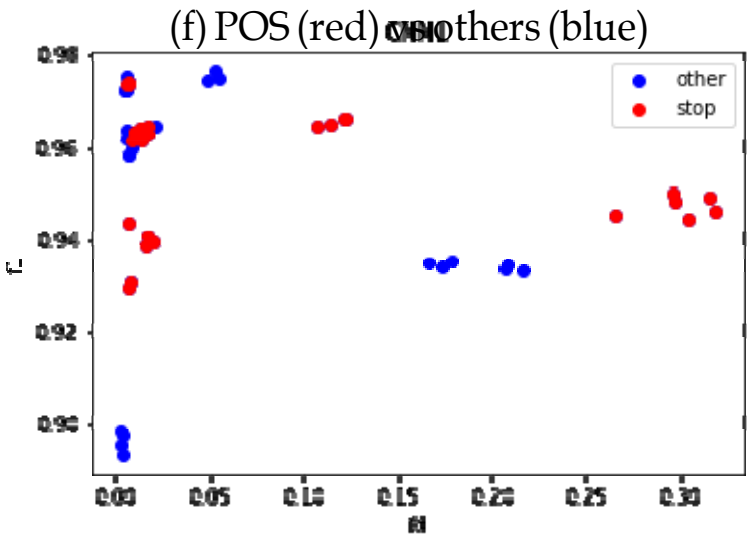

(h) Stopword filtering (red) vs others (blue)

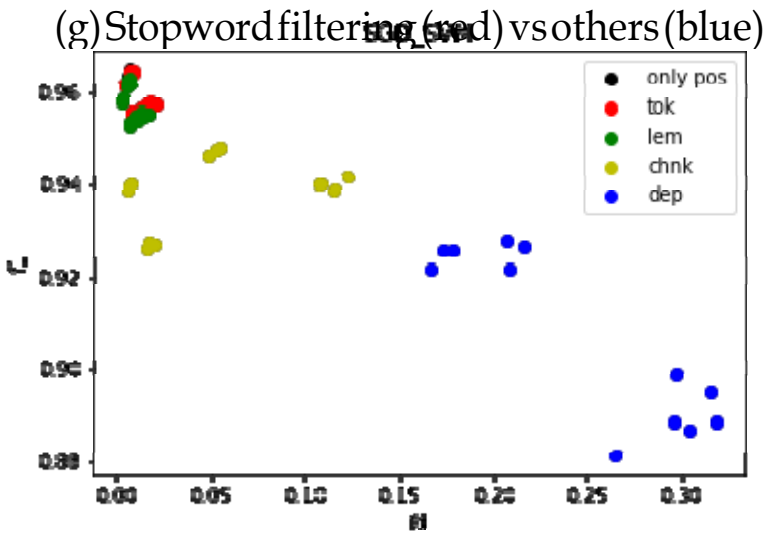

(i) Comparison of TOK (red), LEM (green), CHNK (yellow) and DEP (blue)

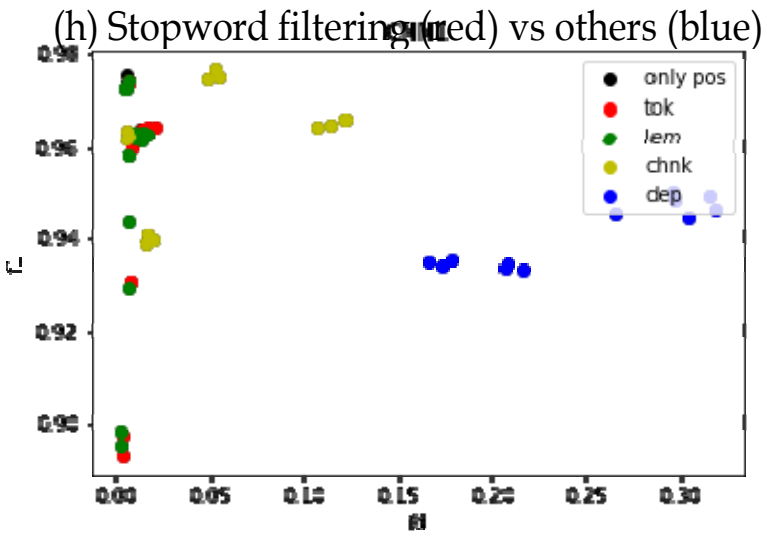

(j) Comparison of TOK (red), LEM (green), CHNK (yellow) and DEP (blue)

Figure 4: Verification Dataset: FD & F1 score for SGD SVM (left) and CNN1 (right)

Table 15: Verification Dataset: Classifiers with best F1, preprocessing type and Pearson's correlation coefficient for FD and F1.

| Classifier | Best F1 | Best PP type | $\rho$(F1, FD) |
|---|---|---|---|
| CNN1 | 0.9766 | CHNKNER | -0.2131 |
| CNN2 | 0.9723 | CHNKNERR | -0.2365 |
| L-BFGS LR | 0.9704 | TOKPOS | -0.8243 |
| MLP | 0.9689 | CHNKSTOP | -0.0410 |
| SGDSVM | 0.9650 | TOK | -0.9127 |
| XGBoost | 0.9506 | TOKPOSS | -0.9345 |
| NaiveBayes | 0.9456 | DEPNERSTOP | 0.8452 |
| RandomForest | 0.9435 | TOKSTOPALPHA | -0.8041 |
| KNN | 0.7911 | CHNK | -0.3000 |

### *5.3. Analysis of Linguistically-backed Preprocessing*

#### *5.3.1. English Cyberbullying Dataset*

From the results it can be seen that most of the classifiers scored highest or close to highest on pure tokens. CNNs also performed quite well on tokens, but also on dependency-based preprocessings. Using lemmas generally got slightly lower scores than tokens presumably due to information loss. Chunking got low performance overall and was clearly outperformed by dependency-based features in CNNs. Using only parts-of-speech tags achieved very low performance and thus it should be only used as a supplement to other preprocessing methods. However, as we can see that the simplest preprocessing, namely, tokens were usually among the highest scoring preprocessing type, this provides the first known empirical proof for using simple words as a default resource for embeddings in practically all previous research on word embeddings. However, it was still possible to outperform simple tokens by a number of preprocessing combinations.

Stopword filtering seemed to be the one of the most effective preprocessing techniques for traditional classifiers, which can be seen from Table 8 as it was used in the majority of the highest scores. The problem with stopword filtering was that the scores fluctuated a lot, having both low and very high scores and scoring high mostly with Logistic Regression and all of the tree based classifiers. An important thing to note is that the preprocessing method had extremely polarized performance with CNNs, scoring either very high or low. Overall, stopword filtering yielded the most top scores of any preprocessing method considering all the classifiers.

Another very effective preprocessing method was Parts of Speech merging (POS),

which achieved high performance overall when added to TOK or LEM. The method also got the highest scores with multiple classifiers, especially SVMs. Adding parts-of-speech information to the respective words usually achieved a higher score than using them as a separate feature. This keeps the information directly connected to the word itself, which seems a better option from the point of view of information preservation.

Using Named Entity Recognition information reduced the classifier performance most of the time, only achieving a high score with one classifier, Newton-LR. The performance of using NER seemed clearly inferior compared to stopword filtering or parts-of-speech information. Replacing words with their NER information seems to cause too much information loss and reduces the performance when comparing to plain tokens. Attaching NER information to the respective words did not improve the performance in most cases but still performed better than replacement.

These results contradict the results obtained in previous research [14], which noticed that NER helped most of the times for cyberbullying (CB) detection in Japanese. This could come from the fact that CB is differently realized in English and Japanese. In Japan, revealing victim's personal information, or "doxxing" is known to be one of the most often used forms of bullying, thus NER, which can pin-point information such as address or phone number often help in classification, while this is not the case for English.

Filtering out non-alphabetic characters also reduced the classifier performance most of the time and also got a high score with only one classifier, kNN, which was the weakest classifier overall. Non-alphabetic tokens, which include e.g., punctuation marks, ellipsis, question- of exclamation marks, seem to carry useful information, at least in the context of cyberbullying detection, as removing them reduced the performance comparing to plain tokens due to information loss.

Trying to generalize the feature set ended up lowering the results in most cases with the exception of the very high scores of stopword filtering using traditional classifiers. This would mean that the stopword filtering sometimes succeeded in removing noise and outliers from the dataset while other generalization methods ended up cutting useful information. Adding information to tokens could be useful in some scenarios as was shown with parts-of-speech tags and using dependency information with CNNs, although using NER was not so successful. Any kind of generalization attempt resulted in a lower performance with CNNs, which shows their ability to assemble more complex patterns from tokens and relations that are unusable by other classifiers.

### *5.3.2. Japanese Cyberbullying Dataset*

With the Japanese dataset, most classifiers scored the highest using lemmas, with tokens being not far behind. MLP and CNNs also performed decently on the dependency-based preprocessings. Again, chunking got low performance overall and got outperformed by dependency-based features in MLP and CNNs. The results further confirm that parts-of-speech tags alone are not viable features.

Using POS information as a supplement showed the highest positive impact on the results, increasing the performance most of the time, which can be seen from Table 10. Majority of the classifiers had their best result include POS. Adding POS information to the word itself seemed slightly more effective than using separate POS tags also with this dataset.

The rest of the preprocessing types showed mixed results depending on the classifier. Most notably, NER seemed to generally slightly increase the performance of Neural Network and tree based models while struggling with other models like Logistic Regression or SVM. These results are in line with previous research [14]. The effect of stopword filtering and alphabetic filtering was minor, usually only changing the results very little for better or worse depending on the preprocessing type.

Comparing to the English Cyberbullying Dataset, lemmas performed slightly better than tokens and the effectiveness of POS and NER were higher in the Japanese dataset. Whereas stopword filtering performed worse. What makes lemmas slightly better for Japanese could be due to Japanese being an agglutinative language. This increases the base token count, and thus feature density, in comparison to English. Slightly decreasing the number of distinct tokens by lemmatizing could be the key to achieving the optimal amount of information for classification of cyberbullying in Japanese.

The use of stopwords has been questioned with the Japanese language and even though some lists exist [67], it is much a more difficult preprocessing technique to use when comparing to English, where the use of stopwords is well established. This is because in Japanese each word/morpheme tends to contain useful information, whether syntactic or semantic, unlike in English, where it is easier to filter out words that do not affect the overall meaning of a sentence. The effect of alphabetic filtering was slightly better, yet it still performed average at best. Again, generalizing the dataset seemed to have an adverse effect on the performance most of the time.

### *5.3.3. Polish Cyberbullying Dataset*

Pure lemmas worked the best for Polish almost without exception. The performance was clearly superior compared to using tokens. This is completely different from the other datasets where tokens and lemmas only had slight differences in scores. Dependency based features performed very poorly for Polish language, even with Neural Networks.

What makes lemmas significantly better for Polish is most likely due to linguistic differences. Polish is a complex declinative language with a sophisticated morphology and grammar. This increases the base distinct token count and thus feature density. The base feature density for tokens is too high for classifiers to correctly generalize on the dataset of this size. Lemmatizing, which puts all variously declensed and conjugated words back into their dictionary forms, cuts the base feature density by about half, decreasing the complexity of the dataset, and thus greatly increasing the achieved score.

Similarly to the English dataset, stopword filtering performed well compared to other preprocessing techniques, which can be seen from Table 13 as it was used in the almost all of the highest scores. The difference being, that the scores were not fluctuating, and the plots (Figure 3) for stopwords look similar to the Japanese POS performance. Stopword filtering was clearly the most effective preprocessing method here and clearly increased the scores overall.

Other preprocessing methods had mixed results, slightly increasing or decreasing the results depending on the classifier and preprocessing type. This also includes POS, which clearly had a positive impact in both previous datasets. The performance of NER and alphabetic filtering were similar to the English dataset.

### *5.3.4. Verification Dataset*

Similarly to the English cyberbullying dataset, it can be seen that most of the traditional classifiers scored highest on pure tokens. The difference being that Neural Network based classifiers performed exceptionally well on chunks, as seen in Table 15. This might have something to do with the dataset size, as it is around twenty times larger compared to the others. Using lemmas generally got slightly lower scores than tokens, similarly to the English CB dataset. Also, dependency based preprocessings performed rather poorly and we were unable to confirm their effectiveness with CNNs given the larger dataset. However, the supplementary preprocessing methods were shown to be very ineffective and had next to no impact on the results. This is completely contrary to the other datasets as there was at least one supplementary preprocessing method that clearly increased the performance for all other datasets. The reason for this could be the simplicity of the learning problem. Most of the classifiers already scored very high, probably close to the maximum achievable score and thus the effects of extra preprocessing were mostly diminished. The effects could be more visible given a more difficult learning problem.

## *5.4. Classifier Stability*

In order to find the best feature sets, we calculated the standard deviations of all cross-validation folds for the English and Japanese datasets. Our aim is to find the high performing classifier-preprocessing pairs that are also stable, meaning their performance should have as low variance as possible within the cross-validation folds. We hypothesise that the well performing, stable classifier-preprocessing pairs would be similar across the datasets and could allow even more efficient model training. F-scores and standard errors for each pair are shown in Tables A.18 and A.19 in Appendix A.

We calculated the stability scores for Japanese and English datasets by substracting the standard error from the F-scores for each classifier-preprocessing pair. The results are shown in Tables 16 and 17. For the English dataset, the classifier-preprocessing pairs that got the highest stability scores are Logistic Regressions and SVMs with tokens and lemmas with parts of speech and their derivatives, highest being tokens with parts of speech and stopword filtering (TOKPOSSTOP).

Table 16: English dataset: Stability score for classifier-preprocessing pairs

| | LBFGS LR | Newton LR | Linear SVM | SGD SVM | KNN | NaiveBayes | RandomForest | AdaBoost | XGBoost | MLP | CNN1 | CNN2 | Avg for pp |
|---|---|---|---|---|---|---|---|---|---|---|---|---|---|
| CHNK | 0.493 | 0.484 | 0.461 | 0.499 | 0.18 | 0.427 | 0.247 | 0.351 | 0.378 | 0.414 | 0.401 | 0.388 | 0.394 |
| CHNKALPHA | 0.422 | 0.422 | 0.403 | 0.42 | 0.268 | 0.38 | 0.323 | 0.314 | 0.385 | 0.31 | 0.25 | 0.154 | 0.338 |
| CHNKNER | 0.48 | 0.487 | 0.473 | 0.503 | 0.197 | 0.421 | 0.224 | 0.331 | 0.381 | 0.404 | 0.458 | 0.355 | 0.393 |
| CHNKNERALPHA | 0.413 | 0.416 | 0.392 | 0.396 | 0.258 | 0.379 | 0.282 | 0.308 | 0.345 | 0.294 | 0.275 | 0.188 | 0.329 |
| CHNKNERR | 0.431 | 0.437 | 0.437 | 0.442 | 0.2 | 0.398 | 0.235 | 0.298 | 0.337 | 0.384 | 0.383 | 0.374 | 0.363 |
| CHNKNERRALPHA | 0.391 | 0.395 | 0.374 | 0.381 | 0.268 | 0.359 | 0.273 | 0.287 | 0.351 | 0.262 | 0.278 | 0.163 | 0.315 |
| CHNKNERRSTOP | 0.421 | 0.417 | 0.41 | 0.43 | 0.199 | 0.37 | 0.31 | 0.313 | 0.352 | 0.328 | 0.138 | 0.121 | 0.317 |
| CHNKNERRSTOPALPHA | 0.338 | 0.335 | 0.303 | 0.32 | 0.171 | 0.322 | 0.316 | 0.27 | 0.35 | 0.178 | 0.137 | -0.03 | 0.251 |
| CHNKNERSTOP | 0.483 | 0.483 | 0.461 | 0.483 | 0.193 | 0.412 | 0.296 | 0.361 | 0.394 | 0.424 | 0.301 | 0.103 | 0.366 |
| CHNKNERSTOPALPHA | 0.405 | 0.396 | 0.365 | 0.408 | 0.168 | 0.347 | 0.358 | 0.324 | 0.365 | 0.204 | 0.091 | 0.12 | 0.296 |
| CHNKSTOP | 0.481 | 0.482 | 0.455 | 0.484 | 0.198 | 0.423 | 0.359 | 0.358 | 0.4 | 0.437 | 0.129 | 0.146 | 0.363 |
| CHNKSTOPALPHA | 0.352 | 0.363 | 0.32 | 0.342 | 0.169 | 0.337 | 0.344 | 0.364 | 0.379 | 0.197 | 0.178 | 0.057 | 0.284 |
| DEP | 0.263 | 0.27 | 0.17 | 0.206 | 0.149 | 0.337 | 0.101 | 0.222 | 0.249 | 0.275 | 0.468 | 0.445 | 0.263 |
| DEPALPHA | 0.26 | 0.266 | 0.216 | 0.245 | 0.158 | 0.312 | 0.129 | 0.2 | 0.258 | 0.206 | 0.21 | 0.202 | 0.222 |
| DEPNER | 0.285 | 0.28 | 0.183 | 0.206 | 0.15 | 0.332 | 0.085 | 0.214 | 0.239 | 0.267 | 0.475 | 0.429 | 0.262 |
| DEPNERALPHA | 0.277 | 0.278 | 0.244 | 0.262 | 0.144 | 0.325 | 0.194 | 0.21 | 0.265 | 0.171 | 0.255 | 0.216 | 0.237 |
| DEPNERR | 0.26 | 0.273 | 0.182 | 0.206 | 0.147 | 0.328 | 0.107 | 0.212 | 0.239 | 0.261 | 0.443 | 0.396 | 0.255 |
| DEPNERRALPHA | 0.254 | 0.25 | 0.222 | 0.243 | 0.156 | 0.303 | 0.119 | 0.207 | 0.241 | 0.153 | 0.274 | 0.207 | 0.219 |
| DEPNERRSTOP | 0.244 | 0.239 | 0.169 | 0.175 | 0.148 | 0.324 | 0.122 | 0.222 | 0.246 | - | 0.501 | 0.453 | 0.258 |
| DEPNERRSTOPALPHA | 0.218 | 0.219 | 0.174 | 0.205 | 0.163 | 0.286 | 0.123 | 0.204 | 0.234 | 0.125 | 0.255 | 0.175 | 0.198 |
| DEPNERSTOP | 0.263 | 0.25 | 0.167 | 0.19 | 0.155 | 0.32 | 0.089 | 0.203 | 0.244 | 0.236 | 0.506 | 0.46 | 0.257 |
| DEPNERSTOPALPHA | 0.222 | 0.217 | 0.204 | 0.213 | 0.146 | 0.256 | 0.191 | 0.183 | 0.23 | 0.099 | 0.309 | 0.223 | 0.208 |
| DEPSTOP | 0.241 | 0.228 | 0.162 | 0.172 | 0.154 | 0.328 | 0.118 | 0.227 | 0.245 | 0.254 | 0.532 | 0.377 | 0.253 |
| DEPSTOPALPHA | 0.215 | 0.222 | 0.178 | 0.201 | 0.167 | 0.285 | 0.124 | 0.205 | 0.23 | 0.132 | 0.277 | 0.239 | 0.206 |
| LEM | 0.601 | 0.599 | 0.594 | 0.608 | 0.308 | 0.494 | 0.471 | 0.498 | 0.519 | 0.528 | 0.477 | 0.4 | 0.508 |
| LEMALPHA | 0.56 | 0.566 | 0.541 | 0.567 | 0.115 | 0.478 | 0.474 | 0.463 | 0.526 | 0.468 | 0.402 | 0.368 | 0.461 |
| LEMNER | 0.597 | 0.599 | 0.605 | 0.616 | 0.301 | 0.486 | 0.455 | 0.483 | 0.515 | 0.506 | 0.49 | 0.345 | 0.5 |
| LEMNERALPHA | 0.566 | 0.568 | 0.562 | 0.573 | 0.322 | 0.472 | 0.455 | 0.46 | 0.504 | 0.469 | 0.459 | 0.387 | 0.483 |
| LEMNERR | 0.523 | 0.518 | 0.525 | 0.526 | 0.266 | 0.458 | 0.417 | 0.423 | 0.478 | 0.463 | 0.455 | 0.432 | 0.457 |
| LEMNERRALPHA | 0.507 | 0.506 | 0.498 | 0.503 | 0.293 | 0.448 | 0.417 | 0.42 | 0.466 | 0.473 | 0.368 | 0.363 | 0.439 |
| LEMNERRSTOP | 0.517 | 0.513 | 0.498 | 0.512 | 0.28 | 0.447 | 0.482 | 0.432 | 0.478 | 0.398 | 0.065 | 0.15 | 0.398 |
| LEMNERRSTOPALPHA | 0.51 | 0.507 | 0.478 | 0.5 | 0.301 | 0.435 | 0.474 | 0.427 | 0.457 | 0.427 | 0.158 | 0.081 | 0.396 |
| LEMNERSTOP | 0.597 | 0.596 | 0.596 | 0.613 | 0.304 | 0.481 | 0.515 | 0.494 | 0.52 | 0.482 | 0.231 | 0.163 | 0.466 |
| LEMNERSTOPALPHA | 0.575 | 0.572 | 0.551 | 0.567 | 0.336 | 0.467 | 0.508 | 0.479 | 0.505 | 0.425 | 0.25 | 0.198 | 0.453 |
| LEMPOS | 0.574 | 0.58 | 0.586 | 0.588 | 0.152 | 0.489 | 0.353 | 0.463 | 0.497 | 0.532 | 0.471 | 0.376 | 0.472 |
| LEMPOSALPHA | 0.564 | 0.562 | 0.561 | 0.559 | 0.312 | 0.472 | 0.356 | 0.441 | 0.475 | 0.435 | 0.451 | 0.445 | 0.469 |
| LEMPOSS | 0.562 | 0.558 | 0.564 | 0.565 | 0.205 | 0.486 | 0.388 | 0.428 | 0.468 | 0.438 | 0.4 | 0.465 | 0.461 |
| LEMPOSSALPHA | 0.56 | 0.554 | 0.54 | 0.549 | 0.152 | 0.475 | 0.41 | 0.408 | 0.463 | 0.45 | 0.386 | 0.402 | 0.446 |
| LEMPOSSSTOP | 0.574 | 0.579 | 0.581 | 0.597 | 0.27 | 0.49 | 0.379 | 0.466 | 0.49 | 0.482 | 0.462 | 0.491 | 0.488 |
| LEMPOSSSTOPALPHA | 0.564 | 0.573 | 0.554 | 0.575 | 0.193 | 0.475 | 0.399 | 0.445 | 0.476 | 0.461 | 0.33 | 0.396 | 0.453 |
| LEMPOSSTOP | 0.593 | 0.59 | 0.605 | 0.602 | 0.357 | 0.486 | 0.447 | 0.489 | 0.504 | 0.43 | 0.455 | 0.437 | 0.5 |
| LEMPOSSTOPALPHA | 0.577 | 0.574 | 0.564 | 0.577 | 0.387 | 0.469 | 0.462 | 0.481 | 0.491 | 0.484 | 0.404 | 0.471 | 0.495 |
| LEMSTOP | 0.604 | 0.601 | 0.592 | 0.61 | 0.321 | 0.492 | 0.525 | 0.504 | 0.533 | 0.468 | 0.206 | 0.122 | 0.465 |
| LEMSTOPALPHA | 0.582 | 0.574 | 0.57 | 0.58 | 0.154 | 0.475 | 0.527 | 0.476 | 0.525 | 0.437 | 0.165 | 0.119 | 0.432 |
| POSS | 0.196 | 0.197 | 0.195 | 0.199 | 0.154 | 0.192 | 0.164 | 0.179 | 0.176 | 0.007 | 0.059 | 0.055 | 0.148 |
| POSSALPHA | 0.198 | 0.196 | 0.194 | 0.208 | 0.169 | 0.192 | 0.156 | 0.187 | 0.168 | 0.012 | 0.057 | 0.053 | 0.149 |
| POSSSTOP | 0.19 | 0.19 | 0.189 | 0.184 | 0.129 | 0.189 | 0.152 | 0.179 | 0.164 | 0.003 | -0.05 | -0.047 | 0.123 |
| POSSSTOPALPHA | 0.186 | 0.186 | 0.183 | 0.185 | 0.11 | 0.181 | 0.149 | 0.181 | 0.157 | 0.01 | -0.053 | 0.003 | 0.123 |
| TOK | 0.611 | 0.606 | 0.611 | 0.623 | 0.295 | 0.495 | 0.439 | 0.494 | 0.525 | 0.498 | 0.468 | 0.403 | 0.506 |
| TOKALPHA | 0.571 | 0.575 | 0.555 | 0.582 | 0.096 | 0.481 | 0.46 | 0.459 | 0.518 | 0.474 | 0.447 | 0.342 | 0.463 |
| TOKNER | 0.604 | 0.606 | 0.606 | 0.606 | 0.265 | 0.483 | 0.432 | 0.483 | 0.519 | 0.496 | 0.489 | 0.44 | 0.502 |
| TOKNERALPHA | 0.574 | 0.577 | 0.563 | 0.587 | 0.303 | 0.466 | 0.431 | 0.467 | 0.517 | 0.466 | 0.447 | 0.391 | 0.482 |
| TOKNERR | 0.526 | 0.526 | 0.518 | 0.527 | 0.263 | 0.465 | 0.407 | 0.408 | 0.473 | 0.458 | 0.455 | 0.356 | 0.449 |
| TOKNERRALPHA | 0.519 | 0.515 | 0.507 | 0.513 | 0.304 | 0.448 | 0.401 | 0.41 | 0.447 | 0.441 | 0.411 | 0.385 | 0.442 |
| TOKNERRSTOP | 0.516 | 0.514 | 0.504 | 0.509 | 0.276 | 0.441 | 0.486 | 0.418 | 0.457 | 0.448 | 0.194 | 0.092 | 0.405 |
| TOKNERRSTOPALPHA | 0.503 | 0.509 | 0.49 | 0.494 | 0.304 | 0.432 | 0.472 | 0.42 | 0.444 | 0.387 | 0.19 | 0.164 | 0.401 |
| TOKNERSTOP | 0.599 | 0.605 | 0.604 | 0.609 | 0.311 | 0.475 | 0.502 | 0.491 | 0.521 | 0.465 | 0.239 | 0.158 | 0.465 |
| TOKNERSTOPALPHA | 0.576 | 0.58 | 0.554 | 0.581 | 0.337 | 0.458 | 0.501 | 0.463 | 0.514 | 0.441 | 0.109 | 0.162 | 0.44 |
| TOKPOS | 0.582 | 0.584 | 0.588 | 0.605 | 0.187 | 0.491 | 0.343 | 0.45 | 0.484 | 0.443 | 0.425 | 0.478 | 0.472 |
| TOKPOSALPHA | 0.572 | 0.569 | 0.564 | 0.576 | 0.385 | 0.476 | 0.349 | 0.428 | 0.469 | 0.455 | 0.365 | 0.448 | 0.471 |
| TOKPOSS | 0.563 | 0.568 | 0.566 | 0.588 | 0.177 | 0.491 | 0.38 | 0.417 | 0.466 | 0.441 | 0.482 | 0.498 | 0.47 |
| TOKPOSSALPHA | 0.57 | 0.565 | 0.567 | 0.575 | 0.101 | 0.481 | 0.404 | 0.415 | 0.468 | 0.351 | 0.374 | 0.383 | 0.438 |
| TOKPOSSSTOP | 0.579 | 0.581 | 0.585 | 0.599 | 0.256 | 0.481 | 0.386 | 0.44 | 0.495 | 0.417 | 0.47 | 0.495 | 0.482 |
| TOKPOSSSTOPALPHA | 0.582 | 0.58 | 0.585 | 0.585 | 0.369 | 0.471 | 0.404 | 0.471 | 0.497 | 0.451 | 0.415 | 0.46 | 0.489 |
| TOKPOSSTOP | 0.599 | 0.597 | 0.612 | 0.613 | 0.359 | 0.475 | 0.449 | 0.503 | 0.507 | 0.451 | 0.506 | 0.465 | 0.511 |
| TOKPOSSTOPALPHA | - | - | - | - | - | - | - | - | - | - | - | - | - |
| TOKSTOP | 0.613 | 0.608 | 0.596 | 0.614 | 0.317 | 0.486 | 0.542 | 0.509 | 0.531 | 0.419 | 0.156 | 0.114 | 0.459 |
| TOKSTOPALPHA | 0.587 | 0.586 | 0.567 | 0.584 | 0.148 | 0.472 | 0.528 | 0.483 | 0.529 | 0.428 | 0.237 | 0.215 | 0.447 |
| Avgforclf | 0.463 | 0.463 | 0.445 | 0.459 | 0.228 | 0.411 | 0.342 | 0.372 | 0.407 | 0.36 | 0.321 | 0.285 | |

Table 17: Japanese dataset: Stability score for classifier-preprocessing pairs

| | LBFGS LR | Newton LR | Linear SVM | SGD SVM | KNN | NaiveBayes | RandomForest | AdaBoost | XGBoost | MLP | CNN1 | CNN2 | Avg for pp |
|---|---|---|---|---|---|---|---|---|---|---|---|---|---|
| CHNK | 0.67 | 0.663 | 0.677 | 0.668 | 0.289 | 0.72 | 0.623 | 0.47 | 0.541 | 0.732 | 0.745 | 0.609 | 0.617 |
| CHNKALPHA | 0.686 | 0.69 | 0.678 | 0.662 | 0.331 | 0.713 | 0.641 | 0.47 | 0.533 | 0.715 | 0.718 | 0.6 | 0.62 |
| CHNKNER | 0.716 | 0.721 | 0.735 | 0.732 | 0.349 | 0.741 | 0.699 | 0.639 | 0.673 | 0.777 | 0.75 | - | 0.685 |
| CHNKNERALPHA | 0.72 | 0.725 | 0.726 | 0.717 | 0.396 | 0.738 | 0.685 | 0.627 | 0.669 | 0.738 | 0.743 | - | 0.68 |
| CHNKNERR | 0.722 | 0.721 | 0.724 | 0.707 | 0.366 | 0.727 | 0.683 | 0.631 | 0.667 | 0.752 | 0.766 | 0.667 | 0.678 |
| CHNKNERRALPHA | 0.687 | 0.696 | 0.709 | 0.7 | 0.366 | 0.722 | 0.678 | 0.615 | 0.649 | 0.717 | 0.743 | 0.632 | 0.66 |
| CHNKNERRSTOP | 0.709 | 0.704 | 0.727 | 0.699 | 0.359 | 0.699 | 0.687 | 0.637 | 0.651 | 0.758 | 0.749 | 0.642 | 0.668 |
| CHNKNERRSTOPALPHA | 0.689 | 0.682 | 0.705 | 0.686 | 0.379 | 0.704 | 0.665 | 0.607 | 0.628 | 0.708 | 0.695 | 0.63 | 0.648 |
| CHNKNERSTOP | 0.714 | 0.715 | 0.731 | 0.752 | 0.383 | 0.738 | 0.699 | 0.643 | 0.662 | 0.757 | 0.766 | - | 0.687 |
| CHNKNERSTOPALPHA | 0.721 | 0.723 | 0.728 | 0.734 | 0.357 | 0.728 | 0.676 | 0.639 | 0.659 | 0.715 | 0.733 | - | 0.674 |
| CHNKSTOP | 0.661 | 0.661 | 0.674 | 0.671 | 0.306 | 0.682 | 0.608 | 0.457 | 0.537 | 0.73 | 0.716 | 0.667 | 0.614 |
| CHNKSTOPALPHA | 0.646 | 0.651 | 0.666 | 0.651 | 0.327 | 0.687 | 0.595 | 0.439 | 0.539 | 0.674 | 0.699 | 0.571 | 0.595 |
| DEP | 0.609 | 0.611 | 0.619 | 0.625 | 0.307 | 0.621 | 0.484 | 0.358 | 0.456 | 0.737 | 0.736 | 0.546 | 0.559 |
| DEPALPHA | 0.595 | 0.594 | 0.614 | 0.609 | 0.27 | 0.613 | 0.456 | 0.35 | 0.444 | 0.723 | 0.686 | 0.588 | 0.545 |
| DEPNER | 0.674 | 0.682 | 0.707 | 0.713 | 0.331 | 0.684 | 0.649 | 0.625 | 0.658 | 0.765 | 0.727 | - | 0.656 |
| DEPNERALPHA | 0.684 | 0.683 | 0.709 | 0.712 | 0.179 | 0.675 | 0.642 | 0.631 | 0.658 | 0.766 | 0.73 | - | 0.643 |
| DEPNERR | 0.612 | 0.617 | 0.615 | 0.625 | 0.279 | 0.616 | 0.486 | 0.336 | 0.45 | 0.756 | 0.747 | 0.553 | 0.558 |
| DEPNERRALPHA | 0.593 | 0.594 | 0.609 | 0.604 | 0.274 | 0.615 | 0.448 | 0.342 | 0.448 | 0.735 | 0.735 | 0.553 | 0.546 |
| DEPNERRSTOP | 0.575 | 0.569 | 0.573 | 0.567 | 0.343 | 0.57 | 0.419 | 0.324 | 0.397 | 0.765 | 0.736 | 0.445 | 0.524 |
| DEPNERRSTOPALPHA | 0.509 | 0.508 | 0.51 | 0.516 | 0.326 | 0.528 | 0.354 | 0.309 | 0.348 | 0.73 | 0.698 | 0.572 | 0.492 |
| DEPNERSTOP | 0.662 | 0.663 | 0.681 | 0.689 | 0.414 | 0.66 | 0.631 | 0.611 | 0.628 | 0.776 | 0.727 | - | 0.649 |
| DEPNERSTOPALPHA | 0.665 | 0.661 | 0.683 | 0.672 | 0.211 | 0.699 | 0.632 | 0.617 | 0.639 | 0.757 | 0.719 | - | 0.632 |
| DEPSTOP | 0.572 | 0.579 | 0.576 | 0.577 | 0.341 | 0.573 | 0.427 | 0.34 | 0.405 | 0.748 | 0.703 | 0.619 | 0.538 |
| DEPSTOPALPHA | 0.518 | 0.518 | 0.522 | 0.528 | 0.33 | 0.535 | 0.369 | 0.327 | 0.363 | 0.718 | 0.716 | 0.579 | 0.502 |
| LEM | 0.783 | 0.777 | 0.8 | 0.776 | 0.477 | 0.803 | 0.739 | 0.589 | 0.683 | 0.846 | 0.837 | 0.692 | 0.734 |
| LEMALPHA | 0.779 | 0.776 | 0.8 | 0.785 | 0.472 | 0.793 | 0.728 | 0.586 | 0.682 | 0.839 | 0.831 | 0.784 | 0.738 |
| LEMNER | 0.778 | 0.769 | 0.795 | 0.794 | 0.523 | 0.794 | 0.768 | 0.705 | 0.74 | 0.853 | 0.845 | 0.595 | 0.747 |
| LEMNERALPHA | 0.781 | 0.785 | 0.807 | 0.809 | 0.51 | 0.798 | 0.771 | 0.697 | 0.746 | 0.838 | 0.829 | 0.744 | 0.76 |
| LEMNERR | 0.765 | 0.758 | 0.777 | 0.767 | 0.468 | 0.786 | 0.752 | 0.7 | 0.726 | 0.825 | 0.79 | 0.632 | 0.729 |
| LEMNERRALPHA | 0.77 | 0.775 | 0.778 | 0.772 | 0.471 | 0.787 | 0.749 | 0.708 | 0.741 | 0.816 | 0.81 | 0.686 | 0.739 |
| LEMNERRSTOP | 0.763 | 0.756 | 0.77 | 0.769 | 0.469 | 0.769 | 0.772 | 0.703 | 0.717 | 0.82 | 0.82 | 0.791 | 0.743 |
| LEMNERRSTOPALPHA | 0.766 | 0.764 | 0.772 | 0.768 | 0.457 | 0.769 | 0.757 | 0.702 | 0.735 | 0.788 | 0.8 | 0.75 | 0.736 |
| LEMNERSTOP | 0.769 | 0.773 | 0.791 | 0.802 | 0.533 | 0.805 | 0.774 | 0.71 | 0.731 | 0.856 | 0.858 | 0.784 | 0.766 |
| LEMNERSTOPALPHA | 0.78 | 0.779 | 0.806 | 0.806 | 0.543 | 0.82 | 0.783 | 0.706 | 0.729 | 0.826 | 0.815 | 0.71 | 0.759 |
| LEMPOS | 0.785 | 0.786 | 0.83 | 0.827 | 0.677 | 0.816 | 0.762 | 0.691 | 0.742 | 0.847 | 0.845 | 0.595 | 0.767 |
| LEMPOSALPHA | 0.781 | 0.768 | 0.821 | 0.827 | 0.674 | 0.802 | 0.741 | 0.675 | 0.739 | 0.839 | 0.845 | 0.647 | 0.763 |
| LEMPOSS | 0.792 | 0.8 | 0.83 | 0.825 | 0.533 | 0.83 | 0.752 | 0.671 | 0.747 | 0.847 | - | 0.755 | 0.762 |
| LEMPOSSALPHA | 0.805 | 0.793 | 0.83 | 0.818 | 0.699 | 0.827 | 0.731 | 0.626 | 0.718 | 0.832 | 0.843 | 0.544 | 0.756 |
| LEMPOSSSTOP | 0.791 | 0.79 | 0.832 | 0.829 | 0.491 | 0.833 | 0.743 | 0.688 | 0.742 | 0.845 | 0.856 | 0.796 | 0.77 |
| LEMPOSSSTOPALPHA | 0.783 | 0.792 | 0.816 | 0.813 | 0.685 | 0.822 | 0.745 | 0.661 | 0.713 | 0.835 | 0.822 | 0.673 | 0.763 |
| LEMPOSSTOP | 0.779 | 0.766 | 0.826 | 0.824 | 0.423 | 0.815 | 0.74 | 0.696 | 0.751 | 0.846 | 0.85 | 0.783 | 0.758 |
| LEMPOSSTOPALPHA | 0.774 | 0.765 | 0.804 | 0.817 | 0.672 | 0.815 | 0.733 | 0.664 | 0.733 | 0.817 | 0.832 | 0.752 | 0.765 |
| LEMSTOP | 0.771 | 0.775 | 0.794 | 0.778 | 0.482 | 0.78 | 0.755 | 0.577 | 0.669 | 0.84 | 0.847 | 0.792 | 0.738 |
| LEMSTOPALPHA | 0.772 | 0.774 | 0.788 | 0.776 | 0.487 | 0.796 | 0.744 | 0.587 | 0.668 | 0.817 | 0.823 | 0.774 | 0.734 |
| POSS | 0.607 | 0.616 | 0.595 | 0.581 | 0.507 | 0.585 | 0.604 | 0.614 | 0.608 | 0.621 | 0.638 | 0.458 | 0.586 |
| POSSALPHA | 0.61 | 0.611 | 0.584 | 0.552 | 0.502 | 0.586 | 0.606 | 0.616 | 0.611 | 0.613 | 0.624 | 0.537 | 0.588 |
| POSSSTOP | 0.607 | 0.627 | 0.604 | 0.587 | 0.47 | 0.588 | 0.616 | 0.615 | 0.599 | 0.607 | 0.599 | 0.462 | 0.582 |
| POSSSTOPALPHA | 0.581 | 0.582 | 0.54 | 0.558 | 0.381 | 0.569 | 0.575 | 0.576 | 0.562 | 0.567 | 0.537 | 0.455 | 0.54 |
| TOK | 0.777 | 0.775 | 0.79 | 0.785 | 0.448 | 0.795 | 0.737 | 0.573 | 0.673 | 0.845 | 0.833 | 0.779 | 0.734 |
| TOKALPHA | 0.781 | 0.774 | 0.788 | 0.784 | 0.449 | 0.798 | 0.73 | 0.578 | 0.66 | 0.827 | 0.834 | 0.78 | 0.732 |
| TOKNER | 0.786 | 0.785 | 0.812 | 0.808 | 0.6 | 0.817 | 0.732 | 0.63 | 0.699 | 0.828 | 0.825 | 0.662 | 0.749 |
| TOKNERALPHA | 0.773 | 0.779 | 0.801 | 0.794 | 0.493 | 0.807 | 0.762 | 0.702 | 0.724 | 0.846 | 0.842 | 0.804 | 0.761 |
| TOKNERR | 0.758 | 0.756 | 0.78 | 0.783 | 0.463 | 0.778 | 0.762 | 0.692 | 0.716 | 0.81 | 0.828 | 0.771 | 0.741 |
| TOKNERRALPHA | 0.769 | 0.765 | 0.778 | 0.768 | 0.441 | 0.778 | 0.749 | 0.703 | 0.725 | 0.815 | 0.803 | 0.696 | 0.733 |
| TOKNERRSTOP | 0.744 | 0.746 | 0.767 | 0.762 | 0.453 | 0.764 | 0.751 | 0.688 | 0.722 | 0.815 | 0.829 | 0.682 | 0.727 |
| TOKNERRSTOPALPHA | 0.761 | 0.752 | 0.774 | 0.77 | 0.444 | 0.751 | 0.749 | 0.696 | 0.732 | 0.802 | 0.794 | 0.723 | 0.729 |
| TOKNERSTOP | 0.777 | 0.774 | 0.803 | 0.798 | 0.494 | 0.808 | 0.759 | 0.7 | 0.735 | 0.827 | 0.83 | 0.792 | 0.758 |
| TOKNERSTOPALPHA | 0.764 | 0.763 | 0.784 | 0.805 | 0.5 | 0.801 | 0.766 | 0.696 | 0.726 | 0.85 | 0.848 | 0.808 | 0.759 |
| TOKPOS | 0.779 | 0.777 | 0.801 | 0.812 | 0.607 | 0.813 | 0.767 | 0.688 | 0.727 | 0.821 | 0.825 | 0.742 | 0.763 |
| TOKPOSALPHA | 0.793 | 0.785 | 0.819 | 0.83 | 0.665 | 0.81 | 0.737 | 0.668 | 0.74 | 0.855 | 0.846 | 0.655 | 0.767 |
| TOKPOSS | 0.79 | 0.798 | 0.821 | 0.812 | 0.504 | 0.831 | 0.75 | 0.662 | 0.725 | 0.845 | - | 0.662 | 0.745 |
| TOKPOSSALPHA | 0.756 | 0.761 | 0.807 | 0.819 | 0.694 | 0.808 | 0.707 | 0.647 | 0.7 | 0.845 | 0.845 | 0.632 | 0.752 |
| TOKPOSSSTOP | 0.798 | 0.79 | 0.807 | 0.814 | 0.677 | 0.826 | 0.738 | 0.676 | 0.739 | 0.833 | 0.8 | 0.621 | 0.76 |
| TOKPOSSSTOPALPHA | 0.794 | 0.803 | 0.816 | 0.822 | 0.477 | 0.826 | 0.728 | 0.674 | 0.743 | 0.845 | 0.863 | 0.74 | 0.761 |
| TOKPOSSTOP | 0.764 | 0.773 | 0.82 | 0.821 | 0.557 | 0.803 | 0.739 | 0.685 | 0.738 | 0.825 | 0.835 | 0.601 | 0.747 |
| TOKPOSSTOPALPHA | - | 0.765 | 0.813 | 0.821 | 0.527 | 0.817 | 0.732 | 0.676 | 0.73 | - | - | - | 0.735 |
| TOKSTOP | 0.776 | 0.774 | 0.787 | 0.778 | 0.451 | 0.786 | 0.737 | 0.568 | 0.662 | 0.844 | 0.848 | 0.759 | 0.731 |
| TOKSTOPALPHA | 0.774 | 0.768 | 0.787 | 0.794 | 0.459 | 0.802 | 0.752 | 0.57 | 0.659 | 0.811 | 0.797 | 0.66 | 0.719 |
| Avgforclf | 0.721 | 0.721 | 0.739 | 0.736 | 0.453 | 0.741 | 0.68 | 0.603 | 0.653 | 0.785 | 0.779 | 0.665 | |

The same preprocessing types, tokens and lemmas with parts of speech and their derivatives, also had high scores for Japanese with logistic regressions and SVMs, although the actual highest performing classifiers were Neural Networks (CNN1, MLP) instead. Tokens and lemmas with named entity recognition also scored high. Based on these results we can say that parts of speech information added to tokens or lemmas should be the preferred feature sets when training as they increase the performance of the most effective classifiers compared to plain tokens or lemmas.

## 6. General Discussion

Overall, the highest performing classifiers were one-layer CNN for Japanese and Verification datasets, SVM for English and Logistic Regression for the Polish dataset. The baseline classifiers, KNN and Naive Bayes, and tree based methods, Random forest, AdaBoost, XGBoost, had mediocre results at best. Using only POS tags as features had extremely low performance with all of the datasets, so they were excluded from the analysis.

An interesting discovery is that using plain tokens rarely had the best performance out of the proposed feature sets as can be seen from Tables 8, 10, 13 and 15. This proves the effectiveness of linguistics-based feature engineering instead of using just words as features. For example, the one-layer CNN's performance with English increased from 0.659 (TOK) F-score to 0.741 (DEPSTOP), indicating that information about structure seems important.

The fact that feature sets with linguistic information managed to outperform the use of tokens clearly shows their potential in feature engineering. for some models, like the one-layer CNN with the English dataset, the increase was truly significant, almost 0.1 F-score using linguistic embeddings. We also discovered that adding parts of speech information to tokens or lemmas produces the most stable and high performance feature sets and they should be preferred over plain tokens or lemmas. This discovery also raises a question: how would linguistic preprocessing affect state-of-the-art pretrained models? For example, RoBERTa [80] fine tuned on the English dataset shows an F-score of 0.797, which is similar to the highest scores by other models. Actually, the best score by SGD SVM is 0.798 which is slightly higher.

It is interesting that a simple method like SVM can outperform a complex modern text classifier when using the right feature set. This shows that they should not be underestimated as with correct preparations, they can reach similar performance to state-of-the art models with much less computational power required. Perhaps the performance of RoBERTa could also be increased by feature engineering and applying embeddings with linguistic information. This needs to be explored further in future research as there is clearly a potential point of improvement.

### 6.1. Feature Density

In general, most of the classifier performances, with the exception of CNNs, had a strong negative correlation with FD. So these classifiers seem to have a weaker performance if a lot of linguistic information is added. These results suggest that for non-CNN classifiers there is no need to consider preprocessings with a high FD, such as chunking or dependencies, as they had a considerably lower performance. The best performance was in most cases between 50-200% of the base FD (TOK) depending on the dataset and classifier.

For CNNs we were expecting a positive correlation with FD, at least with the Japanese dataset, considering the results of the previous research [14]. However, the results were different and we were unable confirm the positive correlation between classifier performance and FD for the Japanese dataset. Still, in some cases such as the English dataset, there was a very weak positive or no correlation between FD and the classifier performance, with the higher FD datasets performing equally or even slightly better. This could mean that there is potential in the higher FD preprocessing types, namely, dependencies for CNNs. As the FD that had the best performance varied slightly throughout the classifiers and language, more exact ideal feature densities need to be confirmed for each classifier using datasets of different sizes and fields to make as accurate ranking of classifiers by FD as possible.

For the English dataset, the best results were usually obtained by slightly increasing the base FD, while for the Japanese and Polish dataset decreasing the base density tended to work better. This could be because Japanese and especially Polish are a lot more complex languages compared to English. This is also indicated by the fact that the LD of Japanese and Polish were higher than English.

The clearly visible optimal range of Feature Densities usually held between 50 to 70% of the proposed feature sets, which means that by ignoring the redundant preprocessings, the total training time could be reduced. Preprocessings using Chunking and Dependencies were usually left outside of the range which means that they add too much information for the classifiers to handle. Neural Networks make an exception here with some of the datasets.

We found out that the optimal feature densities tend to lie in specific ranges for all of the datasets. Even though the ranges were different for each classifier, we can still use this information to further decrease the required amount of work. Instead of running all of the experiments at once, we propose first discarding the FD ranges where the overall weakest feature sets according to the characteristics of the classifiers, POS, CHNK for all, and DEP for other than neural classifiers. Then running a subset of the experiments using the generally most effective feature sets, TOK/LEM+POS and variants and also DEP for neural classifiers. Afterwards, we can run more experiments with similar feature densities as the current peak to find the maximum performance.

We can take the Polish dataset with SGD SVM classifier as an example. First, we discard the feature sets that we assume to have a weak performance (POS, CHNK, DEP). Then train the classifier on the datasets that we found the most effective with English and Japanese datasets (TOK/LEM+POS and variants). At this point we have run 16 of the 68 total experiments and gotten an F-score of 0.498 (LEMPOSSTOP), which is already considerably higher than TOK (0.456). If we are not satisfied with the current high score here, we could start running more experiments with similar FD to LEMPOSSTOP. This way we could find a slightly higher score of 0.500 (LEMSTOP), but this would require eight additional experiments if starting from the FDs closest to LEMPOSSTOP. So, we managed to considerably improve the classifier's performance with less than a quarter of the original amount of experiments. We assume that the method could be also applied to other classifiers than those experimented with in this study, including modern models.

### *6.2. Dataset Complexity*

From the classification results it can be seen that the dataset with the lowest Lexical Density, the verification dataset, was the easiest to classify. While the highest Lexical Density dataset (Polish) was the hardest. However, the Japanese dataset got higher F-scores compared to English even though it had a higher Lexical Density. This means that natural language dataset complexity cannot be measured by using Lexical Density alone, even if it could be used to measure the complexity of the language itself.

There are some interesting points to note however when considering Feature Density. The relative change in Feature Density when lemmatizing each dataset matches the ranking of the dataset scores. This also applies to dependency patterns, except in the case of the Japanese dataset, which actually becomes the most dense, despite achieving the highest scores. From the results we can see that there are huge differences in classification results even when given datasets of a similar size and topic. This means it is not enough to develop the tools and models with just English in mind. Even though most of the research in NLP is done in English, it is a much less complex language when compared to other more morphologically rich languages, thus state of the art results achieved for English are not representative globally for the task in question, but rather locally, for the task done in this particular language. Additionally, even if it cannot be perfectly quantified by a simple measure like Feature Density, the complexity of languages affects the dataset complexity which in return impact the results of the classifiers applied. Even if a model has a high score in English it does not mean it excels with other languages.

### *6.3. Linguistic Preprocessing*

Using token-based preprocessings worked slightly better when classifying the English and Verification datasets. For the Japanese dataset, lemmas performed slightly better compared to tokens and for the Polish dataset, lemmas were clearly superior when compared to using tokens. What causes lemmas to work better for Japanese and Polish is most likely due to linguistic differences. Japanese and Polish are more complex languages than English, which can also be seen as an increase in the base distinct token count and thus feature density. The base feature density for tokens is too high for classifiers to correctly generalize on the dataset of this size. Lemmatizing, which puts all variously declensed and conjugated words back into their dictionary forms, cuts the base feature density, decreasing the complexity of the dataset, and thus increasing the achieved score. This can be seen especially well the case of Polish when the base FD is cut almost in half by lemmatizing, which greatly increases the score.

Chunking and dependency based preprocessings did not perform well compared to tokens or lemmas overall. They showed poor performance with all classifiers and datasets. Only exceptions being with English and Japanese datasets, where dependency based preprocessings showed some potential with the Neural Networks, although still outperformed by tokens or lemmas.

Stopword filtering was to be the one of the most effective preprocessing techniques for both English and Polish. Although for English, the results sometimes fluctuated. For Japanese, stopwords were not effective. This is because in Japanese each word/morpheme tends to contain useful information. This is different from English, where it is easier to filter out words that do not affect the overall meaning of a sentence.

Using parts-of-speech tags was shown to be very effective with both English and especially with Japanese, where it was clearly the most effective preprocessing method. Merging parts-of-speech information to the words themselves usually achieved a higher score than using them separately as features. This keeps the information directly connected to the word itself, which seems a better option from the point of view of information preservation.

Using Named Entity Recognition information had mixed results, reducing or only slightly increasing the classifier performance with English and Polish datasets. Only giving generally positive results with Japanese dataset. Overall, the performance of using NER seemed clearly inferior compared to stopword filtering or parts-of-speech information. Replacing words with their NER information seems to cause too much information loss and reduces the performance, while attaching NER information to the respective words performed somewhat better.

Filtering out non-alphabetic characters had the poorest performance of all the supplementary preprocessing types and reduced the classifier performance most of the time. Non-alphabetic tokens, which include e.g., punctuation marks, ellipsis, question-of exclamation marks, seem to carry useful information, at least in the context of cyberbullying detection, as removing them reduced the performance comparing to plain tokens due to information loss, which should be taken into account in future research.

In most cases, preprocessing had a considerable positive effect on the scores. We were able to find preprocessing methods that outperformed the base method of using pure tokens most of the time, which shows the potential of using linguistic information as a method for increasing classifier performance. Based on the experiments, we suggest omitting pure POS based and CHNK based feature sets for all classifiers due to their overall poor performance. Also, dependency based feature sets should be omitted for non-neural classifiers.

Only in the case of the verification dataset, which was found to be most likely too simple of a learning problem, the preprocessing methods were shown to be ineffective and had next to no impact on the results. The reason for this could be the simplicity of the learning problem. Most of the classifiers probably scored close to the maximum achievable score and thus the effects of extra preprocessing vanished. In the future, the effects of linguistic preprocessing should be also confirmed with state-of-the-art pretrained language models.

### *6.4. Effect on Cyberbullying Detection*

Leaving out the weaker feature sets and classifiers can be used to reduce time and effort when creating cyberbullying detection systems. The savings brought by the method do not only help to protect the environment but also allow quicker and more efficient development of CB detection systems. This is crucial as each day countless number of people become the victims of cyberbullies.

According to our experiments, although there were differences with F-scores for individual preprocessing types, the highest stability score datasets, tokens and lemmas with parts of speech information and their derivatives, were similar for both English and Japanese. The difference being the effectiveness of named entity recognition for Japanese. Also, we can say that from the applied classifiers, only SVMs, logistic regression and neural network models should be considered when dealing with CB.

The fact that the CNN's performance increased drastically when using dependency information instead of plain tokens with the English dataset could hint that structure and syntactic relations between words could be important when classifying CB entries. This needs to be explored further in the future.

Also, SVM (max F1=0.798), MLP (max F1=0.796) and Logistic Regression (max F1=0.793) achieved similar scores to RoBERTa (F1=0.797) on the English dataset, with SGD SVM actually slightly outperforming it with one of the feature sets (TOKPOS). This shows that the method can not only be used to save resources and time, but also to increase the performance of cyberbullying detection models.

### *6.5. Environmental Effect*

To see the concrete environmental effect of the method, we take a look at the English dataset, where the savings in power when training CNN were around 21kWh. According to European Environmental Agency (EEA) [6], the average greenhouse gas emissions from generating electricity was at 275 $g\ CO_2e/$kWh in 2019. This means the emissions generated by the training of CNN could be estimated at around 5.8 $kg\ CO_2e$. For comparison, the average new passenger car in the European Union in 2019, according to EEA, emits around 122 $g\ CO_2e$ per kilometer driven. This means that leaving out the weaker feature sets by utilizing FD could save as much as driving a new car for almost 50 kilometers in greenhouse gas emissions when training a simple CNN model.

Although the effect does not seem that overwhelming, one has to take into account that the models tested here were quite simple. Assuming our hypothesis, if the method would be applied to more resource intensive modern classifiers, the savings would become considerably more significant.

### *6.6. Future Research*

The usage of high Feature Density preprocessings, namely dependency parsing, showed potential with CNNs with English and Japanese datasets and needs to be confirmed with larger datasets in the future. Also, more exact ideal feature densities need to be confirmed for each classifier and language by using datasets of different sizes to make the ranking of classifiers by FD as accurate as possible.

[6] `https://www.eea.europa.eu/`

Due to the linguistic complexity of the Polish language, we would most likely require more data to make the use of neural networks viable. The effect of FD with neural networks in Polish needs to be explored further using a larger dataset in the future.

In order to use FD as a measure to estimate the complexity of a dataset, a more in depth study with more datasets of different sizes, topics and languages is required. Even if the complexity of languages cannot be perfectly quantified by a simple measure like Feature Density, there could be other measures or a combination of them that might be useful in better ranking datasets in terms of complexity.

Linguistic preprocessings had a positive effect on the scores, which shows their potential as a method for increasing classifier performance. In the future, we have to pretrain the linguistic embeddings on a larger corpus instead of relying on a small dataset. The effects of linguistic preprocessing should be also confirmed with other models, such as recurrent neural networks and the state-of-the-art pretrained language models like BERT.

## 7. Conclusions

In this paper we presented our research on Feature Density and linguistically-backed preprocessing methods, applied in cyberbullying detection. Both concepts are relatively novel to the field. We studied the effectiveness of Feature Density using a variety of linguistically-backed feature preprocessing methods to estimate dataset complexity and classifier performance.

From the results we concluded that most of the classifier performances, excluding CNNs, had a strong negative correlation with FD, which means weaker performance if a lot of linguistic information is added. Depending on the dataset and classifier, the best performance was in most cases between 50-200% of the base FD (TOK). For CNNs, we were unable confirm the positive correlation between classifier performance and FD for the Japanese dataset as there was usually only a very weak positive or no correlation between FD and the classifier performance. Still, there could be potential in the higher FD preprocessing types for CNNs. We also discovered that the best results for English were usually obtained by slightly increasing the base feature density, while for the Japanese and Polish dataset it was the opposite. The reason behind this is probably due to the difference in language complexity of English compared to Japanese and Polish.

The results indicated that natural language dataset complexity cannot be measured by using Lexical Density or Feature Density alone, even if it could be used to measure the complexity of the language itself. However, we found out that the relative change in Feature Density when lemmatizing each dataset matches the ranking of the dataset scores. We also noticed that even with datasets of similar sizes and topics, but with different languages, there can be huge differences in classification results. This means it is not enough to develop the tools and models with just English in mind, as it is a much less complex language when compared to others like Japanese or Polish. Which means state-of-the-art results achieved for English are not representative globally for the task in question, but rather locally, for the task done in this particular language.

Using tokens worked slightly better for English while lemmas were better for Japanese and especially Polish. The reason being most likely related to the complexity of the language, as the base feature density for tokens is too high in Japanese and Polish for classifiers to correctly generalize on the dataset of this size and thus lowering the complexity by lemmatizing resulted in an increased performance. Chunking and dependencies generally had a poor performance but dependencies still showed some potential with neural networks. From the supplementary preprocesing methods, stopwords and POS tagging generally resulted in positive results, except for the questionable use of stopwords with the Japanese dataset. NER showed mixed results, being usable mainly with Japanese while alphabetic filtering had poor results with all of the datasets due to non-alphabetic tokens also carrying useful information, at least in the context of cyberbullying detection. In general, the preprocessing had a positive effect on the scores which proves that using linguistically-backed preprocessing can be used to increase classifier performance, at least in the context of cyberbullying. We also discovered that parts of speech information can be used with tokens or lemmas to produce the most stable and high performance feature sets.

We proposed that the method can be applied by first discarding the FD ranges where the overall weakest feature sets are (POS, CHNK, etc.). Then running a small subset of the experiments with a set interval between preprocessing type feature densities and iterate around the most probable peak to find the maximum performance. We concluded that the method could save as much as driving a new car for almost 50 kilometers in greenhouse gas emissions when training a simple CNN model with the English dataset by discarding the weakest feature sets. We assume that the method could also be applied to more modern models than those used in this study.

In the future, the usage of high FD preprocessings and the potential of neural networks needs to be confirmed with larger datasets. We also need to find more exact ideal feature densities for each classifier and language by using datasets of different sizes to make the performance estimation as accurate as possible. The effectiveness of FD as a measure for estimating dataset complexity needs to be further explored with datasets of different sizes, topics and languages and even if the complexity cannot be perfectly quantified by a simple measure like Feature Density, there could be other measures for ranking the datasets. Lastly, the positive effects of linguistic preprocessing need to be confirmed in other fields and with state-of-the-art models.

# Appendix A. F-score and Standard Error of Classifier-Preprocessing Pairs

Table A.18: English dataset: F-scores and average standard error for classifier-preprocessing pairs (F1±stderr)

| | LBFGS LR | Newton LR | Linear SVM | SGD SVM | KNN | NaiveBayes | RandomForest |
|---|---|---|---|---|---|---|---|
| **CHNK** | 0.727±0.234 | 0.726±0.242 | 0.718±0.257 | 0.736±0.237 | 0.570±0.390 | 0.674±0.247 | 0.613±0.366 |
| **CHNKALPHA** | 0.684±0.262 | 0.681±0.259 | 0.669±0.266 | 0.683±0.263 | 0.607±0.339 | 0.643±0.263 | 0.647±0.324 |
| **CHNKNER** | 0.718±0.238 | 0.723±0.236 | 0.721±0.248 | 0.737±0.234 | 0.582±0.385 | 0.669±0.248 | 0.603±0.379 |
| **CHNKNERALPHA** | 0.675±0.262 | 0.676±0.260 | 0.663±0.271 | 0.663±0.267 | 0.599±0.341 | 0.641±0.262 | 0.618±0.336 |
| **CHNKNERR** | 0.688±0.257 | 0.695±0.258 | 0.702±0.265 | 0.699±0.257 | 0.580±0.380 | 0.653±0.255 | 0.603±0.368 |
| **CHNKNERRALPHA** | 0.660±0.269 | 0.663±0.268 | 0.651±0.277 | 0.657±0.276 | 0.603±0.335 | 0.626±0.267 | 0.616±0.343 |
| **CHNKNERRSTOP** | 0.686±0.265 | 0.684±0.267 | 0.684±0.274 | 0.694±0.264 | 0.577±0.378 | 0.629±0.259 | 0.635±0.325 |
| **CHNKNERRSTOPALPHA** | 0.618±0.280 | 0.617±0.282 | 0.591±0.288 | 0.607±0.287 | 0.404±0.233 | 0.598±0.276 | 0.620±0.304 |
| **CHNKNERSTOP** | 0.724±0.241 | 0.724±0.241 | 0.715±0.254 | 0.724±0.241 | 0.582±0.389 | 0.663±0.251 | 0.635±0.339 |
| **CHNKNERSTOPALPHA** | 0.666±0.261 | 0.661±0.265 | 0.644±0.279 | 0.668±0.260 | 0.386±0.218 | 0.615±0.268 | 0.659±0.301 |
| **CHNKSTOP** | 0.722±0.241 | 0.721±0.239 | 0.711±0.256 | 0.723±0.239 | 0.577±0.379 | 0.670±0.247 | 0.667±0.308 |
| **CHNKSTOPALPHA** | 0.629±0.277 | 0.637±0.274 | 0.606±0.286 | 0.619±0.277 | 0.395±0.226 | 0.608±0.271 | 0.649±0.305 |
| **DEP** | 0.617±0.354 | 0.619±0.349 | 0.568±0.398 | 0.587±0.381 | 0.243±0.094 | 0.617±0.280 | 0.536±0.435 |
| **DEPALPHA** | 0.609±0.349 | 0.612±0.346 | 0.588±0.372 | 0.601±0.356 | 0.314±0.156 | 0.600±0.288 | 0.545±0.416 |
| **DEPNER** | 0.624±0.339 | 0.621±0.341 | 0.574±0.391 | 0.585±0.379 | 0.242±0.092 | 0.611±0.279 | 0.528±0.443 |
| **DEPNERALPHA** | 0.585±0.308 | 0.589±0.311 | 0.561±0.317 | 0.579±0.317 | 0.213±0.069 | 0.607±0.282 | 0.578±0.384 |
| **DEPNERR** | 0.610±0.350 | 0.614±0.341 | 0.571±0.389 | 0.587±0.381 | 0.241±0.094 | 0.611±0.283 | 0.533±0.426 |
| **DEPNERRALPHA** | 0.606±0.352 | 0.605±0.355 | 0.589±0.367 | 0.602±0.359 | 0.312±0.156 | 0.596±0.293 | 0.537±0.418 |
| **DEPNERRSTOP** | 0.602±0.358 | 0.599±0.360 | 0.564±0.395 | 0.568±0.393 | 0.273±0.125 | 0.615±0.291 | 0.543±0.421 |
| **DEPNERRSTOPALPHA** | 0.584±0.366 | 0.584±0.365 | 0.560±0.386 | 0.581±0.376 | 0.386±0.223 | 0.599±0.313 | 0.544±0.421 |
| **DEPNERSTOP** | 0.611±0.348 | 0.602±0.352 | 0.564±0.397 | 0.576±0.386 | 0.274±0.119 | 0.604±0.284 | 0.527±0.438 |
| **DEPNERSTOPALPHA** | 0.535±0.313 | 0.531±0.314 | 0.523±0.319 | 0.523±0.310 | 0.297±0.151 | 0.543±0.287 | 0.563±0.372 |
| **DEPSTOP** | 0.606±0.365 | 0.595±0.367 | 0.562±0.400 | 0.571±0.399 | 0.276±0.122 | 0.616±0.288 | 0.544±0.426 |
| **DEPSTOPALPHA** | 0.586±0.371 | 0.587±0.365 | 0.564±0.386 | 0.588±0.387 | 0.388±0.221 | 0.594±0.309 | 0.539±0.415 |
| **LEM** | 0.781±0.180 | 0.786±0.187 | 0.784±0.190 | 0.790±0.182 | 0.634±0.326 | 0.715±0.221 | 0.724±0.253 |
| **LEMALPHA** | 0.755±0.195 | 0.764±0.198 | 0.745±0.204 | 0.765±0.198 | 0.294±0.179 | 0.703±0.225 | 0.718±0.244 |
| **LEMNER** | 0.784±0.187 | 0.782±0.183 | 0.787±0.182 | 0.792±0.176 | 0.631±0.330 | 0.710±0.224 | 0.716±0.261 |
| **LEMNERALPHA** | 0.763±0.197 | 0.764±0.196 | 0.765±0.203 | 0.767±0.194 | 0.637±0.315 | 0.699±0.227 | 0.710±0.255 |
| **LEMNERR** | 0.740±0.217 | 0.737±0.219 | 0.742±0.217 | 0.740±0.214 | 0.601±0.335 | 0.692±0.234 | 0.697±0.280 |
| **LEMNERRALPHA** | 0.729±0.222 | 0.728±0.222 | 0.725±0.227 | 0.725±0.222 | 0.614±0.321 | 0.685±0.237 | 0.699±0.282 |
| **LEMNERRSTOP** | 0.737±0.220 | 0.734±0.221 | 0.726±0.228 | 0.732±0.220 | 0.609±0.329 | 0.682±0.235 | 0.727±0.245 |
| **LEMNERRSTOPALPHA** | 0.732±0.222 | 0.732±0.225 | 0.714±0.236 | 0.727±0.227 | 0.624±0.323 | 0.674±0.239 | 0.723±0.249 |
| **LEMNERSTOP** | 0.782±0.185 | 0.783±0.187 | 0.782±0.186 | 0.792±0.179 | 0.634±0.330 | 0.706±0.225 | 0.745±0.230 |
| **LEMNERSTOPALPHA** | 0.770±0.195 | 0.767±0.195 | 0.752±0.201 | 0.767±0.200 | 0.640±0.304 | 0.693±0.226 | 0.739±0.231 |
| **LEMPOS** | 0.778±0.204 | 0.778±0.198 | 0.788±0.202 | 0.790±0.202 | 0.517±0.365 | 0.711±0.222 | 0.663±0.310 |
| **LEMPOSALPHA** | 0.768±0.204 | 0.772±0.210 | 0.772±0.211 | 0.768±0.209 | 0.522±0.210 | 0.700±0.228 | 0.654±0.298 |
| **LEMPOSS** | 0.764±0.202 | 0.765±0.207 | 0.769±0.205 | 0.767±0.202 | 0.564±0.359 | 0.713±0.227 | 0.658±0.270 |
| **LEMPOSSALPHA** | 0.760±0.200 | 0.758±0.204 | 0.753±0.213 | 0.758±0.209 | 0.406±0.254 | 0.705±0.230 | 0.669±0.259 |
| **LEMPOSSSTOP** | 0.763±0.189 | 0.766±0.187 | 0.767±0.186 | 0.774±0.177 | 0.566±0.296 | 0.709±0.219 | 0.706±0.327 |
| **LEMPOSSSTOPALPHA** | 0.762±0.198 | 0.766±0.193 | 0.748±0.194 | 0.765±0.190 | 0.490±0.297 | 0.702±0.227 | 0.713±0.314 |
| **LEMPOSSTOP** | 0.780±0.187 | 0.781±0.191 | 0.788±0.183 | 0.788±0.186 | 0.642±0.285 | 0.708±0.222 | 0.708±0.261 |
| **LEMPOSSTOPALPHA** | 0.770±0.193 | 0.769±0.195 | 0.766±0.202 | 0.768±0.191 | 0.669±0.282 | 0.696±0.227 | 0.718±0.256 |
| **LEMSTOP** | 0.787±0.183 | 0.786±0.185 | 0.784±0.192 | 0.791±0.181 | 0.641±0.320 | 0.713±0.221 | 0.754±0.229 |
| **LEMSTOPALPHA** | 0.772±0.190 | 0.766±0.192 | 0.766±0.196 | 0.773±0.193 | 0.357±0.203 | 0.702±0.227 | 0.747±0.220 |
| **POSS** | 0.487±0.291 | 0.487±0.290 | 0.488±0.293 | 0.491±0.292 | 0.522±0.368 | 0.498±0.306 | 0.556±0.392 |
| **POSSALPHA** | 0.488±0.290 | 0.486±0.290 | 0.488±0.294 | 0.498±0.290 | 0.526±0.357 | 0.498±0.306 | 0.552±0.396 |
| **POSSSTOP** | 0.477±0.287 | 0.477±0.287 | 0.471±0.282 | 0.467±0.283 | 0.518±0.389 | 0.486±0.297 | 0.540±0.388 |
| **POSSSTOPALPHA** | 0.469±0.283 | 0.470±0.284 | 0.471±0.288 | 0.465±0.280 | 0.517±0.407 | 0.478±0.297 | 0.525±0.376 |
| **TOK** | 0.793±0.182 | 0.788±0.182 | 0.793±0.182 | 0.796±0.173 | 0.632±0.337 | 0.716±0.221 | 0.711±0.272 |
| **TOKALPHA** | 0.768±0.197 | 0.768±0.193 | 0.757±0.202 | 0.773±0.191 | 0.271±0.175 | 0.705±0.224 | 0.721±0.261 |
| **TOKNER** | 0.789±0.185 | 0.785±0.179 | 0.788±0.182 | 0.789±0.183 | 0.609±0.344 | 0.708±0.225 | 0.703±0.271 |
| **TOKNERALPHA** | 0.768±0.194 | 0.771±0.194 | 0.763±0.200 | 0.776±0.189 | 0.628±0.325 | 0.696±0.230 | 0.701±0.270 |
| **TOKNERR** | 0.741±0.215 | 0.744±0.218 | 0.737±0.219 | 0.743±0.216 | 0.600±0.337 | 0.696±0.231 | 0.688±0.281 |
| **TOKNERRALPHA** | 0.734±0.215 | 0.735±0.220 | 0.735±0.228 | 0.730±0.217 | 0.624±0.320 | 0.683±0.235 | 0.681±0.280 |
| **TOKNERRSTOP** | 0.736±0.220 | 0.736±0.222 | 0.728±0.224 | 0.732±0.223 | 0.609±0.333 | 0.680±0.239 | 0.730±0.244 |
| **TOKNERRSTOPALPHA** | 0.728±0.225 | 0.731±0.222 | 0.727±0.237 | 0.723±0.229 | 0.623±0.319 | 0.675±0.243 | 0.721±0.249 |
| **TOKNERSTOP** | 0.785±0.186 | 0.791±0.186 | 0.790±0.186 | 0.790±0.181 | 0.635±0.324 | 0.703±0.228 | 0.732±0.230 |
| **TOKNERSTOPALPHA** | 0.773±0.197 | 0.771±0.191 | 0.762±0.208 | 0.774±0.193 | 0.646±0.309 | 0.691±0.233 | 0.737±0.236 |
| **TOKPOS** | 0.781±0.199 | 0.783±0.199 | 0.791±0.203 | 0.798±0.193 | 0.565±0.378 | 0.713±0.222 | 0.656±0.313 |
| **TOKPOSALPHA** | 0.775±0.203 | 0.775±0.206 | 0.778±0.214 | 0.784±0.208 | 0.576±0.191 | 0.699±0.223 | 0.653±0.304 |
| **TOKPOSS** | 0.766±0.203 | 0.768±0.200 | 0.767±0.201 | 0.783±0.195 | 0.549±0.372 | 0.715±0.224 | 0.648±0.268 |
| **TOKPOSSALPHA** | 0.765±0.195 | 0.761±0.196 | 0.763±0.196 | 0.767±0.192 | 0.378±0.277 | 0.709±0.228 | 0.662±0.258 |
| **TOKPOSSSTOP** | 0.763±0.184 | 0.765±0.184 | 0.767±0.182 | 0.773±0.174 | 0.563±0.307 | 0.704±0.223 | 0.703±0.317 |
| **TOKPOSSSTOPALPHA** | 0.774±0.192 | 0.773±0.193 | 0.774±0.189 | 0.771±0.186 | 0.671±0.302 | 0.694±0.223 | 0.722±0.318 |
| **TOKPOSSTOP** | 0.786±0.187 | 0.783±0.186 | 0.794±0.182 | 0.792±0.179 | 0.645±0.286 | 0.700±0.225 | 0.711±0.262 |
| **TOKPOSSTOPALPHA** | - | - | - | - | - | - | - |
| **TOKSTOP** | 0.793±0.180 | 0.790±0.182 | 0.784±0.188 | 0.794±0.180 | 0.644±0.327 | 0.708±0.222 | 0.758±0.216 |
| **TOKSTOPALPHA** | 0.775±0.188 | 0.776±0.190 | 0.766±0.199 | 0.776±0.192 | 0.342±0.194 | 0.700±0.228 | 0.745±0.217 |
| **Avg clf F1** | 0.705 | 0.705 | 0.696 | 0.704 | 0.508 | 0.660 | 0.655 |
| **Avg clf stderr** | 0.241 | 0.241 | 0.250 | 0.244 | 0.281 | 0.249 | 0.312 |

| | AdaBoost | XGBoost | MLP | CNN1 | CNN2 | Avg pp F1 | Avg pp stderr |
|---|---|---|---|---|---|---|---|
| **CHNK** | 0.649±0.298 | 0.667±0.289 | 0.724±0.310 | 0.657±0.256 | 0.666±0.278 | 0.677 | 0.284 |
| **CHNKALPHA** | 0.616±0.302 | 0.676±0.291 | 0.695±0.385 | 0.587±0.337 | 0.583±0.429 | 0.648 | 0.310 |
| **CHNKNER** | 0.630±0.299 | 0.673±0.292 | 0.722±0.318 | 0.654±0.196 | 0.642±0.287 | 0.673 | 0.280 |
| **CHNKNERALPHA** | 0.609±0.301 | 0.649±0.304 | 0.684±0.390 | 0.557±0.282 | 0.614±0.426 | 0.637 | 0.308 |
| **CHNKNERR** | 0.608±0.310 | 0.642±0.305 | 0.704±0.320 | 0.645±0.262 | 0.662±0.288 | 0.657 | 0.294 |
| **CHNKNERRALPHA** | 0.599±0.312 | 0.653±0.302 | 0.674±0.412 | 0.566±0.288 | 0.600±0.437 | 0.631 | 0.315 |
| **CHNKNERRSTOP** | 0.621±0.308 | 0.652±0.300 | 0.693±0.365 | 0.402±0.264 | 0.344±0.223 | 0.608 | 0.291 |
| **CHNKNERRSTOPALPHA** | 0.582±0.312 | 0.648±0.298 | 0.623±0.445 | 0.451±0.314 | 0.340±0.370 | 0.558 | 0.308 |
| **CHNKNERSTOP** | 0.652±0.291 | 0.679±0.285 | 0.720±0.296 | 0.501±0.200 | 0.298±0.195 | 0.635 | 0.268 |
| **CHNKNERSTOPALPHA** | 0.625±0.301 | 0.656±0.291 | 0.647±0.443 | 0.431±0.340 | 0.406±0.286 | 0.589 | 0.293 |
| **CHNKSTOP** | 0.648±0.290 | 0.679±0.279 | 0.715±0.278 | 0.386±0.257 | 0.342±0.196 | 0.630 | 0.268 |
| **CHNKSTOPALPHA** | 0.654±0.290 | 0.664±0.285 | 0.628±0.431 | 0.455±0.277 | 0.374±0.317 | 0.577 | 0.293 |
| **DEP** | 0.566±0.344 | 0.598±0.349 | 0.594±0.319 | 0.682±0.214 | 0.694±0.249 | 0.577 | 0.314 |
| **DEPALPHA** | 0.552±0.352 | 0.604±0.346 | 0.598±0.392 | 0.606±0.396 | 0.62±0.418 | 0.571 | 0.349 |
| **DEPNER** | 0.564±0.350 | 0.595±0.356 | 0.592±0.325 | 0.686±0.211 | 0.692±0.263 | 0.576 | 0.314 |
| **DEPNERALPHA** | 0.497±0.287 | 0.593±0.328 | 0.603±0.432 | 0.606±0.351 | 0.623±0.407 | 0.553 | 0.316 |
| **DEPNERR** | 0.562±0.350 | 0.596±0.357 | 0.595±0.334 | 0.670±0.227 | 0.695±0.299 | 0.574 | 0.319 |
| **DEPNERRALPHA** | 0.556±0.349 | 0.595±0.354 | 0.593±0.440 | 0.585±0.311 | 0.622±0.415 | 0.567 | 0.347 |
| **DEPNERRSTOP** | 0.572±0.350 | 0.600±0.354 | - | 0.726±0.225 | 0.702±0.249 | 0.579 | 0.320 |
| **DEPNERRSTOPALPHA** | 0.561±0.357 | 0.595±0.361 | 0.574±0.449 | 0.583±0.328 | 0.619±0.444 | 0.564 | 0.366 |
| **DEPNERSTOP** | 0.563±0.360 | 0.604±0.360 | 0.577±0.341 | 0.725±0.219 | 0.708±0.248 | 0.578 | 0.321 |
| **DEPNERSTOPALPHA** | 0.422±0.239 | 0.576±0.346 | 0.564±0.465 | 0.630±0.321 | 0.632±0.409 | 0.528 | 0.321 |
| **DEPSTOP** | 0.576±0.349 | 0.603±0.358 | 0.584±0.330 | 0.741±0.209 | 0.648±0.271 | 0.577 | 0.324 |
| **DEPSTOPALPHA** | 0.568±0.363 | 0.595±0.365 | 0.578±0.446 | 0.629±0.352 | 0.625±0.386 | 0.570 | 0.364 |
| **LEM** | 0.720±0.222 | 0.744±0.225 | 0.786±0.258 | 0.670±0.193 | 0.665±0.265 | 0.733 | 0.225 |
| **LEMALPHA** | 0.705±0.242 | 0.748±0.222 | 0.754±0.286 | 0.610±0.208 | 0.651±0.283 | 0.684 | 0.224 |
| **LEMNER** | 0.720±0.237 | 0.742±0.227 | 0.780±0.274 | 0.680±0.190 | 0.613±0.268 | 0.728 | 0.228 |
| **LEMNERALPHA** | 0.707±0.247 | 0.742±0.238 | 0.768±0.299 | 0.662±0.203 | 0.671±0.284 | 0.721 | 0.238 |
| **LEMNERR** | 0.683±0.260 | 0.724±0.246 | 0.749±0.286 | 0.658±0.203 | 0.663±0.231 | 0.702 | 0.245 |
| **LEMNERRALPHA** | 0.680±0.260 | 0.710±0.244 | 0.740±0.267 | 0.645±0.277 | 0.652±0.289 | 0.694 | 0.256 |
| **LEMNERRSTOP** | 0.690±0.258 | 0.720±0.242 | 0.741±0.343 | 0.371±0.306 | 0.364±0.214 | 0.653 | 0.255 |
| **LEMNERRSTOPALPHA** | 0.682±0.255 | 0.704±0.247 | 0.737±0.310 | 0.372±0.214 | 0.348±0.267 | 0.647 | 0.251 |
| **LEMNERSTOP** | 0.725±0.231 | 0.742±0.222 | 0.780±0.298 | 0.429±0.198 | 0.378±0.215 | 0.690 | 0.224 |
| **LEMNERSTOPALPHA** | 0.716±0.237 | 0.738±0.233 | 0.768±0.343 | 0.460±0.210 | 0.414±0.216 | 0.685 | 0.233 |
| **LEMPOS** | 0.727±0.264 | 0.741±0.244 | 0.783±0.251 | 0.665±0.194 | 0.640±0.264 | 0.715 | 0.243 |
| **LEMPOSALPHA** | 0.713±0.272 | 0.727±0.252 | 0.775±0.340 | 0.664±0.213 | 0.695±0.250 | 0.711 | 0.241 |
| **LEMPOSS** | 0.679±0.251 | 0.717±0.249 | 0.773±0.335 | 0.662±0.262 | 0.736±0.271 | 0.714 | 0.253 |
| **LEMPOSSALPHA** | 0.674±0.266 | 0.712±0.249 | 0.756±0.306 | 0.603±0.217 | 0.715±0.313 | 0.689 | 0.243 |
| **LEMPOSSSTOP** | 0.691±0.225 | 0.720±0.230 | 0.773±0.291 | 0.683±0.221 | 0.725±0.234 | 0.729 | 0.232 |
| **LEMPOSSSTOPALPHA** | 0.681±0.236 | 0.714±0.238 | 0.757±0.296 | 0.593±0.263 | 0.716±0.320 | 0.701 | 0.247 |
| **LEMPOSSTOP** | 0.721±0.232 | 0.735±0.231 | 0.783±0.353 | 0.715±0.260 | 0.707±0.270 | 0.738 | 0.238 |
| **LEMPOSSTOPALPHA** | 0.722±0.241 | 0.730±0.239 | 0.778±0.294 | 0.669±0.265 | 0.698±0.227 | 0.729 | 0.234 |
| **LEMSTOP** | 0.732±0.228 | 0.752±0.219 | 0.789±0.321 | 0.403±0.197 | 0.327±0.205 | 0.688 | 0.224 |
| **LEMSTOPALPHA** | 0.712±0.236 | 0.745±0.220 | 0.764±0.327 | 0.377±0.212 | 0.329±0.210 | 0.651 | 0.219 |
| **POSS** | 0.509±0.330 | 0.555±0.379 | 0.488±0.481 | 0.540±0.481 | 0.536±0.481 | 0.513 | 0.365 |
| **POSSALPHA** | 0.518±0.331 | 0.549±0.381 | 0.493±0.481 | 0.538±0.481 | 0.534±0.481 | 0.514 | 0.365 |
| **POSSSTOP** | 0.496±0.317 | 0.533±0.369 | 0.484±0.481 | 0.431±0.481 | 0.434±0.481 | 0.485 | 0.362 |
| **POSSSTOPALPHA** | 0.484±0.303 | 0.511±0.354 | 0.491±0.481 | 0.428±0.481 | 0.484±0.481 | 0.483 | 0.360 |
| **TOK** | 0.728±0.234 | 0.748±0.223 | 0.796±0.298 | 0.659±0.191 | 0.661±0.258 | 0.735 | 0.229 |
| **TOKALPHA** | 0.705±0.246 | 0.742±0.224 | 0.756±0.282 | 0.643±0.196 | 0.652±0.310 | 0.688 | 0.225 |
| **TOKNER** | 0.722±0.239 | 0.745±0.226 | 0.784±0.288 | 0.684±0.195 | 0.680±0.240 | 0.732 | 0.230 |
| **TOKNERALPHA** | 0.705±0.238 | 0.746±0.229 | 0.775±0.309 | 0.649±0.202 | 0.648±0.257 | 0.719 | 0.236 |
| **TOKNERR** | 0.671±0.263 | 0.719±0.246 | 0.749±0.291 | 0.655±0.200 | 0.631±0.275 | 0.698 | 0.249 |
| **TOKNERRALPHA** | 0.674±0.264 | 0.704±0.257 | 0.748±0.307 | 0.626±0.215 | 0.655±0.270 | 0.694 | 0.252 |
| **TOKNERRSTOP** | 0.678±0.260 | 0.710±0.253 | 0.751±0.303 | 0.406±0.212 | 0.317±0.225 | 0.651 | 0.247 |
| **TOKNERRSTOPALPHA** | 0.680±0.260 | 0.698±0.254 | 0.744±0.357 | 0.412±0.222 | 0.394±0.230 | 0.655 | 0.254 |
| **TOKNERSTOP** | 0.721±0.230 | 0.743±0.222 | 0.790±0.325 | 0.444±0.205 | 0.367±0.209 | 0.691 | 0.226 |
| **TOKNERSTOPALPHA** | 0.704±0.241 | 0.740±0.226 | 0.771±0.330 | 0.371±0.262 | 0.379±0.217 | 0.677 | 0.237 |
| **TOKPOS** | 0.720±0.270 | 0.739±0.255 | 0.787±0.344 | 0.626±0.201 | 0.705±0.227 | 0.722 | 0.250 |
| **TOKPOSALPHA** | 0.705±0.277 | 0.731±0.262 | 0.783±0.328 | 0.633±0.268 | 0.698±0.250 | 0.716 | 0.245 |
| **TOKPOSS** | 0.671±0.254 | 0.715±0.249 | 0.773±0.332 | 0.686±0.204 | 0.729±0.231 | 0.714 | 0.245 |
| **TOKPOSSALPHA** | 0.656±0.241 | 0.709±0.241 | 0.769±0.418 | 0.643±0.269 | 0.658±0.275 | 0.687 | 0.249 |
| **TOKPOSSSTOP** | 0.684±0.244 | 0.724±0.229 | 0.771±0.354 | 0.675±0.205 | 0.722±0.227 | 0.718 | 0.236 |
| **TOKPOSSSTOPALPHA** | 0.713±0.242 | 0.730±0.233 | 0.779±0.328 | 0.680±0.265 | 0.698±0.238 | 0.732 | 0.242 |
| **TOKPOSSTOP** | 0.733±0.230 | 0.739±0.232 | 0.789±0.338 | 0.706±0.200 | 0.691±0.226 | 0.739 | 0.228 |
| **TOKPOSSTOPALPHA** | - | - | - | - | - | - | - |
| **TOKSTOP** | 0.736±0.227 | 0.749±0.218 | 0.787±0.368 | 0.355±0.199 | 0.321±0.207 | 0.685 | 0.226 |
| **TOKSTOPALPHA** | 0.714±0.231 | 0.744±0.215 | 0.765±0.337 | 0.452±0.215 | 0.425±0.210 | 0.665 | 0.218 |
| **Avg clf F1** | 0.649 | 0.682 | 0.703 | 0.580 | 0.576 | | |
| **Avg clf stderr** | 0.277 | 0.275 | 0.347 | 0.257 | 0.289 | | |

Table A.19: Japanese dataset: F-scores and average standard error for classifier-preprocessing pairs (F1, stderr)

| | LBFGS LR | Newton LR | Linear SVM | SGD SVM | KNN | NaiveBayes | RandomForest |
|---|---|---|---|---|---|---|---|
| **CHNK** | 0.710±0.040 | 0.708±0.045 | 0.718±0.041 | 0.715±0.047 | 0.495±0.206 | 0.753±0.033 | 0.677±0.054 |
| **CHNKALPHA** | 0.714±0.028 | 0.719±0.029 | 0.709±0.031 | 0.702±0.040 | 0.510±0.179 | 0.740±0.027 | 0.684±0.043 |
| **CHNKNER** | 0.758±0.042 | 0.758±0.037 | 0.769±0.034 | 0.763±0.031 | 0.523±0.174 | 0.772±0.031 | 0.738±0.039 |
| **CHNKNERALPHA** | 0.751±0.031 | 0.755±0.030 | 0.757±0.031 | 0.751±0.034 | 0.555±0.159 | 0.767±0.029 | 0.725±0.040 |
| **CHNKNERR** | 0.752±0.030 | 0.751±0.030 | 0.756±0.032 | 0.745±0.038 | 0.517±0.151 | 0.757±0.030 | 0.727±0.044 |
| **CHNKNERRALPHA** | 0.733±0.046 | 0.736±0.040 | 0.744±0.035 | 0.737±0.037 | 0.529±0.163 | 0.745±0.023 | 0.712±0.034 |
| **CHNKNERRSTOP** | 0.745±0.036 | 0.744±0.040 | 0.753±0.026 | 0.737±0.038 | 0.524±0.165 | 0.739±0.040 | 0.729±0.042 |
| **CHNKNERRSTOPALPHA** | 0.731±0.042 | 0.730±0.048 | 0.741±0.036 | 0.723±0.037 | 0.539±0.160 | 0.730±0.026 | 0.718±0.053 |
| **CHNKNERSTOP** | 0.754±0.040 | 0.755±0.040 | 0.764±0.033 | 0.776±0.024 | 0.543±0.160 | 0.767±0.029 | 0.738±0.039 |
| **CHNKNERSTOPALPHA** | 0.758±0.037 | 0.755±0.032 | 0.760±0.032 | 0.761±0.027 | 0.528±0.171 | 0.750±0.022 | 0.721±0.045 |
| **CHNKSTOP** | 0.712±0.051 | 0.714±0.053 | 0.724±0.050 | 0.716±0.045 | 0.506±0.200 | 0.728±0.046 | 0.675±0.067 |
| **CHNKSTOPALPHA** | 0.695±0.049 | 0.701±0.050 | 0.714±0.048 | 0.700±0.049 | 0.518±0.191 | 0.721±0.034 | 0.666±0.071 |
| **DEP** | 0.678±0.069 | 0.681±0.070 | 0.685±0.066 | 0.680±0.055 | 0.495±0.188 | 0.682±0.061 | 0.608±0.124 |
| **DEPALPHA** | 0.669±0.074 | 0.668±0.074 | 0.681±0.067 | 0.674±0.065 | 0.464±0.194 | 0.674±0.061 | 0.593±0.137 |
| **DEPNER** | 0.727±0.053 | 0.727±0.045 | 0.742±0.035 | 0.746±0.033 | 0.502±0.171 | 0.724±0.040 | 0.708±0.059 |
| **DEPNERALPHA** | 0.730±0.046 | 0.731±0.048 | 0.741±0.032 | 0.748±0.036 | 0.427±0.248 | 0.718±0.043 | 0.707±0.065 |
| **DEPNERR** | 0.677±0.065 | 0.684±0.067 | 0.681±0.066 | 0.686±0.061 | 0.479±0.200 | 0.676±0.060 | 0.608±0.122 |
| **DEPNERRALPHA** | 0.667±0.074 | 0.671±0.077 | 0.675±0.066 | 0.674±0.070 | 0.461±0.187 | 0.670±0.055 | 0.585±0.137 |
| **DEPNERRSTOP** | 0.659±0.084 | 0.652±0.083 | 0.659±0.086 | 0.649±0.082 | 0.509±0.166 | 0.645±0.075 | 0.571±0.152 |
| **DEPNERRSTOPALPHA** | 0.617±0.108 | 0.617±0.109 | 0.619±0.109 | 0.618±0.102 | 0.493±0.167 | 0.627±0.099 | 0.534±0.180 |
| **DEPNERSTOP** | 0.717±0.055 | 0.716±0.053 | 0.728±0.047 | 0.723±0.034 | 0.544±0.130 | 0.701±0.041 | 0.693±0.062 |
| **DEPNERSTOPALPHA** | 0.715±0.050 | 0.711±0.050 | 0.722±0.039 | 0.713±0.041 | 0.448±0.237 | 0.731±0.032 | 0.695±0.063 |
| **DEPSTOP** | 0.655±0.083 | 0.658±0.079 | 0.658±0.082 | 0.653±0.076 | 0.507±0.166 | 0.647±0.074 | 0.575±0.148 |
| **DEPSTOPALPHA** | 0.627±0.109 | 0.623±0.105 | 0.622±0.100 | 0.622±0.094 | 0.499±0.169 | 0.628±0.093 | 0.537±0.168 |
| **LEM** | 0.803±0.020 | 0.799±0.022 | 0.817±0.017 | 0.802±0.026 | 0.592±0.115 | 0.823±0.020 | 0.771±0.032 |
| **LEMALPHA** | 0.800±0.021 | 0.808±0.032 | 0.824±0.024 | 0.807±0.022 | 0.587±0.115 | 0.815±0.022 | 0.758±0.030 |
| **LEMNER** | 0.806±0.028 | 0.801±0.032 | 0.818±0.023 | 0.816±0.022 | 0.633±0.110 | 0.823±0.029 | 0.790±0.022 |
| **LEMNERALPHA** | 0.809±0.028 | 0.807±0.022 | 0.827±0.020 | 0.823±0.014 | 0.612±0.102 | 0.827±0.029 | 0.797±0.026 |
| **LEMNERR** | 0.794±0.029 | 0.790±0.032 | 0.796±0.019 | 0.791±0.024 | 0.585±0.117 | 0.808±0.022 | 0.778±0.026 |
| **LEMNERRALPHA** | 0.797±0.027 | 0.796±0.021 | 0.805±0.027 | 0.793±0.021 | 0.585±0.114 | 0.811±0.024 | 0.781±0.032 |
| **LEMNERRSTOP** | 0.787±0.024 | 0.784±0.028 | 0.793±0.023 | 0.793±0.024 | 0.590±0.121 | 0.794±0.025 | 0.796±0.024 |
| **LEMNERRSTOPALPHA** | 0.790±0.024 | 0.791±0.027 | 0.798±0.026 | 0.793±0.025 | 0.587±0.130 | 0.796±0.027 | 0.788±0.031 |
| **LEMNERSTOP** | 0.799±0.030 | 0.798±0.025 | 0.819±0.028 | 0.823±0.021 | 0.632±0.099 | 0.831±0.026 | 0.802±0.028 |
| **LEMNERSTOPALPHA** | 0.805±0.025 | 0.808±0.029 | 0.830±0.024 | 0.826±0.020 | 0.643±0.100 | 0.834±0.014 | 0.802±0.019 |
| **LEMPOS** | 0.806±0.021 | 0.808±0.022 | 0.845±0.015 | 0.850±0.023 | 0.716±0.039 | 0.830±0.014 | 0.786±0.024 |
| **LEMPOSALPHA** | 0.807±0.026 | 0.803±0.035 | 0.838±0.017 | 0.846±0.019 | 0.704±0.030 | 0.828±0.026 | 0.778±0.037 |
| **LEMPOSS** | 0.815±0.023 | 0.817±0.017 | 0.851±0.021 | 0.842±0.017 | 0.711±0.178 | 0.855±0.025 | 0.785±0.033 |
| **LEMPOSSALPHA** | 0.825±0.020 | 0.823±0.030 | 0.848±0.018 | 0.837±0.019 | 0.734±0.035 | 0.847±0.020 | 0.770±0.039 |
| **LEMPOSSSTOP** | 0.816±0.025 | 0.818±0.028 | 0.850±0.018 | 0.846±0.017 | 0.529±0.038 | 0.851±0.018 | 0.774±0.031 |
| **LEMPOSSSTOPALPHA** | 0.819±0.036 | 0.817±0.025 | 0.841±0.025 | 0.829±0.016 | 0.722±0.037 | 0.844±0.022 | 0.779±0.034 |
| **LEMPOSSTOP** | 0.802±0.023 | 0.795±0.029 | 0.841±0.015 | 0.847±0.023 | 0.567±0.144 | 0.836±0.021 | 0.770±0.030 |
| **LEMPOSSTOPALPHA** | 0.791±0.017 | 0.789±0.024 | 0.830±0.026 | 0.836±0.019 | 0.700±0.028 | 0.837±0.022 | 0.768±0.035 |
| **LEMSTOP** | 0.799±0.028 | 0.801±0.026 | 0.813±0.019 | 0.802±0.024 | 0.614±0.132 | 0.812±0.032 | 0.785±0.030 |
| **LEMSTOPALPHA** | 0.803±0.031 | 0.804±0.030 | 0.813±0.025 | 0.807±0.031 | 0.619±0.132 | 0.821±0.025 | 0.776±0.032 |
| **POSS** | 0.645±0.038 | 0.646±0.030 | 0.634±0.039 | 0.627±0.046 | 0.558±0.051 | 0.620±0.035 | 0.645±0.041 |
| **POSSALPHA** | 0.646±0.036 | 0.650±0.039 | 0.635±0.051 | 0.619±0.067 | 0.557±0.055 | 0.618±0.032 | 0.644±0.038 |
| **POSSSTOP** | 0.641±0.034 | 0.643±0.016 | 0.634±0.030 | 0.637±0.050 | 0.553±0.083 | 0.626±0.038 | 0.638±0.022 |
| **POSSSTOPALPHA** | 0.615±0.034 | 0.616±0.034 | 0.603±0.063 | 0.610±0.052 | 0.506±0.125 | 0.614±0.045 | 0.610±0.035 |
| **TOK** | 0.801±0.024 | 0.797±0.022 | 0.813±0.023 | 0.806±0.021 | 0.592±0.144 | 0.817±0.022 | 0.766±0.029 |
| **TOKALPHA** | 0.801±0.020 | 0.799±0.025 | 0.807±0.019 | 0.804±0.020 | 0.584±0.135 | 0.817±0.019 | 0.761±0.031 |
| **TOKNER** | 0.814±0.028 | 0.812±0.027 | 0.830±0.018 | 0.828±0.020 | 0.718±0.118 | 0.835±0.018 | 0.763±0.031 |
| **TOKNERALPHA** | 0.798±0.025 | 0.800±0.021 | 0.816±0.015 | 0.820±0.026 | 0.610±0.117 | 0.825±0.018 | 0.790±0.028 |
| **TOKNERR** | 0.789±0.031 | 0.785±0.029 | 0.797±0.017 | 0.799±0.016 | 0.589±0.126 | 0.798±0.020 | 0.783±0.021 |
| **TOKNERRALPHA** | 0.791±0.022 | 0.792±0.027 | 0.802±0.024 | 0.792±0.024 | 0.567±0.126 | 0.797±0.019 | 0.778±0.029 |
| **TOKNERRSTOP** | 0.776±0.032 | 0.778±0.032 | 0.795±0.028 | 0.783±0.021 | 0.582±0.129 | 0.788±0.024 | 0.776±0.025 |
| **TOKNERRSTOPALPHA** | 0.785±0.024 | 0.787±0.035 | 0.796±0.022 | 0.795±0.025 | 0.571±0.127 | 0.783±0.032 | 0.778±0.029 |
| **TOKNERSTOP** | 0.803±0.026 | 0.805±0.031 | 0.825±0.022 | 0.824±0.026 | 0.608±0.114 | 0.822±0.014 | 0.790±0.031 |
| **TOKNERSTOPALPHA** | 0.794±0.030 | 0.794±0.031 | 0.819±0.035 | 0.820±0.015 | 0.616±0.116 | 0.820±0.019 | 0.797±0.031 |
| **TOKPOS** | 0.799±0.020 | 0.796±0.019 | 0.822±0.021 | 0.828±0.016 | 0.630±0.023 | 0.826±0.013 | 0.794±0.027 |
| **TOKPOSALPHA** | 0.808±0.015 | 0.807±0.022 | 0.844±0.025 | 0.852±0.022 | 0.706±0.041 | 0.828±0.018 | 0.777±0.040 |
| **TOKPOSS** | 0.815±0.025 | 0.812±0.014 | 0.843±0.022 | 0.835±0.023 | 0.695±0.191 | 0.849±0.018 | 0.775±0.025 |
| **TOKPOSSALPHA** | 0.788±0.032 | 0.789±0.028 | 0.829±0.022 | 0.833±0.014 | 0.716±0.022 | 0.823±0.015 | 0.752±0.045 |
| **TOKPOSSSTOP** | 0.823±0.025 | 0.814±0.024 | 0.834±0.027 | 0.828±0.014 | 0.713±0.036 | 0.848±0.022 | 0.767±0.029 |
| **TOKPOSSSTOPALPHA** | 0.817±0.023 | 0.822±0.019 | 0.837±0.021 | 0.835±0.013 | 0.518±0.041 | 0.846±0.020 | 0.764±0.036 |
| **TOKPOSSTOP** | 0.802±0.038 | 0.800±0.027 | 0.839±0.019 | 0.840±0.019 | 0.691±0.134 | 0.823±0.020 | 0.771±0.032 |
| **TOKPOSSTOPALPHA** | - | 0.794±0.029 | 0.834±0.021 | 0.844±0.023 | 0.560±0.033 | 0.832±0.015 | 0.766±0.034 |
| **TOKSTOP** | 0.803±0.027 | 0.801±0.027 | 0.815±0.028 | 0.803±0.025 | 0.584±0.133 | 0.809±0.023 | 0.778±0.041 |
| **TOKSTOPALPHA** | 0.800±0.026 | 0.800±0.032 | 0.811±0.024 | 0.813±0.019 | 0.591±0.132 | 0.821±0.019 | 0.779±0.027 |
| **Avg clf F1** | 0.758 | 0.758 | 0.773 | 0.769 | 0.579 | 0.772 | 0.729 |
| **Avg clf stderr** | 0.037 | 0.037 | 0.034 | 0.033 | 0.126 | 0.031 | 0.054 |

| | AdaBoost | XGBoost | MLP | CNN1 | CNN2 | Avg pp F1 | Avg pp stderr |
|---|---|---|---|---|---|---|---|
| **CHNK** | 0.592±0.122 | 0.632±0.091 | 0.761±0.029 | 0.771±0.026 | 0.635±0.026 | 0.681 | 0.063 |
| **CHNKALPHA** | 0.590±0.120 | 0.622±0.089 | 0.738±0.023 | 0.747±0.029 | 0.629±0.029 | 0.675 | 0.056 |
| **CHNKNER** | 0.696±0.057 | 0.714±0.041 | 0.796±0.019 | 0.789±0.039 | - | 0.734 | 0.049 |
| **CHNKNERALPHA** | 0.686±0.059 | 0.712±0.043 | 0.758±0.020 | 0.768±0.025 | - | 0.725 | 0.046 |
| **CHNKNERR** | 0.690±0.059 | 0.713±0.046 | 0.780±0.028 | 0.786±0.020 | 0.699±0.032 | 0.723 | 0.045 |
| **CHNKNERRALPHA** | 0.679±0.064 | 0.703±0.054 | 0.747±0.030 | 0.770±0.027 | 0.669±0.037 | 0.709 | 0.049 |
| **CHNKNERRSTOP** | 0.694±0.057 | 0.700±0.049 | 0.775±0.017 | 0.778±0.029 | 0.665±0.023 | 0.715 | 0.047 |
| **CHNKNERRSTOPALPHA** | 0.676±0.069 | 0.690±0.062 | 0.736±0.028 | 0.727±0.032 | 0.657±0.027 | 0.700 | 0.052 |
| **CHNKNERSTOP** | 0.696±0.053 | 0.709±0.047 | 0.784±0.027 | 0.791±0.025 | - | 0.737 | 0.047 |
| **CHNKNERSTOPALPHA** | 0.692±0.053 | 0.707±0.048 | 0.745±0.030 | 0.752±0.019 | - | 0.718 | 0.047 |
| **CHNKSTOP** | 0.580±0.123 | 0.629±0.092 | 0.753±0.023 | 0.746±0.030 | 0.688±0.021 | 0.681 | 0.067 |
| **CHNKSTOPALPHA** | 0.572±0.133 | 0.628±0.089 | 0.702±0.028 | 0.722±0.023 | 0.604±0.033 | 0.662 | 0.067 |
| **DEP** | 0.534±0.176 | 0.583±0.127 | 0.764±0.027 | 0.765±0.029 | 0.573±0.027 | 0.644 | 0.085 |
| **DEPALPHA** | 0.529±0.179 | 0.571±0.127 | 0.756±0.033 | 0.705±0.019 | 0.621±0.033 | 0.634 | 0.089 |
| **DEPNER** | 0.690±0.065 | 0.707±0.049 | 0.792±0.027 | 0.748±0.021 | - | 0.697 | 0.054 |
| **DEPNERALPHA** | 0.689±0.058 | 0.703±0.045 | 0.782±0.016 | 0.756±0.026 | - | 0.692 | 0.060 |
| **DEPNERR** | 0.520±0.184 | 0.578±0.128 | 0.791±0.035 | 0.773±0.026 | 0.590±0.037 | 0.645 | 0.087 |
| **DEPNERRALPHA** | 0.524±0.182 | 0.576±0.128 | 0.767±0.032 | 0.762±0.027 | 0.587±0.034 | 0.635 | 0.089 |
| **DEPNERRSTOP** | 0.514±0.190 | 0.549±0.152 | 0.783±0.018 | 0.761±0.025 | 0.479±0.034 | 0.619 | 0.095 |
| **DEPNERRSTOPALPHA** | 0.504±0.195 | 0.523±0.175 | 0.755±0.025 | 0.738±0.040 | 0.592±0.020 | 0.603 | 0.111 |
| **DEPNERSTOP** | 0.681±0.070 | 0.690±0.062 | 0.802±0.026 | 0.750±0.023 | - | 0.700 | 0.055 |
| **DEPNERSTOPALPHA** | 0.683±0.066 | 0.685±0.046 | 0.779±0.022 | 0.739±0.020 | - | 0.691 | 0.061 |
| **DEPSTOP** | 0.523±0.183 | 0.556±0.151 | 0.776±0.028 | 0.731±0.028 | 0.659±0.040 | 0.633 | 0.095 |
| **DEPSTOPALPHA** | 0.516±0.189 | 0.535±0.172 | 0.746±0.028 | 0.743±0.027 | 0.608±0.029 | 0.609 | 0.107 |
| **LEM** | 0.669±0.080 | 0.723±0.040 | 0.863±0.017 | 0.868±0.031 | 0.721±0.029 | 0.771 | 0.037 |
| **LEMALPHA** | 0.666±0.080 | 0.727±0.045 | 0.854±0.015 | 0.853±0.022 | 0.805±0.021 | 0.775 | 0.037 |
| **LEMNER** | 0.747±0.042 | 0.775±0.035 | 0.870±0.017 | 0.865±0.020 | 0.616±0.021 | 0.780 | 0.033 |
| **LEMNERALPHA** | 0.744±0.047 | 0.776±0.030 | 0.857±0.019 | 0.860±0.031 | 0.772±0.028 | 0.793 | 0.033 |
| **LEMNERR** | 0.744±0.044 | 0.763±0.037 | 0.843±0.018 | 0.811±0.021 | 0.654±0.022 | 0.763 | 0.034 |
| **LEMNERRALPHA** | 0.748±0.040 | 0.768±0.027 | 0.845±0.029 | 0.828±0.018 | 0.708±0.022 | 0.772 | 0.033 |
| **LEMNERRSTOP** | 0.740±0.037 | 0.756±0.039 | 0.839±0.019 | 0.836±0.016 | 0.807±0.016 | 0.776 | 0.033 |
| **LEMNERRSTOPALPHA** | 0.743±0.041 | 0.764±0.029 | 0.810±0.022 | 0.814±0.014 | 0.776±0.026 | 0.771 | 0.035 |
| **LEMNERSTOP** | 0.742±0.032 | 0.767±0.036 | 0.875±0.019 | 0.870±0.012 | 0.807±0.023 | 0.797 | 0.032 |
| **LEMNERSTOPALPHA** | 0.744±0.038 | 0.769±0.040 | 0.848±0.022 | 0.849±0.034 | 0.741±0.031 | 0.792 | 0.033 |
| **LEMPOS** | 0.742±0.051 | 0.768±0.026 | 0.868±0.021 | 0.870±0.025 | 0.619±0.024 | 0.792 | 0.025 |
| **LEMPOSALPHA** | 0.728±0.053 | 0.772±0.033 | 0.856±0.017 | 0.860±0.015 | 0.663±0.016 | 0.790 | 0.027 |
| **LEMPOSS** | 0.718±0.047 | 0.780±0.033 | 0.866±0.019 | - | 0.768±0.013 | 0.807 | 0.039 |
| **LEMPOSSALPHA** | 0.690±0.064 | 0.759±0.041 | 0.856±0.024 | 0.867±0.024 | 0.572±0.028 | 0.786 | 0.030 |
| **LEMPOSSSTOP** | 0.718±0.030 | 0.764±0.022 | 0.868±0.023 | 0.876±0.020 | 0.820±0.024 | 0.794 | 0.025 |
| **LEMPOSSSTOPALPHA** | 0.686±0.025 | 0.743±0.030 | 0.850±0.015 | 0.851±0.029 | 0.688±0.015 | 0.789 | 0.026 |
| **LEMPOSSTOP** | 0.730±0.034 | 0.769±0.018 | 0.868±0.022 | 0.877±0.027 | 0.808±0.025 | 0.793 | 0.034 |
| **LEMPOSSTOPALPHA** | 0.707±0.043 | 0.754±0.021 | 0.835±0.018 | 0.860±0.028 | 0.783±0.031 | 0.791 | 0.026 |
| **LEMSTOP** | 0.659±0.082 | 0.723±0.054 | 0.858±0.018 | 0.862±0.015 | 0.809±0.017 | 0.778 | 0.040 |
| **LEMSTOPALPHA** | 0.665±0.078 | 0.727±0.059 | 0.842±0.025 | 0.844±0.021 | 0.801±0.027 | 0.777 | 0.043 |
| **POSS** | 0.643±0.029 | 0.632±0.024 | 0.645±0.024 | 0.670±0.032 | 0.490±0.032 | 0.621 | 0.035 |
| **POSSALPHA** | 0.646±0.030 | 0.640±0.029 | 0.649±0.036 | 0.658±0.034 | 0.567±0.030 | 0.627 | 0.040 |
| **POSSSTOP** | 0.643±0.028 | 0.630±0.031 | 0.635±0.028 | 0.623±0.024 | 0.492±0.030 | 0.616 | 0.035 |
| **POSSSTOPALPHA** | 0.609±0.033 | 0.595±0.033 | 0.614±0.047 | 0.587±0.050 | 0.508±0.053 | 0.591 | 0.050 |
| **TOK** | 0.647±0.074 | 0.724±0.051 | 0.864±0.019 | 0.857±0.024 | 0.792±0.013 | 0.773 | 0.039 |
| **TOKALPHA** | 0.655±0.077 | 0.712±0.052 | 0.847±0.020 | 0.848±0.014 | 0.798±0.018 | 0.769 | 0.037 |
| **TOKNER** | 0.667±0.037 | 0.731±0.032 | 0.846±0.018 | 0.847±0.022 | 0.686±0.024 | 0.781 | 0.033 |
| **TOKNERALPHA** | 0.734±0.032 | 0.762±0.038 | 0.870±0.024 | 0.864±0.022 | 0.834±0.030 | 0.794 | 0.033 |
| **TOKNERR** | 0.729±0.037 | 0.756±0.040 | 0.838±0.028 | 0.844±0.016 | 0.793±0.022 | 0.775 | 0.034 |
| **TOKNERRALPHA** | 0.737±0.034 | 0.759±0.034 | 0.836±0.021 | 0.821±0.018 | 0.722±0.026 | 0.766 | 0.034 |
| **TOKNERRSTOP** | 0.727±0.039 | 0.756±0.034 | 0.835±0.020 | 0.843±0.014 | 0.709±0.027 | 0.762 | 0.035 |
| **TOKNERRSTOPALPHA** | 0.736±0.040 | 0.761±0.029 | 0.818±0.016 | 0.817±0.023 | 0.742±0.019 | 0.764 | 0.035 |
| **TOKNERSTOP** | 0.745±0.045 | 0.768±0.033 | 0.842±0.015 | 0.857±0.027 | 0.811±0.019 | 0.792 | 0.034 |
| **TOKNERSTOPALPHA** | 0.735±0.039 | 0.759±0.033 | 0.878±0.028 | 0.875±0.027 | 0.849±0.041 | 0.796 | 0.037 |
| **TOKPOS** | 0.730±0.042 | 0.763±0.036 | 0.846±0.025 | 0.847±0.022 | 0.773±0.031 | 0.788 | 0.025 |
| **TOKPOSALPHA** | 0.724±0.056 | 0.770±0.030 | 0.870±0.015 | 0.865±0.019 | 0.674±0.019 | 0.794 | 0.027 |
| **TOKPOSS** | 0.719±0.057 | 0.756±0.031 | 0.876±0.031 | - | 0.683±0.021 | 0.795 | 0.042 |
| **TOKPOSSALPHA** | 0.715±0.068 | 0.748±0.048 | 0.864±0.019 | 0.865±0.020 | 0.646±0.014 | 0.781 | 0.029 |
| **TOKPOSSSTOP** | 0.701±0.025 | 0.760±0.021 | 0.855±0.022 | 0.821±0.021 | 0.643±0.022 | 0.784 | 0.024 |
| **TOKPOSSSTOPALPHA** | 0.703±0.029 | 0.764±0.021 | 0.876±0.031 | 0.883±0.020 | 0.759±0.019 | 0.785 | 0.024 |
| **TOKPOSSTOP** | 0.715±0.030 | 0.763±0.025 | 0.839±0.014 | 0.857±0.022 | 0.615±0.014 | 0.780 | 0.033 |
| **TOKPOSSTOPALPHA** | 0.720±0.044 | 0.762±0.032 | - | - | - | 0.775 | 0.029 |
| **TOKSTOP** | 0.646±0.078 | 0.717±0.055 | 0.857±0.013 | 0.861±0.013 | 0.782±0.023 | 0.771 | 0.041 |
| **TOKSTOPALPHA** | 0.650±0.080 | 0.717±0.058 | 0.827±0.016 | 0.818±0.021 | 0.710±0.050 | 0.761 | 0.042 |
| **Avg clf F1** | 0.673 | 0.707 | 0.809 | 0.806 | 0.688 | | |
| **Avg clf stderr** | 0.049 | 0.023 | 0.026 | 0.031 | 0.036 | | |